\documentclass{surveymeta}
\usepackage{amsmath,amsfonts,amssymb}
\usepackage{algorithmic}
\usepackage{algorithm}
\usepackage{array}
\usepackage{textcomp}
\usepackage{url}
\usepackage{verbatim}
\usepackage{graphicx}
\usepackage[nocompress]{cite}
\usepackage{booktabs, multirow, makecell, adjustbox, pifont, colortbl, xcolor}
\usepackage{threeparttable}
\usepackage{comment}
\usepackage{marvosym}
\usepackage{enumitem}
\usepackage{tikz}
\usepackage{ragged2e}
\usepackage{longtable}
\usepackage{booktabs}
\usetikzlibrary{calc}
\usepackage{fontawesome5}
\definecolor{zebragray}{HTML}{F7F5F1}         
\definecolor{textgray}{HTML}{6B7084}          

\definecolor{blueBg}{HTML}{EAF0FA}     \definecolor{blueBd}{HTML}{7B9ACC}
\definecolor{greenBg}{HTML}{E8F4E8}    \definecolor{greenBd}{HTML}{6EA26E}
\definecolor{orangeBg}{HTML}{FDF3E6}   \definecolor{orangeBd}{HTML}{C4974A}
\definecolor{purpleBg}{HTML}{F0EAF8}   \definecolor{purpleBd}{HTML}{9B7EC0}
\definecolor{pinkBg}{HTML}{FAE8EC}     \definecolor{pinkBd}{HTML}{C47888}
\definecolor{tealBg}{HTML}{E6F2F0}     \definecolor{tealBd}{HTML}{6EA8A0}
\definecolor{grayBg}{HTML}{F0F0F0}     \definecolor{grayBd}{HTML}{A0A0A0}

\definecolor{qaBg}{HTML}{EAF0FA}       \definecolor{qaBd}{HTML}{7B9ACC}
\definecolor{lfBg}{HTML}{F0EAF8}       \definecolor{lfBd}{HTML}{9B7EC0}
\definecolor{maBg}{HTML}{FAE8EC}       \definecolor{maBd}{HTML}{C47888}
\definecolor{coBg}{HTML}{FDF3E6}       \definecolor{coBd}{HTML}{C4974A}
\definecolor{agBg}{HTML}{E8F4E8}       \definecolor{agBd}{HTML}{6EA26E}
\definecolor{muBg}{HTML}{F0F0F0}       \definecolor{muBd}{HTML}{A0A0A0}

\definecolor{dotE}{HTML}{3D5A80}
\definecolor{dotS}{HTML}{6EA26E}
\definecolor{dotP}{HTML}{C4974A}

\definecolor{contribframe}{HTML}{8B6834}
\definecolor{contribtitle}{HTML}{8B6834}
\definecolor{contribback}{HTML}{F9F6F0}
\definecolor{contriblabel}{HTML}{5C3D2E}
\definecolor{contribbullet}{HTML}{8B6834}

\newcommand{\cmark}{{\color{black}\ding{51}}}
\newcommand{\na}{{\color{textgray}--}}
\newcommand{\ygray}[1]{{\color{textgray}#1}}
\newcommand{\rowstrut}{\rule[-0.55ex]{0pt}{2.35ex}}

\newcolumntype{L}[1]{>{\RaggedRight\arraybackslash}m{#1}}
\newcolumntype{C}[1]{>{\centering\arraybackslash}m{#1}}

\newcommand{\badge}[3]{%
\tikz[baseline=(X.base)]\node
(X)[draw=#2,fill=#1,rounded corners=1.6pt,
inner xsep=4pt,inner ysep=1.2pt,line width=0.4pt,font=\footnotesize]{#3};%
}

\newcommand{\weakcap}[1]{%
\tikz[baseline=(X.base)]\node
(X)[draw=grayBd,fill=grayBg,rounded corners=1.5pt,
inner xsep=3.5pt,inner ysep=1.0pt,line width=0.35pt,font=\footnotesize]{#1};%
}

\newcommand{\KE}{\badge{blueBg}{blueBd}{KE}}
\newcommand{\PE}{\badge{greenBg}{greenBd}{PE}}
\newcommand{\ESS}{\badge{orangeBg}{orangeBd}{ESS}}
\newcommand{\VRP}{\badge{purpleBg}{purpleBd}{VRP}}
\newcommand{\OV}{\badge{pinkBg}{pinkBd}{OV}}
\newcommand{\RC}{\badge{tealBg}{tealBd}{RC}}

\newcommand{\QA}{\badge{qaBg}{qaBd}{QA}}
\newcommand{\LF}{\badge{lfBg}{lfBd}{LF}}
\newcommand{\MA}{\badge{maBg}{maBd}{Ma}}
\newcommand{\CO}{\badge{coBg}{coBd}{Co}}
\newcommand{\AG}{\badge{agBg}{agBd}{Ag}}
\newcommand{\MU}{\badge{muBg}{muBd}{Mu}}
\newcommand{\Proactive}{\badge{blueBg}{blueBd}{Proactive}}
\newcommand{\Progressive}{\badge{orangeBg}{orangeBd}{Progressive}}
\newcommand{\MCTSb}{\badge{greenBg}{greenBd}{MCTS}}
\newcommand{\AStarb}{\badge{purpleBg}{purpleBd}{A*}}
\newcommand{\Retrieveb}{\badge{tealBg}{tealBd}{Retrieve}}
\newcommand{\Diffusionb}{\badge{pinkBg}{pinkBd}{Diffusion}}
\newcommand{\Greedyb}{\badge{grayBg}{grayBd}{Greedy}}

\newcommand{\Flat}{\badge{blueBg}{blueBd}{Flat}}
\newcommand{\Tiered}{\badge{purpleBg}{purpleBd}{Tiered}}
\newcommand{\Graph}{\badge{greenBg}{greenBd}{Graph}}
\newcommand{\TreeS}{\badge{orangeBg}{orangeBd}{Tree}}
\newcommand{\Typed}{\badge{pinkBg}{pinkBd}{Typed}}
\newcommand{\Unified}{\badge{tealBg}{tealBd}{Unified}}

\newcommand{\ClosedSet}{\badge{greenBg}{greenBd}{Closed}}
\newcommand{\OpenSet}{\badge{blueBg}{blueBd}{Open}}
\newcommand{\DynamicSet}{\badge{orangeBg}{orangeBd}{Dynamic}}

\newcommand{\dotfilled}[1]{\tikz[baseline=-0.55ex]\fill[#1] (0,0) circle (0.55ex);}
\newcommand{\dotempty}{\tikz[baseline=-0.55ex]\draw[textgray!70, line width=0.35pt] (0,0) circle (0.55ex);}

\newcommand{\memE}{\dotfilled{dotE}}
\newcommand{\memS}{\dotfilled{dotS}}
\newcommand{\memP}{\dotfilled{dotP}}

\newcommand{\caphollow}[2]{%
\tikz[baseline=(X.base)]\node
(X)[draw=#2,fill=white,rounded corners=1.6pt,
inner xsep=3.8pt,inner ysep=1.0pt,line width=0.35pt,font=\footnotesize]{#1};%
}
\newcommand{\Rpri}{\badge{blueBg}{blueBd}{R}}
\newcommand{\Mpri}{\badge{greenBg}{greenBd}{M}}
\newcommand{\Ppri}{\badge{orangeBg}{orangeBd}{P}}
\newcommand{\Tpri}{\badge{tealBg}{tealBd}{T}}
\newcommand{\Rsec}{\caphollow{R}{blueBd}}
\newcommand{\Msec}{\caphollow{M}{greenBd}}
\newcommand{\Psec}{\caphollow{P}{orangeBd}}
\newcommand{\Tsec}{\caphollow{T}{tealBd}}
\newcommand{\memES}{\memE\hspace{2pt}\memS\hspace{2pt}\dotempty}
\newcommand{\memEonly}{\memE\hspace{2pt}\dotempty\hspace{2pt}\dotempty}
\newcommand{\memSonly}{\dotempty\hspace{2pt}\memS\hspace{2pt}\dotempty}
\newcommand{\memESP}{\memE\hspace{2pt}\memS\hspace{2pt}\memP}

\definecolor{aTwoABg}{HTML}{EAF0FA}
\definecolor{aTwoABd}{HTML}{7B9ACC}

\definecolor{aTwoTBg}{HTML}{FDF3E6}
\definecolor{aTwoTBd}{HTML}{C4974A}

\definecolor{aTwoSBg}{HTML}{E6F2F0}
\definecolor{aTwoSBd}{HTML}{6EA8A0}

\newcommand{\AIIA}{\badge{aTwoABg}{aTwoABd}{A2A}}
\newcommand{\AIIT}{\badge{aTwoTBg}{aTwoTBd}{A2T}}
\newcommand{\AIIS}{\badge{aTwoSBg}{aTwoSBd}{A2S}}

\newcommand{\PDcap}{\weakcap{PD}}
\newcommand{\JSONRPCcap}{\weakcap{JSON-RPC}}
\newcommand{\JSONLDcap}{\weakcap{JSON-LD}}
\newcommand{\JSONLDNLcap}{\weakcap{JSON-LD+NL}}
\newcommand{\JSONLDTDcap}{\weakcap{JSON-LD+TD}}
\newcommand{\JSONRPCSSEcap}{\weakcap{JSON-RPC/SSE}}
\newcommand{\RESTHTTPcap}{\weakcap{REST/HTTP}}

\definecolor{distBg}{HTML}{EAF0FA}
\definecolor{distBd}{HTML}{7B9ACC}

\definecolor{centBg}{HTML}{FDF3E6}
\definecolor{centBd}{HTML}{C4974A}

\definecolor{hybBg}{HTML}{E6F2F0}
\definecolor{hybBd}{HTML}{6EA8A0}

\newcommand{\DistributedB}{\badge{distBg}{distBd}{Distributed}}
\newcommand{\CentralizedB}{\badge{centBg}{centBd}{Centralized}}
\newcommand{\HybridB}{\badge{hybBg}{hybBd}{Hybrid}}

\newcommand{\CoopCap}{\weakcap{Coop}}
\newcommand{\CompCap}{\weakcap{Comp}}

\definecolor{judgeBg}{HTML}{EAF0FA}
\definecolor{judgeBd}{HTML}{7B9ACC}

\definecolor{learnBg}{HTML}{FDF3E6}
\definecolor{learnBd}{HTML}{C4974A}

\definecolor{causalBg}{HTML}{F0EAF8}
\definecolor{causalBd}{HTML}{9B7EC0}

\definecolor{humanBg}{HTML}{E8F4E8}
\definecolor{humanBd}{HTML}{6EA26E}

\definecolor{analytBg}{HTML}{E6F2F0}
\definecolor{analytBd}{HTML}{6EA8A0}

\newcommand{\JudgeB}{\badge{judgeBg}{judgeBd}{LLM-As-Judge}}

\newcommand{\RLB}{\badge{learnBg}{learnBd}{RL}}
\newcommand{\KDB}{\badge{learnBg}{learnBd}{KD}}
\newcommand{\SFTRLB}{\badge{learnBg}{learnBd}{SFT+RL}}
\newcommand{\SFTB}{\badge{learnBg}{learnBd}{SFT}}
\newcommand{\SFTRLCLB}{\badge{learnBg}{learnBd}{SFT+RL+CL}}
\newcommand{\CLB}{\badge{learnBg}{learnBd}{CL}}

\newcommand{\CounterfactualB}{\badge{causalBg}{causalBd}{Counterfactual}}

\newcommand{\CfASEB}{\badge{causalBg}{causalBd}{Cf-ASE}}
\newcommand{\TCFEB}{\badge{causalBg}{causalBd}{TCFE}}

\newcommand{\PreActB}{\badge{humanBg}{humanBd}{PreAct}}
\newcommand{\HITLB}{\badge{humanBg}{humanBd}{HITL}}

\newcommand{\SAB}{\badge{analytBg}{analytBd}{SA}}
\newcommand{\StageCompB}{\badge{analytBg}{analytBd}{Stage-comparison}}

\newcommand{\GraphBasedB}{\badge{analytBg}{analytBd}{Graph-based}}

\newcommand{\OfflineCap}{\weakcap{Offline}}
\newcommand{\OnlineCap}{\weakcap{Online}}
\newcommand{\BothCap}{\weakcap{Both}}
\newcommand{\SelfCorrB}{\badge{selfcorrBg}{selfcorrBd}{Self-Corr}}
\newcommand{\CausalDiscB}{\badge{causalBg}{causalBd}{Causal-Disc}}
\newcommand{\CausalGraphB}{\badge{causalBg}{causalBd}{Causal-Graph}}

\definecolor{selfcorrBg}{HTML}{FDF3E6}    \definecolor{selfcorrBd}{HTML}{C4974A}
\definecolor{causalBg}{HTML}{F0EAF8}      \definecolor{causalBd}{HTML}{9B7EC0}

\definecolor{locBg}{HTML}{EAF0FA}
\definecolor{locBd}{HTML}{7B9ACC}

\definecolor{catBg}{HTML}{FDF3E6}
\definecolor{catBd}{HTML}{C4974A}

\definecolor{intBg}{HTML}{E8F4E8}
\definecolor{intBd}{HTML}{6EA26E}

\definecolor{traceBg}{HTML}{F0EAF8}
\definecolor{traceBd}{HTML}{9B7EC0}

\newcommand{\PromptB}{\badge{blueBg}{blueBd}{Prompt}}
\newcommand{\MemoryB}{\badge{greenBg}{greenBd}{Memory}}
\newcommand{\ParamB}{\badge{orangeBg}{orangeBd}{Param.}}

\newcommand{\GraphB}{\badge{purpleBg}{purpleBd}{Graph}}
\newcommand{\RoleB}{\badge{pinkBg}{pinkBd}{Role}}

\newcommand{\KSB}{\badge{blueBg}{blueBd}{KS}}
\newcommand{\GenB}{\badge{greenBg}{greenBd}{Gen}}
\newcommand{\LocalizationB}{\badge{locBg}{locBd}{Localization}}
\newcommand{\CategoryB}{\badge{catBg}{catBd}{Category}}



\shorttitle{Survey: Multi-Agent Collaboration, Attribution \& Evolution}

\title{\texorpdfstring{\raisebox{-0.1\height}{\includegraphics[height=1.15em]{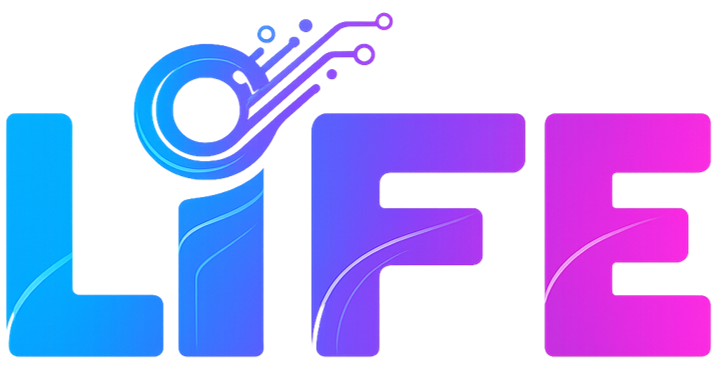}}\hspace{0.7em}Beyond Individual Intelligence: Surveying Collaboration, Failure Attribution, and Self-Evolution in LLM-based Multi-Agent Systems}}
\author{
\begin{tabular}{c}
\mbox{Shihao Qi\textsuperscript{\textdagger,1,2}},\quad
\mbox{Jie Ma\textsuperscript{\textdagger,\Letter,1,3}},\quad
\mbox{Rui Xing\textsuperscript{\textdagger,1,4}},\quad
\mbox{Wei Guo\textsuperscript{\textdagger,1,5}},\quad
\mbox{Xiao Huang\textsuperscript{\textdagger,1,4}}\\[0.35em]
\mbox{Zhitao Gao\textsuperscript{2,6}},\quad
\mbox{Jianhao Deng\textsuperscript{2}},\quad
\mbox{Jun Liu\textsuperscript{1,2,6}},\quad
\mbox{Lingling Zhang\textsuperscript{1,2,6}},\quad
\mbox{Bifan Wei\textsuperscript{1,6}}\\[0.35em]
\mbox{Boqian Yang\textsuperscript{2}},\quad
\mbox{Pinghui Wang\textsuperscript{1,5}},\quad
\mbox{Jianwen
Sun\textsuperscript{7}},\quad
\mbox{Jing Tao\textsuperscript{1,3}}\\[0.35em]
\mbox{Yaqiang Wu\textsuperscript{2,8}},\quad
\mbox{Hui Liu\textsuperscript{8}},\quad
\mbox{Yu Yao\textsuperscript{9}},\quad
\mbox{Tongliang Liu\textsuperscript{9}}
\end{tabular}
}

\affiliation{
\makebox[\linewidth][c]{%
\parbox{0.88\linewidth}{\centering
\textsuperscript{1} MOE KLINNS Lab\\
\textsuperscript{2} School of Computer Science and Technology, Xi'an Jiaotong University\\
\textsuperscript{3} School of Cyber Science and Engineering, Xi'an Jiaotong University\\
\textsuperscript{4} School of Software Engineering, Xi'an Jiaotong University\\
\textsuperscript{5} School of Control Science and Engineering, Xi'an Jiaotong University\\
\textsuperscript{6} Shaanxi Provincial Key Laboratory of Big Data Knowledge Engineering\\
\textsuperscript{7} Laboratory for AI and New Forms of Education, Central China Normal University\\
\textsuperscript{8} Lenovo AI Technology Center, CTOO, Lenovo\\
\textsuperscript{9} Sydney AI Centre, The University of Sydney
}
}
}

\contribution{\textsuperscript{\textdagger} Equal contribution \quad \textsuperscript{\Letter} Corresponding author}

\metadata[GitHub]{{\color{surveyaccent}\faGithub\ \href{https://github.com/mira-ai-lab/Awesome_MAS_Evolution}{\texttt{\nolinkurl{https://github.com/mira-ai-lab/awesome-mas-life}}}}}
\abstract{LLM-based autonomous agents have demonstrated strong capabilities in reasoning, planning, and tool use, yet remain limited when tasks require sustained coordination across roles, tools, and environments. Multi-agent systems address this limitation through structured collaboration among specialized agents, but tighter coordination also amplifies a less explored risk: errors can propagate across agents and interaction rounds, producing failures that are difficult to diagnose and, even once identified, rarely translate into structural self-improvement. Existing surveys have separately covered individual agent capabilities, multi-agent collaboration, or agent self-evolution, but treat these topics in isolation—leaving the causal dependencies among them largely unexamined. This survey provides a unified and comprehensive review organized around four causally linked stages, which we term the \textbf{LIFE} progression: \textbf{L}ay the capability foundation, \textbf{I}ntegrate agents through collaboration, \textbf{F}ind faults through attribution, and \textbf{E}volve through autonomous self-improvement. We review the capability foundations of individual agents, the organizational mechanisms of multi-agent collaboration, the methodological landscape of failure attribution, and the hierarchical design space of self-evolution. Throughout, we formally characterize the dependencies between adjacent stages, revealing how each stage both depends on and constrains the next. Beyond synthesizing existing work, we identify open challenges at the boundaries between LIFE stages and propose a cross-stage research agenda for closed-loop multi-agent systems capable of continuously diagnosing failures, reorganizing collaborative structures, and refining agent behaviors, thereby extending current human-engineered coordination frameworks toward more self-organizing and resilient forms of collective intelligence. By bridging these previously fragmented research threads into a coherent progression, this survey aims to offer both a systematic reference for current research and a conceptual roadmap toward autonomous, self-improving multi-agent intelligence.}

\begin{document}
\maketitle
\clearpage
\tableofcontents
\clearpage
\section{Introduction}
Large language models (LLMs) have rapidly evolved from fluent text generators into systems capable of complex reasoning, long-horizon planning, and interaction with external environments~\cite{openai2025gpt5,deepseek2024v3,yang2025qwen3,anthropic2026opus46}. This progression has been accompanied by the emergence of chain-of-thought capabilities~\cite{wei2022cot,kojima2022zeroshot}, advances in scalable instruction tuning~\cite{wei2022finetuned}, and reinforcement learning--driven reasoning~\cite{guo2025deepseekr1}, which together have substantially expanded the ability of LLMs to tackle mathematical problem-solving, competitive programming, and professional examinations~\cite{guo2025deepseekr1,openai2026gpt54}. Despite these advances, individual LLMs still struggle to maintain coherent behavior over extended interactions, adapt to rapidly changing task conditions, and coordinate multi-step actions that span diverse tools and environments~\cite{wang2024survey,xi2025rise}. These gaps become especially pronounced in real-world deployments that demand sustained autonomy, including autonomous software engineering~\cite{jimenez2024swebench,qian2024chatdev}, AI-driven scientific discovery~\cite{boiko2023autonomous,bran2024chemcrow}, and embodied robotic control~\cite{zitkovich2023rt2,huang2023innermonologue}, where success hinges not on a single inference pass but on persistent observation, iterative decision-making, and closed-loop adaptation. This recognition has catalyzed the emergence of LLM-based autonomous agents: systems that augment foundation models with explicit modules for perception, memory, planning, and tool use, operating through continuous cycles of reasoning, action, and reflection~\cite{wang2024survey,xi2025rise,yao2023react,shinn2023reflexion}.

While these capabilities have significantly raised the ceiling of what a single agent can accomplish, a fundamental tension emerges as tasks grow in scope and duration: the diverse competencies they demand become increasingly difficult to concentrate within a single agent without encountering conflicts in prompt design, context allocation, and behavioral objectives~\cite{pan2025whymultiagentfail}. Multi-agent collaboration offers a principled resolution by decomposing complex tasks into specialized roles that operate through structured communication and coordinated orchestration~\cite{hong2024metagpt,guo2024multiagentsurvey}. Recent systems have moved well beyond early explorations: role-specialized agent teams now collaboratively solve tasks that require sustained multi-step coordination across planning, execution, and verification~\cite{wu2024autogen,park2025maporl}, while standardized communication protocols such as MCP and A2A are beginning to establish the infrastructure for open, interoperable agent ecosystems~\cite{anthropic2024mcp,google2025a2a}. These developments signal a shift from building individual capable agents to engineering the organizational structures that govern how agents coordinate at scale. Yet this very capacity for tight coordination introduces a less examined consequence: as agents become more deeply interdependent, the system's collective behavior grows increasingly difficult to predict, diagnose, and improve when failures occur.

In tightly coupled multi-agent systems, a localized error such as a hallucinated fact, a misrouted message, or an incorrect tool invocation can propagate through successive interaction rounds, triggering cascading failures that obscure the original root cause~\cite{pan2025whymultiagentfail,zhang2025whichagent}. As execution trajectories grow longer and inter-agent dependencies deepen, the causal chain linking a failure to its downstream consequences becomes increasingly opaque, rendering manual identification of the responsible agent and the decisive step both inefficient and unscalable~\cite{deshpande2025trail,ma2025diagnosing,wang2026chief}. Meanwhile, even when root causes are successfully identified, existing multi-agent systems largely lack the ability to translate diagnostic insights into structural adaptation, whether by reorganizing coordination topologies, revising role assignments, or refining collaboration policies in light of observed failure patterns~\cite{gao2025selfevolvingsurvey,dang2025evolvingorchestration}. These two deficits in attribution and in self-correction are deeply intertwined: without reliable diagnosis, improvement efforts lack direction; without the ability to act on diagnostic results, attribution yields limited practical value~\cite{zhu2025agenterrorbench}. This coupling exposes a fundamental asymmetry: the capacity to build increasingly sophisticated multi-agent systems has far outpaced the machinery available for understanding why they fail and enabling them to improve autonomously. Bridging this gap requires a closed-loop framework in which failure diagnosis directly informs structural self-improvement.

The growing significance of {LLM}-based agent systems has prompted a substantial body of survey work, yet existing reviews remain compartmentalized along the boundaries identified above. One line of surveys systematically examines the core capabilities of individual agents, covering reasoning, memory, planning, and tool use as modular components of a single decision-making system~\cite{wang2024survey,xi2025rise,zhang2025memorysurvey,wei2025plangenllms,liu2025foundationagents}. A second line shifts focus to multi-agent collaboration, analyzing how role specialization, communication protocols, orchestration topologies, and interaction patterns enable collective problem-solving~\cite{guo2024multiagentsurvey,li2024llmbasedmas,tran2025collaborationmechanisms,yan2025beyond,ehtesham2025interop,han2024maschallenges}. A third, more recent line addresses agent self-evolution, investigating how agents can autonomously refine their prompts, memory, tools, and workflows through environmental feedback~\cite{fang2025comprehensiveselfevolving,gao2025selfevolvingsurvey}. These lines collectively cover considerable ground, but they leave the post-deployment failure loop largely unexamined. Notably, no existing survey provides a systematic treatment of multi-agent failure attribution, despite its emergence as a distinct and rapidly growing research area~\cite{zhang2025whichagent,pan2025whymultiagentfail,zhu2025agenterrorbench}. More fundamentally, no prior work has examined the complete operational lifecycle of multi-agent systems as an integrated whole: from foundational single-agent capabilities, through collaborative organization, to failure diagnosis and autonomous structural improvement. This survey is motivated by precisely this gap.

\begin{figure*}[t]
    \centering

    \includegraphics[width=\textwidth]{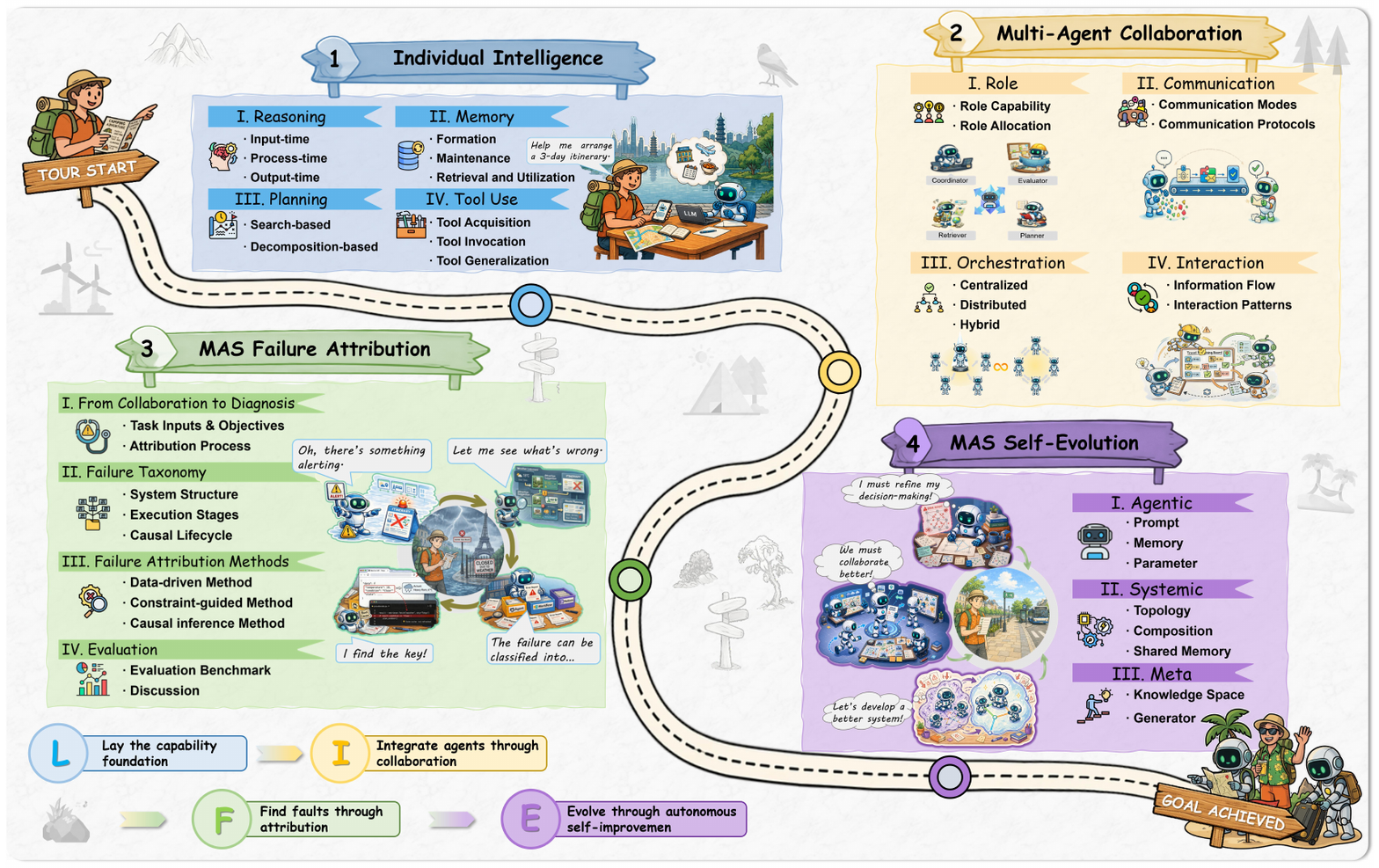}
    \caption{Illustrative overview of the operational lifecycle for LLM-based multi-agent systems, using a collaborative travel-planning scenario as a running example. The winding pathway connects four causally linked stages from \textsc{Start} to \textsc{Destination}: \ding{182}~\textbf{Individual Intelligence}, \ding{183}~\textbf{Multi-Agent Collaboration}, \ding{184}~\textbf{Multi-Agent Attribution}, and \ding{185}~\textbf{Multi-Agent System Self-Evolution}. Each stage is accompanied by a taxonomy panel summarizing the key dimensions and sub-topics reviewed in this survey.}
    \label{fig:overview}
\end{figure*}

To the best of our knowledge, this is the first survey that covers the complete operational lifecycle of LLM-based multi-agent systems, from foundational single-agent capabilities, through collaborative organization, to failure diagnosis and autonomous structural improvement. Figure~\ref{fig:overview} provides a visual overview of the operational lifecycle and the representative work reviewed at each stage. We organize this lifecycle as a causally linked progression termed \textbf{LIFE}, and summarize our main contributions below.
\begin{center}
\begin{tcolorbox}[
    enhanced,
    colframe=contribframe,
    colback=contribback,
    coltitle=white,
    colbacktitle=contribtitle,
    fonttitle=\normalsize\bfseries\sffamily,
    title={\normalsize Contributions},
    arc=2mm,
    boxrule=0.35mm,
    boxsep=3pt,
    left=5pt,
    right=5pt,
    top=4pt,
    bottom=4pt]

\begin{itemize}[
    leftmargin=*,
    itemsep=6pt,
    labelsep=5pt,
    label={\color{contribbullet}$\blacktriangleright$}
]

\item \textbf{LIFE-oriented perspective.} We organize LLM-based multi-agent systems through the LIFE progression, a unified analytical lens that connects individual intelligence, collaboration, failure attribution, and self-evolution as four causally linked stages, moving beyond prior surveys that largely treat these topics separately.
\item \textbf{Stage-wise structured synthesis.} Within each stage, we provide systematic taxonomies and comparative analyses of the field, covering core capabilities for individual intelligence, organizational mechanisms for collaboration, methodological families for failure attribution, and hierarchical scopes for self-evolution.
\item \textbf{Attribution-driven evolution loop.} Enabled by the LIFE perspective, we highlight an attribution–evolution closed loop: failure attribution not only explains what went wrong, but also narrows the search space for targeted self-improvement. In turn, collaborative structures shape what failures can be observed and attributed, a coupling that remains underexplored in prior survey literature.
\item \textbf{Challenges, agenda, and resources.} We identify open challenges that arise at stage boundaries, outline a cross-stage research agenda, and curate a publicly maintained repository of structured literature, taxonomy visualizations, and comparative tables to support future research and community use.
\end{itemize}

\end{tcolorbox}
\end{center}

\begin{figure*}[t]
    \centering
    \label{overview}
    \includegraphics[width=\textwidth]{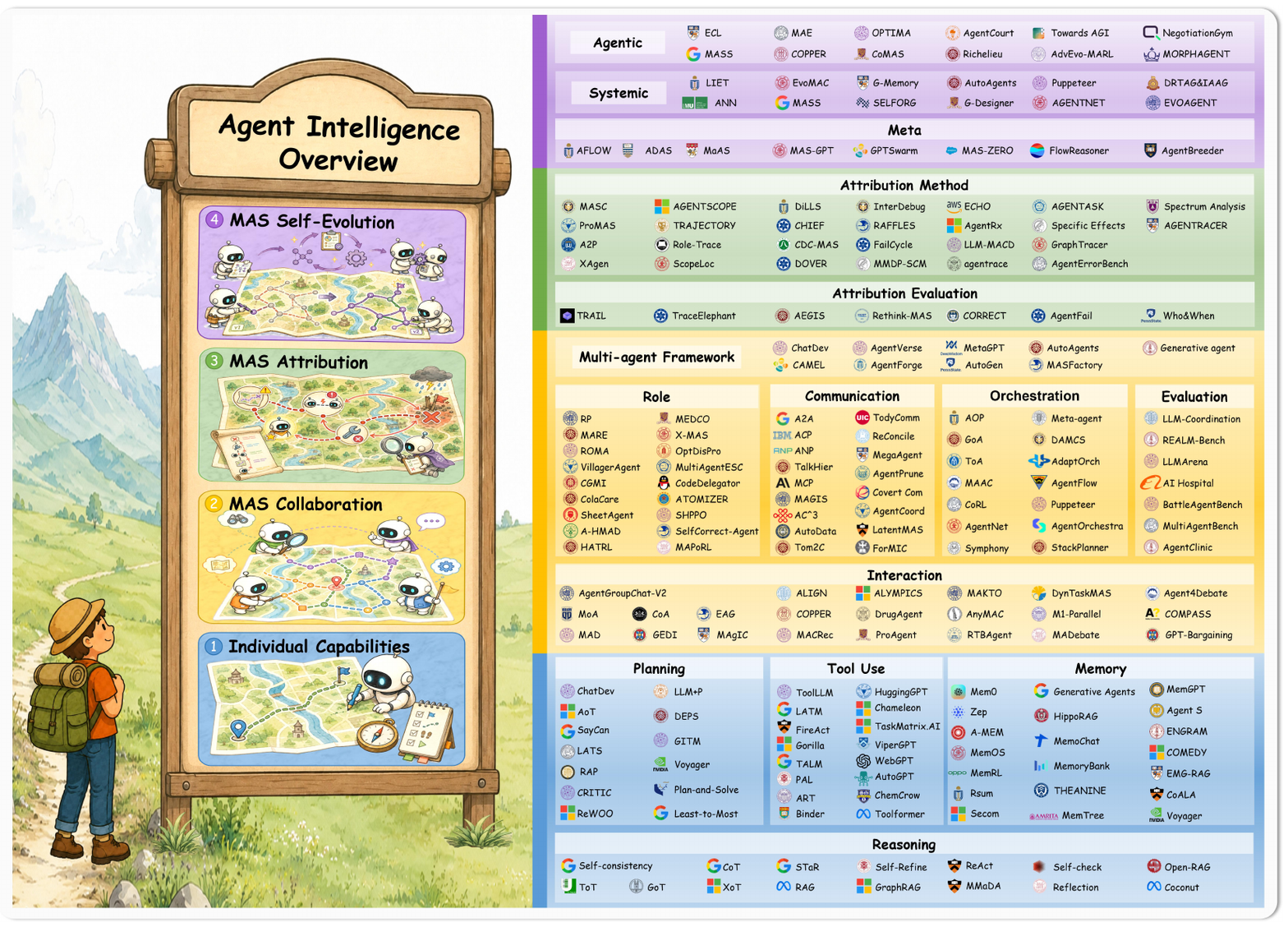}
    \caption{Overview of LLM-based multi-agent systems organized along an operational lifecycle. The right side summarizes representative topics and systems in each stage, and the left-side axis visualizes the cross-stage progression from action to evolution.}
    \label{fig:overview_lifecycle}
\end{figure*}

\section{Individual Intelligence}
\label{sec:individual_intelligence}

\subsection{From LLM to LLM-based Agent: An Architectural Overview}

\label{subsec:overview}

While large language models have demonstrated remarkable capabilities in language understanding and generation~\cite{openai2023gpt4,touvron2023llama}, their limitations become evident in complex real-world scenarios such as automated software development~\cite{jimenez2024swebench}, scientific experiment design~\cite{bran2024chemcrow}, and multi-step web navigation~\cite{zhou2024webarena}. In these settings, LLMs operate as stateless response generators: they lack persistent memory, cannot actively perceive or act upon external environments, and remain confined to the static knowledge encoded during training. These constraints motivate the transition from reactive text generators to LLM-based autonomous agents~\cite{wang2024survey}, which extend the foundation model with the ability to perceive environmental states, formulate goals, execute actions through external tools, and adapt behavior based on feedback---operating through an iterative loop of observation, reasoning, action, and reflection~\cite{yao2023react,shinn2023reflexion}. To systematically characterize the capabilities that enable such autonomy, we organize the core competencies of an individual agent into four dimensions: \textit{reasoning}, \textit{memory}, \textit{planning}, and \textit{tool use}~\cite{wang2024survey,xi2023rise}. Below we formalize their interactions within a unified decision-making framework.
\begin{figure*}[t]
    \centering
    \label{individual}
    \includegraphics[width=\textwidth]{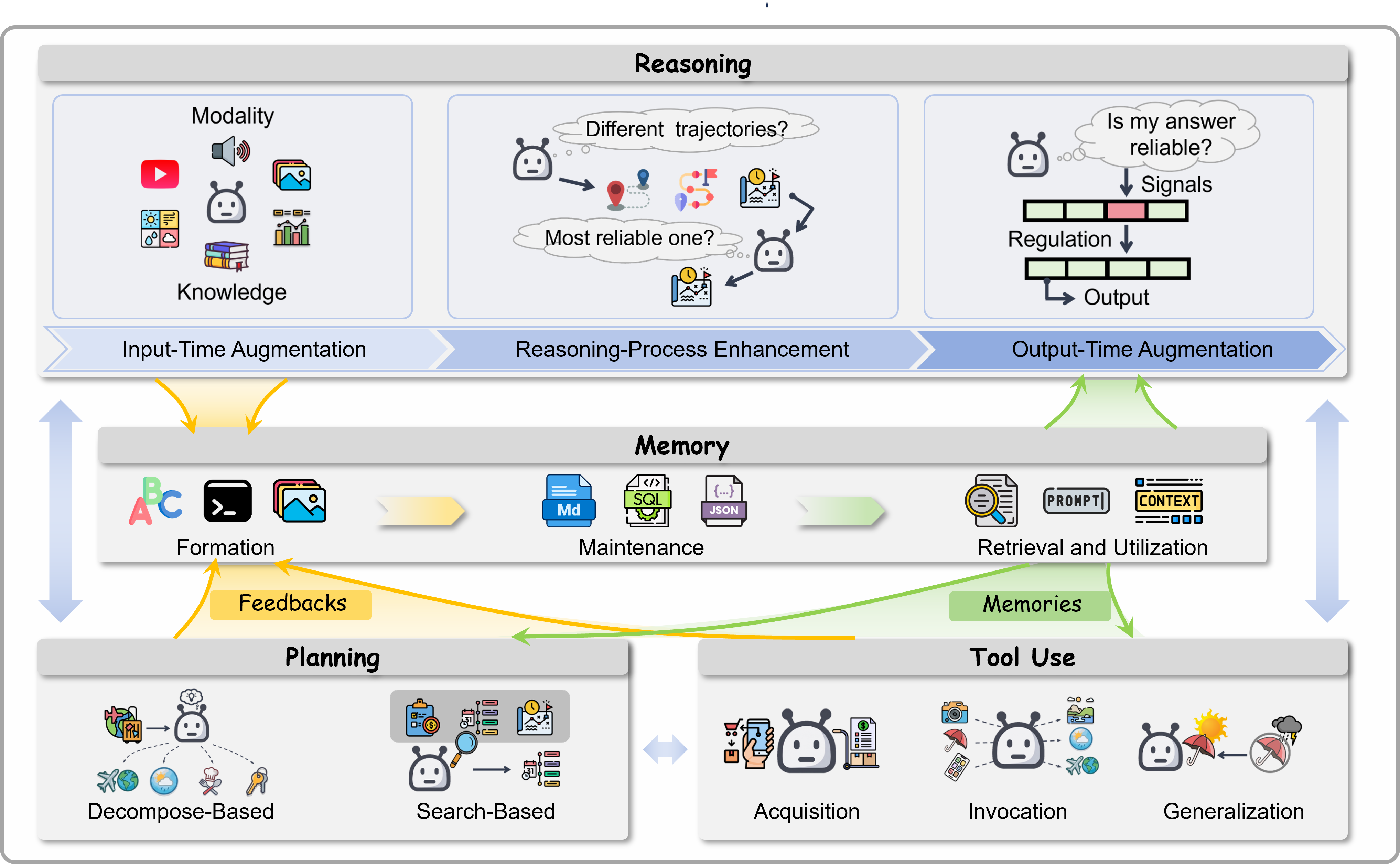}
   \caption{Architecture of individual agent capabilities organized along the reasoning pipeline. Memory serves as a cross-cutting module throughout this process, providing retrieved context and incorporating execution feedback. Planning and tool use are further elaborated as core components of the processing stage.}
    \label{fig:individual}
\end{figure*}

We model an LLM-based agent as a structured sequential decision-making system with explicit internal functional decomposition. The agent $\mathfrak{a}$ is defined by the tuple:
\begin{equation}
\mathfrak{a}
=
(\mathcal{X},\; \mathcal{U},\; \mathcal{O},\; \mathcal{M},\; R,\; \mathcal{P},\; \mathcal{T}),
\end{equation}
where $\mathcal{X}$ denotes the observation space, $\mathcal{U}$ the action space, $\mathcal{O}$ the space of tool-returned observations, and $\mathcal{M}$ the memory state space. The functions $R$, $\mathcal{P}$, and $\mathcal{T}$ correspond to reasoning, planning, and tool execution, respectively. At each discrete time step $t = 1, 2, \dots$, the agent receives an observation $x_t \in \mathcal{X}$ and maintains an internal memory state $m_t \in \mathcal{M}$. The decision-making process is factorized into the following stages.

First, the agent retrieves task-relevant context from its memory state:

\begin{equation}
c_t = \phi(x_t,\, m_t).
\end{equation}

Second, a reasoning module produces an intermediate representation based on the current observation and retrieved context:

\begin{equation}
r_t = R(x_t,\, c_t).
\end{equation}

Third, a planning module maps a high-level goal $g$ and the inferred context into a (possibly multi-step) action plan:

\begin{equation}
u_{t:t+k} = \mathcal{P}(g, r_t, c_t).
\end{equation}

An executable action is then selected from the plan via an individual action policy that may incorporate stochasticity:

\begin{equation}
u_t \sim \pi_{\mathrm{act}}(\cdot \mid u_{t:t+k},\, r_t,\, c_t).
\end{equation}

The selected action is executed through external tools, yielding an observation:

\begin{equation}
o_t \sim \mathcal{T}(\cdot \mid u_t).
\end{equation}

Finally, let $h_t = (x_t, u_t, o_t)$ denote the interaction tuple. The memory state is updated by incorporating the new experience:

\begin{equation}
m_{t+1} = \psi(m_t,\, h_t).
\end{equation}

Overall, the agent defines a closed-loop system in which the four capabilities interact dynamically rather than operating in isolation: reasoning guides tool selection, planning coordinates multi-step tool invocations, memory preserves intermediate results and learned strategies to inform future decisions, and tool execution grounds the agent's actions in external environments. Figure~\ref{fig:individual} illustrates this architecture. The resulting individual action policy $\pi_{\mathrm{act}}$ is no longer a monolithic input-output mapping, but is factorized through the composition of functionally specialized modules, enabling interpretable and structured decision-making.

\subsection{Reasoning}
\label{reasoning}

Reasoning constitutes the cognitive foundation of LLM-based agents, underpinning their capacity for decision-making, problem-solving, and coherent inference across complex task scenarios~\cite{wang2024survey,xi2023rise}. Chain-of-Thought (CoT) prompting~\cite{wei2022cot} elicits intermediate reasoning steps through carefully designed prompts and substantially improves performance on complex tasks without modifying model parameters. Building on this insight, a rich body of subsequent research has explored diverse strategies to further strengthen the reasoning ability of LLM-based agents~\cite{huang2023towards,sun2025survey}. In the context of agent systems, the intrinsic reasoning capability embedded within the foundation model is generally treated as a given prior; the central challenge thus shifts to how external mechanisms can bridge the gap between a model's static parametric knowledge and the dynamic demands of real-world tasks. Following this perspective, we organize existing methods along the information flow of the reasoning pipeline, categorizing them into three stages: (1)~\textit{input-stage enhancement}, which expands the informational boundary accessible to the model through external knowledge retrieval and multimodal perception; (2)~\textit{reasoning-process enhancement}, which optimizes the reasoning trajectory via search space expansion and path verification; and (3)~\textit{output-stage regulation}, which improves the reliability and factual accuracy of generated results through post-hoc evaluation and adaptive correction.

\subsubsection{Input-Stage Enhancement}
\label{input}
Input-stage enhancement expands the information available to the model before reasoning begins, shifting from sole reliance on parametric knowledge toward dynamic integration of external information. This category encompasses two main directions: knowledge augmentation and modality augmentation.

\paragraph{Knowledge Augmentation}

Retrieval-Augmented Generation (RAG)~\cite{lewis2020retrieval} couples a parametric generator with a non-parametric retrieval module, enabling the model to fetch relevant documents from external corpora at inference time. Self-RAG~\cite{asai2024selfrag} improves upon this by training the model to adaptively decide when to retrieve and to critique both retrieved passages and its own generations through reflection tokens. Beyond the retrieve-then-read paradigm, several alternative strategies have emerged. On the structured knowledge side, SubgraphRAG~\cite{li2025subgraphrag} extracts query-relevant subgraphs from large-scale knowledge graphs, and Think-on-Graph~\cite{sun2024thinkongraph} enables LLMs to interactively explore knowledge graphs through iterative beam search over entity--relation paths, tightly coupling retrieval with multi-hop reasoning. On the knowledge generation side, Knowledge Card~\cite{feng2024knowledgecard} trains domain-specialized small models as modular knowledge repositories that generate background documents on demand, augmenting closed-source LLMs without relying on retrieval infrastructure. RuAG~\cite{zhang2025ruag} takes a compression-oriented approach, distilling external data into compact first-order logic rules via LLM-aided Monte Carlo Tree Search and injecting them directly into prompts.

\paragraph{Modality Augmentation}

Many real-world reasoning tasks involve non-textual evidence that cannot be adequately captured in words alone. Multimodal-CoT~\cite{zhang2024multimodalcot} first demonstrates that fusing visual features into chain-of-thought reasoning yields substantial gains on science QA, even with sub-billion-parameter models. However, it treats visual input as a static feature. Visual Sketchpad~\cite{hu2024sketchpad} lifts this restriction by equipping models with a drawing canvas for creating intermediate visual artifacts during reasoning, and CogCoM~\cite{qi2025cogcom} internalizes step-by-step visual manipulations, such as region grounding and zooming, while producing interpretable reasoning traces.

\subsubsection{Reasoning-Process Enhancement}

Input-stage methods enrich the information available before inference begins; the techniques discussed below instead intervene on the reasoning trajectory itself. We organize them along two dimensions: \emph{search-space expansion}, which increases the coverage of candidate solutions, and \emph{path verification}, which provides the selection pressure needed to identify the most reliable outcome.

\paragraph{Search Space Expansion}

CoT prompting and its zero-shot variant~\cite{wei2022cot,kojima2022zeroshotcot} activate step-by-step decomposition through appropriate prompts. Tree of Thoughts~\cite{yao2023tot} and Graph of Thoughts~\cite{besta2024got} generalize the search topology from a single chain to tree- and graph-structured spaces supporting branching, backtracking, and thought merging. These methods, however, rely on hand-designed prompting protocols, leaving open how to scale reasoning more systematically.

Recent work addresses this from two angles. On the learning side, DeepSeek-R1~\cite{deepseek2025r1} trains a model via large-scale reinforcement learning to produce extended chains of thought autonomously, exhibiting emergent self-verification and spontaneous backtracking that arise purely from reward optimization. CPO~\cite{zhang2024cpo} takes a distillation perspective, collecting step-level preference pairs from tree search and fine-tuning via DPO so that ordinary decoding reproduces the quality of deliberate search at lower cost. On the resource-allocation side, Snell et al.~\cite{snell2024scaling} show that routing easy prompts to lightweight revision and hard prompts to parallel sampling or PRM-guided search enables a smaller model to match a substantially larger one under comparable compute budgets.

\paragraph{Reasoning Path Verification}

Expanding the search space produces richer candidate solutions, yet without effective selection the additional diversity may introduce as many erroneous paths as correct ones. Self-Consistency~\cite{wang2023selfconsistency} addresses this at the output level by sampling multiple chains and returning the majority answer, while Self-Refine~\cite{madaan2023selfrefine} and Reflexion~\cite{shinn2023reflexion} implement iterative critique loops, with Reflexion further maintaining an episodic memory of past failures. These approaches improve reliability without external supervision but evaluate entire outputs rather than pinpointing which step first goes wrong. Process reward models (PRMs) address this limitation by scoring each reasoning step individually. Lightman et al.~\cite{lightman2024lets} train a PRM on human-annotated step-level labels and show that step-wise supervision substantially outperforms outcome-level reward models. Math-Shepherd~\cite{wang2024mathshepherd} eliminates the annotation bottleneck by estimating step correctness through Monte Carlo rollouts, matching human-supervised PRMs while scaling to larger training sets. WizardMath~\cite{luo2025wizardmath} further integrates PRMs into both PPO training and inference-time verification, confirming that process supervision is most effective when it shapes both the learned policy and the selection mechanism.

\subsubsection{Output-Stage Regulation}

The strategies above improve reasoning quality from the perspectives of information supply and search mechanisms. However, model outputs may still contain factual errors---commonly termed hallucination~\cite{ji2023hallucination}. Output-stage regulation addresses this risk along two complementary dimensions: \emph{output reliability assessment}, which evaluates the factual consistency and trustworthiness of generated content, and \emph{response behavior regulation}, which dynamically adjusts the model's generation strategy based on reliability signals to keep error risk within acceptable bounds.

\paragraph{Output Reliability Assessment}

Reliability assessment aims to evaluate the factual trustworthiness of generated content. Evidence-based methods verify claims against external knowledge: FActScore~\cite{min2023factscore} decomposes long-form text into atomic claims and checks each against a knowledge base, while FacTool~\cite{chern2023factool} and Factcheck-GPT~\cite{wang2024factcheckgpt} extend this pipeline to code generation, mathematical reasoning, and scientific literature with finer-grained error localization. An alternative line of work detects hallucination without external knowledge. SelfCheckGPT~\cite{manakul2023selfcheckgpt} detects hallucination by measuring agreement across multiple independent samples without any external reference, and Semantic Entropy~\cite{kuhn2023semantic} refines this by clustering semantically equivalent outputs before estimating uncertainty, capturing the model's true confidence at the meaning level rather than the surface form.

Deeper approaches extract reliability signals directly from model internals. HaloScope~\cite{du2024haloscope} uses membership estimation scores from hidden states to distinguish factual from hallucinated generations in an unsupervised manner, while HalluLens~\cite{bang2025hallulens} provides a systematic taxonomy distinguishing extrinsic from intrinsic hallucinations. Chain-of-Verification (CoVe)~\cite{dhuliawala2024cove} takes an active approach: after generating an initial answer, the model automatically produces verification questions, independently answers them, and revises inconsistencies; RARR~\cite{gao2023rarr} follows a similar verify-then-revise pattern using retrieved external evidence.

\paragraph{Response Behavior Regulation}

Reliability assessment provides error-identification signals; response behavior regulation acts on these signals to constrain the model's generation. Existing mechanisms divide into inference-time decoding adjustments and training-time alignment.

At the decoding level, contrastive decoding suppresses hallucination by amplifying differences between distinct signal sources. DoLa~\cite{chuang2024dola} contrasts logit distributions from higher versus lower Transformer layers to surface factual knowledge encoded in deeper representations. CAD~\cite{shi2024cad} contrasts output distributions with and without context to strengthen reliance on retrieved evidence, and ITI~\cite{li2024iti} directly shifts attention-head activations along pre-identified truthfulness directions during inference. More recent extensions include SADI~\cite{sadi2025iclr}, which introduces semantically aligned dynamic intervention, and MCD~\cite{mcd2025neurips}, which extends contrastive decoding to a three-model architecture for tighter hallucination suppression.

At the training level, FactTune~\cite{tian2024facttune} constructs factuality preference pairs from automated evaluation scores and fine-tunes via DPO, requiring no human annotation. For queries exceeding the model's knowledge boundaries, R-Tuning~\cite{zhang2024rtuning} explicitly teaches the model to output refusal signals on out-of-scope samples, and calibrated RL approaches~\cite{wu2025calibratedrl} incentivize stochastic abstention when confidence is insufficient. CRITIC~\cite{gou2024critic} takes a tool-interactive approach, having the model verify its own outputs against external tools such as calculators and search engines before committing to a final answer. SEAL~\cite{zweiger2025seal} trains models to embed calibrated confidence directly into their natural-language responses, enabling downstream systems to assess reliability without probing internal states.

\begin{table}[!htbp]
\centering
\caption{Comparison of reasoning enhancement methods (Section~\ref{reasoning}), organized by intervention stage and sub-type.}
\label{tab:section3-methods}
\footnotesize
\renewcommand{\arraystretch}{1.05}
\setlength{\tabcolsep}{3.1pt}
\begin{adjustbox}{max width=\textwidth}
\begin{tabular}{@{}L{2.75cm} C{1.05cm} C{1.82cm} C{0.95cm} C{1.20cm} C{0.78cm} C{0.58cm} C{0.58cm} C{0.68cm} C{0.50cm}@{}}
\toprule
\textbf{Method} & \textbf{Sub-type} & \textbf{Task Scope} & \makecell{\textbf{Prim.}\\\textbf{Source}} & \textbf{Correction} & \makecell{\textbf{Ext.}\\\textbf{Ret.}} & \textbf{Tool} & \textbf{TF} & \textbf{TTC} & \textbf{Year} \\
\midrule
\rowcolor{gray!12}
\multicolumn{10}{c}{\rowstrut\small\itshape A.\ Input-Stage Enhancement} \\
\midrule
\rowstrut ViperGPT~\cite{suris2023vipergpt}       & \PE   & \MU               & Inter  & \na    & \na    & \cmark & \cmark & \na    & \ygray{2023} \\
\rowcolor{zebragray}
\rowstrut RAPTOR~\cite{sarthi2024raptor}          & \KE   & \QA\ \LF          & Evid   & \na    & \cmark & \na    & \cmark & \na    & \ygray{2024} \\
\rowstrut Self-RAG~\cite{asai2024selfrag}         & \KE   & \QA\ \LF          & Evid   & SF     & \cmark & \na    & \na    & \na    & \ygray{2024} \\
\rowcolor{zebragray}
\rowstrut LLaVA~\cite{liu2023llava}               & \PE   & \MU               & Inter  & \na    & \na    & \cmark & \na    & \na    & \ygray{2024} \\
\rowstrut TopViewRS~\cite{li2024topviewrs}        & \PE   & \MU               & Inter  & \na    & \na    & \na    & \cmark & \na    & \ygray{2024} \\
\rowcolor{zebragray}
\rowstrut VisCoT~\cite{shao2024viscot}            & \PE   & \MU               & Inter  & \na    & \na    & \na    & \cmark & \na    & \ygray{2024} \\
\rowstrut HippoRAG2~\cite{gutierrez2025hipporag2} & \KE   & \QA\ \LF          & Evid   & \na    & \cmark & \na    & \cmark & \na    & \ygray{2025} \\
\rowcolor{zebragray}
\rowstrut Theanine~\cite{pneg2025timeline}         & \KE   & \QA\ \LF          & Evid   & \na    & \cmark & \na    & \cmark & \na    & \ygray{2025} \\
\rowstrut SubgraphRAG~\cite{li2025subgraphrag}    & \KE   & \QA               & Evid   & \na    & \cmark & \na    & \cmark & \na    & \ygray{2025} \\
\midrule
\rowcolor{gray!12}
\multicolumn{10}{c}{\rowstrut\small\itshape B.\ Reasoning-Process Enhancement} \\
\midrule
\rowstrut Self-Consistency~\cite{wang2023selfconsistency} & \VRP & \MA\ \QA     & Consis & SC     & \na    & \na    & \cmark & \cmark & \ygray{2023} \\
\rowcolor{zebragray}
\rowstrut LATS~\cite{zhou2024lats}                & \ESS  & \MU               & Inter  & Refl   & \na    & \cmark & \cmark & \cmark & \ygray{2024} \\
\rowstrut CoC~\cite{li2024chainofcode}            & \ESS  & \MA\ \QA\ \CO     & Tool   & \na    & \na    & \cmark & \cmark & \na    & \ygray{2024} \\
\rowcolor{zebragray}
\rowstrut AnyTool~\cite{du2024anytool}            & \ESS  & \AG               & Tool   & SF     & \na    & \cmark & \cmark & \na    & \ygray{2024} \\
\rowstrut RAIN~\cite{li2024rain}                  & \ESS  & \QA               & Inter  & SF     & \na    & \na    & \cmark & \cmark & \ygray{2024} \\
\rowcolor{zebragray}
\rowstrut Math-Shepherd~\cite{wang2024mathshepherd} & \VRP & \MA            & Log    & \na    & \na    & \na    & \na    & \cmark & \ygray{2024} \\
\rowstrut CPO~\cite{zhang2024cpo}                 & \VRP  & \MA               & Log    & \na    & \na    & \na    & \na    & \na    & \ygray{2024} \\
\rowcolor{zebragray}
\rowstrut rStar~\cite{qi2025rstar}                & \ESS  & \MA               & Consis & SC     & \na    & \na    & \cmark & \cmark & \ygray{2025} \\
\rowstrut rStar-Math~\cite{guan2025rstarmath}     & \ESS  & \MA               & Consis & SC     & \na    & \na    & \na    & \cmark & \ygray{2025} \\
\rowcolor{zebragray}
\rowstrut TTS~\cite{snell2025tts}                 & \ESS  & \MA               & Log    & \na    & \na    & \na    & \cmark & \cmark & \ygray{2025} \\
\rowstrut TOPS~\cite{yang2025tops}                & \ESS  & \MA               & Log    & \na    & \na    & \na    & \cmark & \cmark & \ygray{2025} \\
\rowcolor{zebragray}
\rowstrut Rewarding Progress~\cite{setlur2025rewarding} & \VRP & \MA        & Log    & \na    & \na    & \na    & \na    & \cmark & \ygray{2025} \\
\rowstrut WizardMath~\cite{luo2025wizardmath}     & \VRP  & \MA               & Log    & \na    & \na    & \na    & \na    & \cmark & \ygray{2025} \\
\midrule
\rowcolor{gray!12}
\multicolumn{10}{c}{\rowstrut\small\itshape C.\ Output-Stage Regulation} \\
\midrule
\rowstrut FActScore~\cite{min2023factscore}       & \OV   & \LF               & Evid   & \na    & \cmark & \na    & \cmark & \na    & \ygray{2023} \\
\rowcolor{zebragray}
\rowstrut FacTool~\cite{chern2023factool}         & \OV   & \MU               & Evid   & \na    & \cmark & \cmark & \cmark & \na    & \ygray{2023} \\
\rowstrut SelfCheckGPT~\cite{manakul2023selfcheckgpt} & \OV & \LF          & Consis & \na    & \na    & \na    & \cmark & \na    & \ygray{2023} \\
\rowcolor{zebragray}
\rowstrut Semantic Entropy~\cite{farquhar2024semantic} & \OV & \QA\ \LF    & Inter  & \na    & \na    & \na    & \cmark & \na    & \ygray{2023} \\
\rowstrut RARR~\cite{gao2023rarr}                 & \OV   & \QA\ \LF          & Evid   & TV     & \cmark & \na    & \cmark & \na    & \ygray{2023} \\
\rowcolor{zebragray}
\rowstrut ITI~\cite{li2024iti}                    & \RC   & \QA               & Inter  & \na    & \na    & \na    & \cmark & \na    & \ygray{2023} \\
\rowstrut Reflexion~\cite{shinn2023reflexion}     & \RC   & \AG\ \CO          & Inter  & Refl   & \na    & \na    & \cmark & \cmark & \ygray{2023} \\
\rowcolor{zebragray}
\rowstrut L2R~\cite{cao2023l2r}                   & \RC   & \QA               & Inter  & \na    & \na    & \na    & \na    & \na    & \ygray{2023} \\
\rowstrut Factcheck-GPT~\cite{wang2024factcheckgpt} & \OV & \QA\ \LF       & Evid   & \na    & \cmark & \na    & \cmark & \na    & \ygray{2024} \\
\rowcolor{zebragray}
\rowstrut HaloScope~\cite{du2024haloscope}        & \OV   & \QA\ \LF          & Inter  & \na    & \na    & \na    & \cmark & \na    & \ygray{2024} \\
\rowstrut DoLa~\cite{chuang2024dola}              & \RC   & \QA\ \LF          & Inter  & \na    & \na    & \na    & \cmark & \na    & \ygray{2024} \\
\rowcolor{zebragray}
\rowstrut CAD~\cite{shi2024cad}                   & \RC   & \QA\ \LF          & Evid   & \na    & \na    & \na    & \cmark & \na    & \ygray{2024} \\
\rowstrut FactTune~\cite{tian2024facttune}        & \RC   & \QA\ \LF          & Evid   & \na    & \na    & \na    & \na    & \na    & \ygray{2024} \\
\rowcolor{zebragray}
\rowstrut CoVe~\cite{dhuliawala2024cove}          & \RC   & \QA\ \LF          & Log    & SF     & \na    & \na    & \cmark & \cmark & \ygray{2024} \\
\rowstrut CRITIC~\cite{gou2024critic}             & \RC   & \QA\ \MA\ \CO     & Tool   & TV     & \na    & \cmark & \cmark & \na    & \ygray{2024} \\
\rowcolor{zebragray}
\rowstrut R-Tuning~\cite{zhang2024rtuning}        & \RC   & \QA               & Inter  & \na    & \na    & \na    & \na    & \na    & \ygray{2024} \\
\rowstrut HalluLens~\cite{bang2025hallulens}      & \OV   & \MU               & Inter  & \na    & \na    & \na    & \cmark & \na    & \ygray{2025} \\
\rowcolor{zebragray}
\rowstrut SADI~\cite{sadi2025iclr}                & \RC   & \QA               & Inter  & \na    & \na    & \na    & \cmark & \na    & \ygray{2025} \\
\rowstrut MCD~\cite{mcd2025neurips}               & \RC   & \QA\ \LF          & Inter  & \na    & \na    & \na    & \cmark & \na    & \ygray{2025} \\
\rowcolor{zebragray}
\rowstrut SEAL~\cite{zweiger2025seal}             & \RC   & \QA\ \LF          & Inter  & \na    & \na    & \na    & \na    & \na    & \ygray{2025} \\
\rowstrut SMC steering~\cite{grand2025smc}        & \RC   & \QA\ \CO          & Log    & \na    & \na    & \na    & \cmark & \na    & \ygray{2025} \\
\bottomrule
\end{tabular}
\end{adjustbox}
\vspace{3pt}
\parbox{\textwidth}{\footnotesize
\textit{Sub-type}: \KE\ Knowledge Enhancement; \PE\ Perceptual Enhancement; \ESS\ Expanding Search Space; \VRP\ Verification of Reasoning Paths; \OV\ Output Verification; \RC\ Response Correction. \;
\textit{Task Scope}: evaluation domains --- \MA\ Math; \QA\ Question Answering; \CO\ Code; \LF\ Long-form Generation; \AG\ Agent; \MU\ Multi-task. \;
\textit{Prim.\ Source}: primary information source --- Evid = external evidence; Inter = internal states; Consis = cross-sample consistency; Log = logic/verification; Tool = tool-based. \;
\textit{Correction}: error-correction mechanism --- SF = self-feedback; Refl = reflection; TV = tool-verify; SC = self-consistency. \;
\textit{Ext.\ Ret.}: uses external retrieval. \;
\textit{TF}: training-free. \;
\textit{TTC}: supports test-time compute scaling. 
}
\end{table}

\subsection{Memory}
\label{memory}

Reasoning, as formalized above, is inherently episodic. However, complex real-world tasks often span multiple episodes, requiring the agent to leverage lessons from prior interactions to improve future performance. To address this need, recent research has introduced explicit memory mechanisms that enable agents to record, consolidate, and recall their own task-execution experiences~\cite{zhang2024memorymechanism,wu2025humantoai}. In the agent's cognitive flow, reasoning generates the raw material of experience, memory consolidates and organizes that material, and the consolidated memory in turn informs the planning and tool-use strategies discussed in subsequent sections. We specifically focus on \emph{agent-level experiential memory}, rather than model-level parametric knowledge or static external knowledge bases already discussed under retrieval-augmented generation (Section~\ref{input}). Under this scoping, memory serves as the substrate for a closed loop: interaction produces experience, experience is consolidated into memory, and memory is retrieved to inform better actions in subsequent interactions. This loop not only strengthens the individual agent's reasoning and planning but also underpins the multi-agent coordination, attribution, and evolution discussed in later sections.
 
This closed-loop perspective naturally suggests three research questions that correspond to the three stages of the memory lifecycle: how raw interaction traces are transformed into storable entries (\textit{memory formation}), how accumulated memories are structured, updated, and selectively forgotten over time (\textit{memory maintenance}), and how relevant memories are recalled and converted into actionable context at decision time (\textit{memory retrieval and utilization}).

\subsubsection{Memory Formation}
\label{sec:mem-formation}

Memory formation concerns how an agent transforms its stream of interactions into discrete, storable memory entries. Existing approaches differ primarily in the degree of abstraction applied during this transformation, and can be grouped into three categories: direct recording, abstractive distillation, and typed routing.

\paragraph{Direct Recording}

Direct recording retains raw interaction traces with minimal processing. Generative Agents~\cite{park2023generative} introduces the \emph{memory stream}, a timestamped log of all observations the agent perceives, serving as the foundational data layer for higher-level memory construction. MemGPT~\cite{packer2024memgpt} frames this process through an operating-system metaphor, writing interaction segments to persistent archival storage via self-directed function calls.

\paragraph{Abstractive Distillation}

Abstractive distillation compresses raw experience into more compact representations, spanning a spectrum of granularity. At the coarsest level, the \emph{reflection} mechanism of Generative Agents~\cite{park2023generative} periodically synthesizes recent observations into higher-level insights that are stored alongside raw entries. MemoChat~\cite{lu2023memochat} fine-tunes LLMs to compose structured memos through iterative memorization--retrieval--response cycles. Recursive summarization~\cite{wang2025recursum} maintains a continuously updated global summary through hierarchical compression of session-level memories. At a finer granularity, Mem0~\cite{chhikara2025mem0} extracts individual facts from each exchange via a two-phase extraction-and-update pipeline. SeCom~\cite{pan2025secom} demonstrates that the segmentation granularity itself critically affects quality, showing that topically coherent segments with compression-based denoising outperform both turn-level and session-level alternatives. 

\paragraph{Typed Routing}

Typed routing goes beyond deciding how much to abstract and additionally determines where each entry should be stored. This is a natural extension of abstractive distillation: once experience is processed, routing it into semantically distinct stores enables more targeted downstream retrieval. ENGRAM~\cite{patel2025engram} classifies each turn via an LLM-based router into episodic, semantic, or procedural stores with normalized schemas; ablation experiments confirm that collapsing types into a single store substantially degrades retrieval quality. CoALA~\cite{sumers2024cognitive} provides the theoretical grounding for such designs, formally defining working, episodic, semantic, and procedural memory as distinct components of a language agent's cognitive architecture. 

Notably, the type of memory formed at this stage directly determines what downstream capabilities can draw upon: episodic traces support reasoning-time self-correction (Section~\ref{reasoning}), semantic knowledge informs experience-driven planning (Section~\ref{planning}), and procedural entries enable cross-task skill reuse (Section~\ref{planning}).

\subsubsection{Memory Maintenance}
\label{sec:mem-maintenance}

As interactions accumulate, the agent's memory grows without bound, yet the capacity to attend to stored content at any given moment remains limited. Memory maintenance addresses this tension through two complementary aspects: the static structure that organizes stored entries, and the dynamic mechanisms that keep the store current and manageable over time.

\paragraph{Storage Structure}

The choice of storage structure determines the representational capacity and retrieval affordances of the memory system. The simplest approach is a flat, append-only list, as exemplified by the memory stream of Generative Agents~\cite{park2023generative}. Flat storage, however, provides no mechanism for relating or aggregating entries, limiting its scalability.

Graph-based structures introduce relational organization. Zep~\cite{rasmussen2025zep} builds a temporally-aware knowledge graph with three hierarchical sub-graphs (episode, semantic entity, and community layers) and a bitemporal data model that tracks both event time and ingestion time. HippoRAG~\cite{gutierrez2024hipporag} draws on hippocampal indexing theory, constructing a schemaless knowledge graph via open information extraction and retrieving through Personalized PageRank. Editable Memory Graph~\cite{kang2024emg} supports dynamic memory modification through explicit graph editing operations for personalized agent behavior.

Hierarchical structures offer multi-granularity abstraction. MemTree~\cite{aadhithya2025memtree} organizes memories as a dynamic tree with varying abstraction levels across depths, routing new information to the most semantically similar branch. A-MEM~\cite{xu2025amem} adopts a Zettelkasten-inspired note-based architecture with LLM-generated metadata and dynamically constructed inter-note links that evolve as new knowledge accretes.

\paragraph{Dynamic Maintenance}

Beyond static organization, long-lived agents require mechanisms to update, compress, or discard stored memories as facts change and relevance decays. Existing approaches span a spectrum from passive decay to active policy learning.

Decay-based methods model forgetting as a continuous process. MemoryBank~\cite{zhong2024memorybank} applies the Ebbinghaus forgetting curve, exponentially decaying each entry's strength since last access and removing entries below a threshold. Timeline-based memory management~\cite{pneg2025timeline} organizes dialogue history along a temporal axis for selective retention based on narrative relevance.

LLM-driven policies make maintenance decisions through explicit reasoning. Mem0~\cite{chhikara2025mem0} uses an LLM to choose among four atomic operations (add, update, delete, or no-op) for each incoming fact. A-MEM~\cite{xu2025amem} implements memory evolution, retroactively refining existing notes when new experience arrives. COMEDY~\cite{chen2025comedy} progressively compresses older interaction segments while preserving key factual content. MemOS~\cite{gao2025memos} elevates memory to a first-class operating-system resource with unified lifecycle management primitives across multiple memory types.

Learned policies optimize maintenance through training signals. MemRL~\cite{yan2026memrl} trains a Q-function that estimates the future utility of each stored experience, retaining high-utility entries and discarding low-utility ones. MemGPT~\cite{packer2024memgpt} takes a resource-management approach, evicting the least relevant segments from main context to archival storage when the context window approaches its limit, analogous to page swapping in virtual memory systems.

\subsubsection{Memory Retrieval and Utilization}
\label{sec:mem-retrieval}

Stored memories become useful only when recalled at the right moment and in the right form. Memory retrieval and utilization closes the experiential loop by converting accumulated state back into actionable context.

\paragraph{Retrieval Mechanisms}

The most widely adopted baseline is dense vector retrieval, where memory entries and the current query are embedded and ranked by cosine similarity~\cite{chhikara2025mem0,zhong2024memorybank,packer2024memgpt}. This approach treats all entries as interchangeable and ignores temporal or importance signals.

Multi-factor scoring addresses this limitation by combining orthogonal dimensions. Generative Agents~\cite{park2023generative} weight recency, LLM-assigned importance, and embedding relevance into a composite score, a design that remains influential across subsequent systems. Type-aware retrieval further exploits the structure introduced during formation: ENGRAM~\cite{patel2025engram} retrieves top-$k$ neighbors independently from each typed store before merging results, ensuring balanced representation across evidence modalities. Zep~\cite{rasmussen2025zep} combines semantic search, BM25 keyword matching, and graph traversal, routing queries through the most effective path for each type. Utility-driven retrieval moves beyond relevance to task-level value: MemRL~\cite{yan2026memrl} first filters candidates by similarity, then re-ranks by a learned Q-value estimating each memory's contribution to task success.

\paragraph{Integration Patterns}
How retrieved memories enter the agent's decision process varies substantially across systems. The simplest approach serializes relevant entries directly into the prompt for the LLM to consume via in-context learning~\cite{chhikara2025mem0,zhong2024memorybank}, requiring no architectural modification but leaving the agent with no control over what is injected or when. A more active pattern uses retrieved memories to shape the agent's reasoning and planning trajectories. Generative Agents~\cite{park2023generative} feed retrieved reflections into a planning module to synthesize proactive daily schedules, so that past experience directly influences the structure of future behavior rather than merely providing supplementary context. Agent~S~\cite{agashe2025agents} similarly retrieves relevant past task completions and uses them to constrain its plan search space, channeling exploration toward previously effective strategies. At the far end of this spectrum, Voyager~\cite{wang2023voyager} retrieves verified code-based skills from a growing library and composes them to solve new tasks, effectively treating memory as a reusable action repertoire that bypasses deliberative planning entirely.


\begin{table}[!htbp]
\centering
\caption{Comparison of memory enhancement methods (Section~\ref{memory}), organized along the memory lifecycle.}
\label{tab:memory-methods}
\footnotesize
\renewcommand{\arraystretch}{1.05}
\setlength{\tabcolsep}{4.5pt}
\begin{adjustbox}{max width=1.05\textwidth}
\begin{tabular}{@{}L{2.6cm} C{1.65cm} C{1.35cm} C{1.55cm} C{1.65cm} C{1.0cm} C{1.55cm} C{0.7cm}@{}}
\toprule
\textbf{Method} & \textbf{Formation} & \textbf{Storage} & \textbf{Maintenance} & \textbf{Retrieval} & \makecell{\textbf{Mem.}\\\textbf{Type}} & \makecell{\textbf{Task}\\\textbf{Domain}} & \textbf{Year} \\
\midrule
\rowstrut Generative Agents~\cite{park2023generative} & Record, Distill & \Flat    & Append      & Multi-factor    & \memES    & Simulation   & \ygray{2023} \\
\rowcolor{zebragray}
\rowstrut MemoChat~\cite{lu2023memochat}              & Distill         & \Flat    & Append      & Similarity      & \memES    & Conversation & \ygray{2023} \\
\rowstrut Reflexion~\cite{shinn2023reflexion}         & Distill         & \Flat    & Append      & Injection       & \memEonly & General      & \ygray{2023} \\
\rowcolor{zebragray}
\rowstrut MemoryBank~\cite{zhong2024memorybank}       & Record          & \Flat    & Decay       & Similarity      & \memES    & Conversation & \ygray{2024} \\
\rowstrut MemGPT~\cite{packer2024memgpt}              & Record          & \Tiered  & Eviction    & Autonomous      & \memEonly & Conv, QA     & \ygray{2024} \\
\rowcolor{zebragray}
\rowstrut HippoRAG~\cite{gutierrez2024hipporag}       & Distill         & \Graph   & Incremental & Graph trav.     & \memSonly & QA           & \ygray{2024} \\
\rowstrut EMG~\cite{kang2024emg}                      & Distill         & \Graph   & Rewrite     & Similarity      & \memES    & Conversation & \ygray{2024} \\
\rowcolor{zebragray}
\rowstrut SeCom~\cite{pan2025secom}                   & Distill         & \Flat    & Append      & Similarity      & \memES    & Conversation & \ygray{2025} \\
\rowstrut ENGRAM~\cite{patel2025engram}               & Routing         & \Typed   & Append      & Type-aware      & \memESP   & Conversation & \ygray{2025} \\
\rowcolor{zebragray}
\rowstrut Mem0~\cite{chhikara2025mem0}                & Distill         & \Graph   & CRUD        & Similarity      & \memSonly & Conversation & \ygray{2025} \\
\rowstrut MemTree~\cite{aadhithya2025memtree}         & Distill         & \TreeS   & Compression & Tree trav.      & \memSonly & QA           & \ygray{2025} \\
\rowcolor{zebragray}
\rowstrut A-MEM~\cite{xu2025amem}                     & Distill         & \Graph   & Evolution   & Similarity      & \memSonly & Conversation & \ygray{2025} \\
\rowstrut Zep~\cite{rasmussen2025zep}                 & Distill         & \Graph   & Versioning  & Hybrid          & \memES    & Conv, Ent    & \ygray{2025} \\
\rowcolor{zebragray}
\rowstrut RecurSum~\cite{wang2025recursum}            & Distill         & \Flat    & Compression & Similarity      & \memSonly & Conversation & \ygray{2025} \\
\rowstrut COMEDY~\cite{chen2025comedy}                & Distill         & \Flat    & Compression & Similarity      & \memES    & Conversation & \ygray{2025} \\
\rowcolor{zebragray}
\rowstrut MemOS~\cite{gao2025memos}                   & Routing         & \Unified & Lifecycle   & Routed      & \memESP   & General      & \ygray{2025} \\
\rowstrut MemRL~\cite{yan2026memrl}                   & Record          & \Flat    & Learned     & Utility-driven  & \memEonly & General      & \ygray{2026} \\
\bottomrule
\end{tabular}
\end{adjustbox}
\vspace{3pt}
\parbox{\textwidth}{\footnotesize
\textit{Formation}: how raw experience becomes memory --- Record = raw logging; Distill = abstractive summarization; Routing = typed classification. \;
\textit{Storage}: data structure holding memories --- \Flat\ flat store; \Tiered\ main context + archival; \Graph\ graph-structured; \TreeS\ tree-structured; \Typed\ type-separated stores; \Unified\ multi-type unified store. \;
\textit{Maintenance}: how memories are updated or discarded --- Decay = time-based forgetting; Eviction = budget-constrained removal; CRUD = create/read/update/delete; Evolution = retroactive refinement; Versioning = bitemporal tracking; Lifecycle = unified management; Learned = RL-based selection. \;
\textit{Retrieval}: how relevant entries are recalled --- Multi-factor = recency + importance + relevance; Injection = full buffer injection; Autonomous = self-directed via function calls; Graph/Tree trav.\ = traversal over corresponding structure; Type-aware = per-type top-$k$ then merge; Hybrid = semantic + keyword + graph; Routed = multi-route retrieval; Utility-driven = similarity + Q-value re-ranking. \;
\textit{Mem.\ Type}: E (Episodic), S (Semantic), P (Procedural); filled = supported, hollow = absent. \;
\textit{Task Domain}: Conv = Conversation; Ent = Enterprise.
}
\end{table}

\subsection{Planning}
\label{planning}

Planning refers to the ability of an agent to convert a high-level goal into a structured sequence of executable actions. Unlike single-step reasoning, planning must account for sub-task dependencies, environmental feedback, and potential execution failures over extended horizons~\cite{wei2025plangenllms,xie2024travelplanner}. We organize existing methods along two paradigms: \textit{decomposition-based planning}, which structures goals into sub-tasks prior to or during execution, and \textit{search-based planning}, which explores alternative plans through systematic evaluation and backtracking.

\subsubsection{Decomposition-Based Planning}
\label{plan-decomp}

Decomposition-based methods reduce a complex goal into manageable sub-tasks, differing primarily in \emph{when} the decomposition occurs relative to execution.

\paragraph{Proactive Decomposition}

Proactive decomposition generates a complete plan before any action is taken. Least-to-Most prompting~\cite{zhou2023leasttomost} introduces the idea of progressively breaking a problem into simpler sub-problems and solving them in sequence, while Plan-and-Solve prompting~\cite{wang2023planandsolve} replaces the generic ``think step by step'' instruction with an explicit planning phase followed by sequential execution. HuggingGPT~\cite{shen2023hugginggpt} extends this idea to multi-modal task orchestration: given a user request, an LLM generates a task dependency graph whose nodes are dispatched to specialist models hosted on Hugging Face. LLMCompiler~\cite{kim2024llmcompiler} draws an analogy to classical compilers, producing a directed acyclic graph (DAG) of function calls that are executed in parallel wherever dependencies permit.

When the action space is grounded in physical or simulated environments, proactive decomposition often takes the form of hierarchical task networks. LLM-Planner~\cite{song2023llmplanner} generates few-shot grounded plans for embodied agents, while LLM+P~\cite{liu2023llmp} translates natural-language goals into PDDL specifications and delegates solving to a classical planner, combining the linguistic flexibility of LLMs with the optimality guarantees of symbolic planners. More recent work explores richer plan representations: Planning Anything~\cite{zheng2025planninganything} formulates planning problems as optimization programs that are solved by formal solvers, and Skeleton-of-Thought~\cite{ning2024skeleton} first generates a plan skeleton and then fills in details in parallel to reduce latency. Plan-and-Act~\cite{erdogan2025planandact} specifically targets long-horizon web tasks by separating a high-level plan from low-level action execution, demonstrating that explicit proactive planning significantly outperforms reactive approaches on complex web benchmarks.

\paragraph{Progressive Decomposition}

Progressive decomposition interleaves planning with execution, refining or extending the plan based on intermediate observations. ReAct~\cite{yao2023react} establishes the foundational pattern by alternating reasoning traces with environment actions in a single prompt, enabling the agent to adjust its next step based on the most recent observation. The verbal self-reflection mechanism introduced by Reflexion~\cite{shinn2023reflexion} (discussed in Section~\ref{reasoning}) further augments this loop, allowing agents to generate natural-language critiques of failed episodes and consult them in subsequent attempts.

ADaPT~\cite{prasad2024adapt} introduces recursive decomposition---when the executor fails on a sub-task, a planner module further decomposes it into finer-grained steps, continuing until each step is within the model's execution capability. SelfGoal~\cite{yang2025selfgoal} dynamically constructs and refines a hierarchical goal tree (GoalTree) during interaction, selecting the most relevant sub-goals for the current state and decomposing them further when guidance is insufficiently specific. In open-world settings, Voyager~\cite{wang2023voyager} combines incremental planning with a growing skill library: the agent proposes sub-goals, writes executable code to achieve them, and stores verified skills for future reuse, enabling continual capability expansion in Minecraft. CodeAct~\cite{wang2024codeact} reframes the action space itself as executable code, allowing the agent to express complex, multi-step plans as Python programs that interact with the environment through function calls, naturally supporting iteration and branching. DEPS~\cite{wang2023deps} follows a Describe-Explain-Plan-Select loop in which the agent generates a plan, explains its rationale, and selects among alternatives based on environmental feedback. KnowAgent~\cite{zhu2024knowagent} injects domain-specific action knowledge into the planning loop, constraining the agent's plan space to reduce hallucinated actions and improve plan feasibility.

\subsubsection{Search-Based Planning}
\label{plan-search}

While decomposition-based methods commit to a single plan trajectory (possibly with replanning), search-based methods explicitly explore multiple candidate paths and select among them through evaluation, pruning, or backtracking. The tree search mechanisms introduced by Tree of Thoughts~\cite{yao2024tot} and Language Agent Tree Search~\cite{zhou2024lats} (discussed in Section~\ref{reasoning}) provide the algorithmic foundation for this paradigm; here we focus on methods that apply and extend these ideas specifically to agent planning scenarios.

\paragraph{Step-Level Search}

Step-level search expands and evaluates candidates at each decision point. MCTS has proven particularly effective for LLM-based planning. LLM-MCTS~\cite{zhao2023llmmcts} uses the LLM as a commonsense prior policy to guide MCTS rollouts in large-scale household task planning. PromptAgent~\cite{wang2024promptagent} applies MCTS to search over the prompt space itself, treating prompt optimization as a planning problem with strategic exploration. RAP~\cite{hao2023rap} formulates reasoning as planning with a world model: the LLM simultaneously serves as both the world model (predicting next states) and the reasoning agent, with MCTS balancing exploration and exploitation. More recently, REST-MCTS*~\cite{zhang2024restmcts} integrates process reward models to guide tree search, and AFlow~\cite{zhang2025aflow} uses MCTS at a higher level of abstraction, searching over entire agentic workflows composed of reusable operators rather than individual actions. Beyond tree search, ToolChain*~\cite{zhuang2024toolchain} formulates tool-use planning as an A* search problem over chains of API calls, and DiffuSearch~\cite{li2025diffusearch} proposes implicit search via discrete diffusion modeling, bypassing explicit tree expansion by predicting future states to inform current action selection.

\paragraph{Plan-Level Search}

Rather than searching step by step, plan-level search generates and compares multiple complete or partial plans. This approach is natural when global coherence matters more than local optimality. The multi-plan selection paradigm identified by Huang et al.~\cite{huang2024understanding} represents an early systematization, where the LLM generates several candidate plans and a selection mechanism (voting, scoring, or verification) chooses the best one. PlanRAG~\cite{lee2024planrag} combines plan generation with retrieval-augmented decision making: the agent generates candidate plans, retrieves relevant evidence from a knowledge graph, and re-ranks plans based on retrieved information. Agent~S~\cite{agashe2025agents} employs a hierarchical planning architecture with experience-augmented search, using a manager agent to generate high-level plans and a worker agent to execute individual steps, with an experience library that enables the agent to learn from past task completions.

\begin{table}[!htbp]
\centering
\caption{Comparison of planning enhancement methods (Section~\ref{planning}), organized by planning paradigm.}
\label{tab:planning-methods}
\footnotesize
\renewcommand{\arraystretch}{1.25}
\setlength{\tabcolsep}{6pt}
\begin{adjustbox}{max width=\textwidth}
\begin{tabular}{@{}L{2.85cm} C{1.60cm} C{0.95cm} C{0.95cm} C{0.72cm} C{1.15cm} C{0.52cm}@{}}
\toprule
\textbf{Method} & \textbf{Strategy} & \textbf{Structure} & \makecell{\textbf{Exec}\\\textbf{Mode}} & \textbf{Replan} & \makecell{\textbf{Task}\\\textbf{Domain}} & \textbf{Year} \\
\midrule
\rowcolor{gray!12}
\multicolumn{7}{c}{\rowstrut\small\itshape A.\ Decomposition-Based Planning} \\
\midrule
\rowstrut Least-to-Most~\cite{zhou2023leasttomost}           & \Proactive   & Seq     & Seq   & \na    & General  & \ygray{2023} \\
\rowcolor{zebragray}
\rowstrut Plan-and-Solve~\cite{wang2023planandsolve}         & \Proactive   & Seq     & Seq   & \na    & General  & \ygray{2023} \\
\rowstrut HuggingGPT~\cite{shen2023hugginggpt}               & \Proactive   & DAG     & Hier  & \na    & General  & \ygray{2023} \\
\rowcolor{zebragray}
\rowstrut LLM-Planner~\cite{song2023llmplanner}              & \Proactive   & Seq     & Seq   & \cmark & Robotics & \ygray{2023} \\
\rowstrut LLM+P~\cite{liu2023llmp}                           & \Proactive   & Seq     & Seq   & \na    & Robotics & \ygray{2023} \\
\rowcolor{zebragray}
\rowstrut ReAct~\cite{yao2023react}                          & \Progressive & Seq     & Intlv & \na    & General  & \ygray{2023} \\
\rowstrut DEPS~\cite{wang2023deps}                           & \Progressive & Seq     & Intlv & \cmark & General  & \ygray{2023} \\
\rowcolor{zebragray}
\rowstrut LLMCompiler~\cite{kim2024llmcompiler}              & \Proactive   & DAG     & Hier  & \cmark & Tool     & \ygray{2024} \\
\rowstrut Skeleton-of-Thought~\cite{ning2024skeleton}        & \Proactive   & Tree    & Hier  & \na    & General  & \ygray{2024} \\
\rowcolor{zebragray}
\rowstrut Voyager~\cite{wang2023voyager}                     & \Progressive & Seq     & Intlv & \cmark & General  & \ygray{2024} \\
\rowstrut ADaPT~\cite{prasad2024adapt}                       & \Progressive & Tree    & Intlv & \cmark & General  & \ygray{2024} \\
\rowcolor{zebragray}
\rowstrut CodeAct~\cite{wang2024codeact}                     & \Progressive & Program & Intlv & \cmark & General  & \ygray{2024} \\
\rowstrut KnowAgent~\cite{zhu2024knowagent}                  & \Progressive & Seq     & Intlv & \cmark & General  & \ygray{2024} \\
\rowcolor{zebragray}
\rowstrut Planning Anything~\cite{zheng2025planninganything} & \Proactive   & Program & Seq   & \na    & General  & \ygray{2025} \\
\rowstrut Plan-and-Act~\cite{erdogan2025planandact}          & \Proactive   & Seq     & Hier  & \na    & Web      & \ygray{2025} \\
\rowcolor{zebragray}
\rowstrut SelfGoal~\cite{yang2025selfgoal}                   & \Progressive & Tree    & Intlv & \cmark & General  & \ygray{2025} \\
\midrule
\rowcolor{gray!12}
\multicolumn{7}{c}{\rowstrut\small\itshape B.\ Search-Based Planning} \\
\midrule
\rowstrut LLM-MCTS~\cite{zhao2023llmmcts}                    & \MCTSb       & State   & Seq   & \na    & Robotics & \ygray{2023} \\
\rowcolor{zebragray}
\rowstrut RAP~\cite{hao2023rap}                              & \MCTSb       & State   & Seq   & \na    & General  & \ygray{2023} \\
\rowstrut PromptAgent~\cite{wang2024promptagent}             & \MCTSb       & State   & Seq   & \na    & General  & \ygray{2024} \\
\rowcolor{zebragray}
\rowstrut ToolChain*~\cite{zhuang2024toolchain}              & \AStarb      & Action  & Seq   & \na    & Tool     & \ygray{2024} \\
\rowstrut REST-MCTS*~\cite{zhang2024restmcts}                & \MCTSb       & State   & Seq   & \na    & General  & \ygray{2024} \\
\rowcolor{zebragray}
\rowstrut PlanRAG~\cite{lee2024planrag}                      & \Retrieveb   & Plan    & Seq   & \cmark & QA       & \ygray{2024} \\
\rowstrut AFlow~\cite{zhang2025aflow}                        & \MCTSb       & Plan    & Seq   & \na    & General  & \ygray{2025} \\
\rowcolor{zebragray}
\rowstrut DiffuSearch~\cite{li2025diffusearch}               & \Diffusionb  & State   & Seq   & \na    & General  & \ygray{2025} \\
\rowstrut Agent S~\cite{agashe2025agents}                    & \Greedyb     & Plan    & Hier  & \cmark & Web      & \ygray{2025} \\
\bottomrule
\end{tabular}
\end{adjustbox}
\vspace{3pt}
\parbox{\textwidth}{\footnotesize
\textit{Strategy}: temporal relation between planning and execution (Part~A) or search algorithm (Part~B) --- \Proactive\ full plan before execution; \Progressive\ interleaved plan-execute loop; \MCTSb\ Monte Carlo Tree Search; \AStarb\ heuristic A* search; \Retrieveb\ retrieval-guided; \Diffusionb\ diffusion-based; \Greedyb\ greedy. \;
\textit{Structure}: plan representation (Part~A) or search space (Part~B) --- Seq = sequential; DAG = directed acyclic graph; Tree = tree-structured; Program = programmatic; State/Action/Plan = corresponding search space. \;
\textit{Exec Mode}: Seq = sequential; Hier = hierarchical; Intlv = interleaved with planning.
}
\end{table}

\subsection{Tool Use}
\label{tool-use}

Tool use extends an agent's action space beyond text generation by enabling it to invoke external functions, APIs, code interpreters, web services, and other external systems~\cite{wang2024codeact}. Unlike reasoning or planning, which operate within the model's internal representation, tool use requires the agent to cross the boundary between language and execution---generating structured calls, processing external returns, and integrating results back into the reasoning flow. We organize existing work along three dimensions that capture the lifecycle of this capability: \textit{tool capability acquisition}, which concerns how models learn to use tools; \textit{tool invocation}, which concerns the control-flow patterns through which tools are called during task execution; and \textit{tool generalization}, which concerns how learned tool-use abilities transfer to unseen tools and environments.

\subsubsection{Tool Capability Acquisition}

Tool capability acquisition addresses how a model learns to recognize when a tool is needed, select the appropriate one, generate valid invocation parameters, and incorporate the returned results. Existing approaches can be broadly divided into trajectory-based learning, which internalizes tool-use behavior from demonstration sequences, and protocol-based alignment, which grounds the model's understanding in tool interface descriptions.

\paragraph{Trajectory-Based Learning}

Toolformer~\cite{schick2023toolformer} pioneer self-supervised tool learning by having the model annotate its own training corpus with API calls and retaining only those that reduce perplexity, enabling autonomous acquisition of tool-use behavior without task-specific supervision. ToolLLM~\cite{qin2024toolllm} scales this idea to 16,000+ real-world APIs by using ChatGPT to generate diverse instruction--trajectory pairs and training open-source models via supervised fine-tuning, while APIGen~\cite{liu2024apigen} further improves data quality through a three-stage verification pipeline covering format, execution, and semantic correctness. GPT4Tools~\cite{yang2023gpt4tools} and ToolACE~\cite{liu2025toolace} explore complementary directions in data synthesis: the former uses self-instruction from GPT-4, while the latter introduces tool self-evolution to systematically expand API diversity and dialog complexity.

Beyond static supervision, reinforcement learning enables models to refine tool-use strategies through interaction. ToolPlanner~\cite{wu2024toolplanner} employs a two-stage RL framework with path planning and dual feedback mechanisms for task completion and instruction following. WebRL~\cite{qi2025webrl} applies self-evolving curriculum RL to web agent training, progressively generating harder tasks from failed attempts and using a learned outcome reward model.

\paragraph{Protocol-Based Alignment}

Rather than learning from execution traces, protocol-based methods align the model's representation with tool interface specifications. Gorilla~\cite{patil2024gorilla} introduces Retriever-Aware Training (RAT), which conditions fine-tuning on retrieved API documentation so that the model learns to ground its calls in up-to-date specifications and adapt to documentation changes at test time. ToolkenGPT~\cite{hao2023toolkengpt} takes a representation-level approach, encoding each tool as a special token in the model's vocabulary so that tool invocation becomes a natural extension of next-token prediction without modifying the frozen language model. EasyTool~\cite{yuan2025easytool} addresses the upstream bottleneck of documentation quality by transforming diverse and verbose tool documentation into unified, concise instructions, significantly reducing token consumption while improving tool-use accuracy. At the ecosystem level, the Model Context Protocol (MCP)~\cite{anthropic2024mcp} standardizes agent-tool integration through a universal JSON-RPC interface, replacing per-tool custom connectors with a single interoperable protocol.

\subsubsection{Tool Invocation}

Once tool-use capability has been acquired, the central question shifts to how tools are invoked during task execution. We distinguish three control-flow patterns of increasing complexity: one-shot calling, closed-loop calling, and workflow-based orchestration.

\paragraph{One-Shot Calling}

In the simplest setting, a single user instruction triggers a single tool call whose result directly produces the final response. This pattern underlies the basic function-calling interfaces of commercial APIs~\cite{openai2023functioncalling} and is the primary evaluation target of benchmarks such as BFCL~\cite{patil2025bfcl}, which systematically assesses serial, parallel, and multi-language function calls using Abstract Syntax Tree matching. Toolformer's single-step API insertion~\cite{schick2023toolformer} and Gorilla's syntax-aware call generation~\cite{patil2024gorilla} both operate in this regime, where the model must select the right tool and produce well-formed parameters in a single generation pass.

\paragraph{Closed-Loop Calling}

Many tasks require multiple rounds of tool interaction, where each tool's output informs the next decision. CRITIC~\cite{gou2024critic} exemplifies this pattern: the model generates an initial response, verifies it against external tools such as calculators or search engines, and revises based on the feedback. AnyTool~\cite{du2024anytool} implements a hierarchical, self-reflective closed loop over a large API pool, dynamically selecting and invoking tools across multiple rounds. ReAct~\cite{yao2023react} provides the foundational interleaving of reasoning traces and tool actions that underpins most closed-loop tool-use systems.

\paragraph{Workflow-Based Orchestration}

For complex goals requiring coordinated use of multiple tools, explicit workflow structures govern the invocation order and data flow. LLMCompiler~\cite{kim2024llmcompiler} generates a DAG of function calls with dependency tracking, enabling parallel execution wherever possible. WorkflowLLM~\cite{fan2025workflowllm} constructs a large-scale benchmark of real-world workflows from Apple Shortcuts and fine-tunes models to orchestrate complex multi-tool sequences with nested branches and loops. GPTSwarm~\cite{zhuge2024gptswarm} models the orchestration as an optimizable graph of LLM invocations, while AFlow~\cite{zhang2025aflow} uses MCTS to automatically search over workflow structures composed of reusable operators.

\subsubsection{Tool Generalization}

A practical agent must handle tools it has never seen during training and retrieve appropriate tools from large, evolving libraries. Tool generalization addresses these challenges along two complementary directions: adapting to unseen tools and discovering relevant tools from large collections.

\paragraph{Unseen Tool Generalization}

Generalizing to previously unseen tools requires the model to rely on documentation comprehension rather than memorized call patterns. GenTool~\cite{liu2025gentool} systematically trains for this through two complementary strategies---zero-to-one generalization (handling entirely new tools) and weak-to-strong generalization (transferring from simpler to more complex tools). TOOLVERIFIER~\cite{mekala2024toolverifier} improves generalization by generating contrastive questions that help the model self-verify its tool selection and parameter generation for unfamiliar APIs. Gorilla's retriever-aware architecture~\cite{patil2024gorilla} also contributes here, as its conditioning on retrieved documentation enables adaptation to test-time API changes without retraining. ToolLLM~\cite{qin2024toolllm} demonstrates strong zero-shot generalization on the out-of-distribution APIBench benchmark despite being trained on a disjoint API set.

\begin{table}[!htbp]
\centering
\caption{Comparison of tool-use enhancement methods (Section~\ref{tool-use}).}
\label{tab:tooluse-methods}
\footnotesize
\renewcommand{\arraystretch}{1.14}
\setlength{\tabcolsep}{4.5pt}
\begin{adjustbox}{max width=1.05\textwidth}
\begin{tabular}{@{}L{2.4cm} C{1.1cm} C{1.8cm} C{1.3cm} C{1.85cm} C{1.5cm} C{1.1cm} C{0.8cm}@{}}
\toprule
\textbf{Method} & \textbf{Training} & \makecell{\textbf{Tool}\\\textbf{Interface}} & \textbf{Tool Set} & \makecell{\textbf{Control}\\\textbf{Flow}} & \makecell{\textbf{Gen}\\\textbf{Target}} & \makecell{\textbf{Task}\\\textbf{Domain}} & \textbf{Year} \\
\midrule
\rowstrut Toolformer~\cite{schick2023toolformer} & Self-sup & Func-Call & \ClosedSet & One-shot & In-dist & General & \ygray{2023} \\
\rowcolor{zebragray}
\rowstrut ToolkenGPT~\cite{hao2023toolkengpt} & SFT & Func-Call & \ClosedSet & One-shot & New-Task & General & \ygray{2023} \\
\rowstrut GPT4Tools~\cite{yang2023gpt4tools} & SFT & Func-Call & \ClosedSet & One-shot & New-Task & General & \ygray{2023} \\
\rowcolor{zebragray}
\rowstrut CRITIC~\cite{gou2024critic} & Prompt & Func-Call & \OpenSet & Closed-loop & In-dist & General & \ygray{2024} \\
\rowstrut Gorilla~\cite{patil2024gorilla} & SFT & Func-Call & \DynamicSet & One-shot & New-Tool & Code & \ygray{2024} \\
\rowcolor{zebragray}
\rowstrut ToolLLM~\cite{qin2024toolllm} & SFT & Func-Call & \OpenSet & Closed-loop & New-Tool & General & \ygray{2024} \\
\rowstrut AnyTool~\cite{du2024anytool} & Prompt & Func-Call & \OpenSet & Closed-loop & New-Tool & General & \ygray{2024} \\
\rowcolor{zebragray}
\rowstrut APIGen~\cite{liu2024apigen} & SFT & Func-Call & \OpenSet & One-shot & New-Tool & General & \ygray{2024} \\
\rowstrut ToolPlanner~\cite{wu2024toolplanner} & RL & Func-Call & \OpenSet & Closed-loop & New-Task & General & \ygray{2024} \\
\rowcolor{zebragray}
\rowstrut Re-Invoke~\cite{chen2024reinvoke} & Prompt & Retrieval & \OpenSet & One-shot & New-Tool & General & \ygray{2024} \\
\rowstrut TOOLVERIFIER~\cite{mekala2024toolverifier}& SFT & Func-Call & \DynamicSet & One-shot & New-Tool & General & \ygray{2024} \\
\rowcolor{zebragray}
\rowstrut MetaTool~\cite{huang2024metatool} & Prompt & Func-Call & \OpenSet & One-shot & New-Tool & General & \ygray{2024} \\
\rowstrut ToolACE~\cite{liu2025toolace} & SFT & Func-Call & \DynamicSet & One-shot & New-Tool & General & \ygray{2025} \\
\rowcolor{zebragray}
\rowstrut EasyTool~\cite{yuan2025easytool} & Prompt & Func-Call & \DynamicSet & One-shot & New-Tool & General & \ygray{2025} \\
\rowstrut WebRL~\cite{qi2025webrl} & RL & Web & \OpenSet & Closed-loop & New-Env & Agent & \ygray{2025} \\
\rowcolor{zebragray}
\rowstrut WorkflowLLM~\cite{fan2025workflowllm} & SFT & Func-Call & \OpenSet & Workflow & New-Tool & General & \ygray{2025} \\
\rowstrut GenTool~\cite{liu2025gentool} & SFT & Func-Call & \DynamicSet & One-shot & New-Tool & General & \ygray{2025} \\
\rowcolor{zebragray}
\rowstrut BFCL~\cite{patil2025bfcl} & --- & Func-Call & \DynamicSet & One-shot & --- & General & \ygray{2025} \\
\rowstrut ToolRet~\cite{shi2025toolret} & --- & Retrieval & \OpenSet & --- & New-Tool & General & \ygray{2025} \\
\bottomrule
\end{tabular}
\end{adjustbox}
\vspace{3pt}
\parbox{\textwidth}{\footnotesize
\textit{Training}: learning paradigm --- Self-sup = self-supervised; SFT = supervised fine-tuning; RL = reinforcement learning; Prompt = prompting only (no training). \;
\textit{Tool Interface}: interaction modality --- Func-Call = function-calling; Retrieval = retrieval-based selection; Web = direct web interaction. \;
\textit{Tool Set}: tool availability at test time --- \ClosedSet\ fixed set, known at training; \OpenSet\ extensible to unseen tools; \DynamicSet\ available tools change across tasks. \;
\textit{Control Flow}: invocation pattern --- One-shot = single invocation; Closed-loop = iterative invocation with feedback; Workflow = multi-step orchestration. \;
\textit{Gen Target}: generalization target evaluated --- In-dist = in-distribution only; New-Task = unseen tasks; New-Tool = unseen tools; New-Env = unseen environments.
}
\end{table}

\paragraph{Tool Discovery and Retrieval}

As tool libraries scale to thousands of APIs, identifying the right tool for a given query becomes a retrieval problem in its own right. ToolLLM~\cite{qin2024toolllm} trains a neural API retriever that recommends relevant APIs from a large pool, while Re-Invoke~\cite{chen2024reinvoke} proposes an unsupervised approach that enriches tool documents with synthetic queries and extracts user intents through multi-view similarity ranking, achieving strong zero-shot retrieval without any labeled query--tool pairs. ToolRet~\cite{shi2025toolret} provides the first comprehensive tool retrieval benchmark and reveals that existing information retrieval models, despite strong performance on general retrieval tasks, perform poorly on tool retrieval, highlighting a significant gap between conventional document retrieval and tool-oriented retrieval. MetaTool~\cite{huang2024metatool} addresses the upstream decision of whether to use a tool at all and which tool to select, framing these as meta-cognitive capabilities that precede the invocation itself.

A related line of work abstracts tool discovery to the level of reusable skills. SkillWeaver~\cite{shi2025skillweaver} enables web agents to autonomously synthesize reusable skills as APIs through iterative environment exploration, building transferable skill libraries. SAGE~\cite{wang2025sage} enhances this paradigm with reinforcement learning, using sequential rollouts across similar tasks to improve skill acquisition.

\subsection{Evaluation}

The capabilities reviewed in Sections~\ref{reasoning}--\ref{tool-use} require systematic benchmarking to identify where individual agents succeed and where they fall short.
As summarized in Table~\ref{tab:single-agent-benchmarks}, the evaluation landscape for LLM-based agents has expanded rapidly in recent years, exhibiting two notable trends.
First, benchmarks have moved from testing isolated skills toward joint evaluation of reasoning, planning, and tool use within unified task settings.
Second, evaluation criteria have shifted from purely outcome-based metrics (e.g., success rate or exact match) toward process-level signals such as action trajectories, subgoal completion, and execution traces, which provide finer diagnostic resolution.

Comprehensive benchmarks aim to stress-test multiple capabilities simultaneously.
AgentBench~\cite{liu2023agentbench} is among the first to evaluate agents across eight diverse environments with trajectory-level assessment.
GAIA~\cite{mialon2024gaia} and MINT~\cite{wang2024mint} further combine web interaction and tool use within multi-turn settings.
More recent efforts raise the bar in evaluation granularity: AgentBoard~\cite{ma2024agentboard} introduces subgoal-based progress metrics that capture partial task completion, while TheAgentCompany~\cite{xu2024theagentcompany} situates evaluation in realistic enterprise workflows requiring long-horizon coordination across web and code environments.

Beyond comprehensive suites, a large body of work targets specific execution environments or individual capabilities.
In web environments, WebArena~\cite{zhou2024webarena} constructs self-hosted websites for reproducible long-horizon evaluation, VisualWebArena~\cite{koh2024visualwebarena} extends this to multimodal web tasks, and Mind2Web~\cite{deng2024mind2web} provides large-scale cross-website coverage.
BrowseComp~\cite{wei2025browsecomp} further challenges agents with deeply entangled, hard-to-find information retrieval.
OSWorld~\cite{xie2024osworld} and AndroidWorld~\cite{rawles2024androidworld} bring evaluation to desktop and mobile operating systems, adopting state verification as the primary evaluation signal.
In code and software engineering, SWE-bench~\cite{jimenez2024swebench} has become the de facto standard for assessing agents on real GitHub issue resolution over extended interaction horizons.
For tool use specifically, ToolBench~\cite{qin2023toolllm} offers large-scale API evaluation with over 12,000 instances, BFCL~\cite{patil2025bfcl} introduces AST-level matching for precise assessment of function-call correctness, and AppWorld~\cite{trivedi2024appworld} and $\tau$-bench~\cite{yao2024taubench} embed tool use within richer application ecosystems and dialogue contexts.
Planning benchmarks such as ALFWorld~\cite{shridhar2021alfworld} and ScienceWorld~\cite{wang2022scienceworld} evaluate long-horizon action sequences in text-based worlds.
Memory evaluation remains comparatively nascent: LoCoMo~\cite{maharana2024locomo} tests long-conversation memory through entailment-based metrics, LongMemEval~\cite{wu2025longmemeval} systematically probes multi-session recall, and MemoryAgentBench~\cite{hu2026memoryagentbench} couples memory assessment with application-level interaction.

Recent benchmarks have also expanded into domain-specific and safety-critical settings.
In science and machine learning, MLE-bench~\cite{chan2024mlebench}, CORE-Bench~\cite{siegel2024corebench}, and ScienceAgentBench~\cite{chen2025scienceagentbench} evaluate agents on automated experimentation and code-based research tasks, while PaperBench~\cite{starace2025paperbench} assesses the ability to reproduce research papers using rubric-based subgoal evaluation.
Enterprise benchmarks such as WorkArena++~\cite{drouin2024workarena} and CRMArena~\cite{huang2024crmarena} test agents in realistic business applications, though both remain only partially open.
On the safety front, Cybench~\cite{zhang2025cybench} targets cybersecurity challenges and AgentHarm~\cite{andriushchenko2025agentharm} evaluates susceptibility to harmful tool-use behaviors.
Despite this breadth, several gaps persist.
Cross-capability evaluation that examines how improvements in one module (e.g., memory) affect performance in another (e.g., planning) remains largely absent.
Efficiency metrics such as token consumption, latency, and cost are rarely incorporated into benchmark design.
These gaps suggest that future evaluation efforts should move toward integrated, resource-aware assessment protocols.
Moreover, the capability bottlenecks revealed by single-agent benchmarks provide a direct motivation for the multi-agent collaboration strategies discussed in the next section.

\begin{table}[p]
\centering
\caption{Single-agent capability benchmarks, organized by evaluation domain.}
\label{tab:single-agent-benchmarks}
\renewcommand{\arraystretch}{1.0}
\setlength{\tabcolsep}{4.2pt}
\begin{footnotesize}
\begin{adjustbox}{max width=1.05\textwidth}
\begin{tabular}{@{}L{2.40cm} C{2.20cm} C{1.55cm} C{1.15cm} C{1.35cm} C{1.15cm} C{1.05cm} C{0.48cm} C{0.52cm}@{}}
\toprule
\textbf{Benchmark} & \textbf{Capability} & \textbf{Environment} & \textbf{Interaction} & \makecell{\textbf{Observable}\\\textbf{Signal}} & \makecell{\textbf{Eval.}\\\textbf{Method}} & \textbf{Scale} & \textbf{Open} & \textbf{Year} \\
\midrule
\rowcolor{gray!12}
\multicolumn{9}{c}{\rowstrut\small\itshape A.\ Comprehensive} \\
\midrule
\rowstrut AgentBench         & \Rsec\;\Psec\;\Tsec         & Mixed          & Multi  & ActTraj   & SR       & ${\sim}$1K  & \cmark & \ygray{2023} \\
\rowcolor{zebragray}
\rowstrut GAIA               & \Rpri\;\Psec\;\Tpri         & Web + Tool     & Multi  & Outcome   & EM       & 466         & P      & \ygray{2023} \\
\rowstrut MINT               & \Rpri\;\Psec\;\Tpri         & Mixed          & Multi  & ToolSeq   & SR       & 586         & \cmark & \ygray{2023} \\
\rowcolor{zebragray}
\rowstrut AgentBoard         & \Rsec\;\Ppri\;\Tsec         & Mixed          & Multi  & Subgoal   & Comp     & 1,013       & \cmark & \ygray{2024} \\
\rowstrut TheAgentCompany    & \Rsec\;\Ppri\;\Tpri         & Web + Code     & Long   & ExecTrace & Comp     & 175         & \cmark & \ygray{2024} \\
\midrule
\rowcolor{gray!12}
\multicolumn{9}{c}{\rowstrut\small\itshape B.\ Web} \\
\midrule
\rowstrut WebArena           & \Rsec\;\Ppri\;\Tpri         & Web            & Long   & ActTraj   & Exec     & 812         & \cmark & \ygray{2023} \\
\rowcolor{zebragray}
\rowstrut Mind2Web           & \Rsec\;\Ppri\;\Tsec         & Web            & Multi  & ActTraj   & Comp     & 2,350       & \cmark & \ygray{2023} \\
\rowstrut VisualWebArena     & \Rsec\;\Ppri\;\Tpri         & Web            & Long   & ActTraj   & Exec     & 910         & \cmark & \ygray{2024} \\
\rowcolor{zebragray}
\rowstrut BrowseComp         & \Rpri\;\Psec\;\Tpri         & Web            & Multi  & Outcome   & EM       & 1,266       & \cmark & \ygray{2025} \\
\midrule
\rowcolor{gray!12}
\multicolumn{9}{c}{\rowstrut\small\itshape C.\ Desktop \& Mobile} \\
\midrule
\rowstrut OSWorld            & \Rsec\;\Ppri\;\Tpri         & Desktop        & Long   & StateVer  & Exec     & 369         & \cmark & \ygray{2024} \\
\rowcolor{zebragray}
\rowstrut AndroidWorld       & \Rsec\;\Ppri\;\Tpri         & Mobile         & Long   & StateVer  & SR       & 116         & \cmark & \ygray{2024} \\
\rowstrut Mobile-Bench       & \Rsec\;\Ppri\;\Tsec         & Mobile         & Multi  & ActTraj   & Comp     & 832         & \cmark & \ygray{2024} \\
\midrule
\rowcolor{gray!12}
\multicolumn{9}{c}{\rowstrut\small\itshape D.\ Code \& Software} \\
\midrule
\rowstrut SWE-bench          & \Rsec\;\Ppri\;\Tpri         & Code           & Long   & ExecTrace & Exec     & 2,294       & \cmark & \ygray{2024} \\
\rowcolor{zebragray}
\rowstrut Terminal-Bench     & \Rsec\;\Ppri\;\Tpri         & Terminal       & Long   & ExecTrace & Exec     & 89          & \cmark & \ygray{2025} \\
\midrule
\rowcolor{gray!12}
\multicolumn{9}{c}{\rowstrut\small\itshape E.\ Tool Use} \\
\midrule
\rowstrut ToolBench          & \Rsec\;\Psec\;\Tpri         & API            & Multi  & ToolSeq   & LLM-J    & 12,657      & \cmark & \ygray{2023} \\
\rowcolor{zebragray}
\rowstrut API-Bank           & \Rsec\;\Psec\;\Tpri         & API            & Multi  & ToolSeq   & Comp     & 314         & \cmark & \ygray{2023} \\
\rowstrut T-Eval             & \Rsec\;\Psec\;\Tpri         & API            & Multi  & ToolSeq   & Step Acc & 553         & \cmark & \ygray{2024} \\
\rowcolor{zebragray}
\rowstrut AppWorld           & \Rsec\;\Ppri\;\Tpri         & App            & Long   & StateVer  & Exec     & 750         & \cmark & \ygray{2024} \\
\rowstrut $\tau$-bench       & \Rsec\;\Ppri\;\Tpri         & Dialogue+Tool  & Long   & StateVer  & Exec     & 168         & \cmark & \ygray{2024} \\
\rowcolor{zebragray}
\rowstrut BFCL               & \Rsec\;\Psec\;\Tpri         & API            & Multi  & ToolSeq   & AST+Exec & 2,000+      & \cmark & \ygray{2025} \\
\rowstrut ToolSandbox        & \Rsec\;\Msec\;\Psec\;\Tpri  & Dialogue+Tool  & Long   & Milestone & Comp     & 180         & \cmark & \ygray{2025} \\
\midrule
\rowcolor{gray!12}
\multicolumn{9}{c}{\rowstrut\small\itshape F.\ Planning} \\
\midrule
\rowstrut ALFWorld           & \Rsec\;\Ppri\;\Tsec         & Text World     & Long   & ActTraj   & SR       & 134         & \cmark & \ygray{2021} \\
\rowcolor{zebragray}
\rowstrut ScienceWorld       & \Rpri\;\Msec\;\Ppri         & Text World     & Long   & ActTraj   & SR       & 300         & \cmark & \ygray{2022} \\
\midrule
\rowcolor{gray!12}
\multicolumn{9}{c}{\rowstrut\small\itshape G.\ Memory} \\
\midrule
\rowstrut LoCoMo             & \Rsec\;\Mpri                & Dialogue       & Long   & NLI       & F1+Human & 10 convs    & \cmark & \ygray{2024} \\
\rowcolor{zebragray}
\rowstrut LongMemEval        & \Rsec\;\Mpri                & Dialogue       & Long   & Subgoal   & Acc      & 500         & \cmark & \ygray{2025} \\
\rowstrut MemoryAgentBench   & \Rsec\;\Mpri\;\Psec         & Dialogue+App   & Long   & ActTraj   & Comp     & 1,000+      & P      & \ygray{2026} \\
\midrule
\rowcolor{gray!12}
\multicolumn{9}{c}{\rowstrut\small\itshape H.\ Science \& ML} \\
\midrule
\rowstrut MLE-bench          & \Rpri\;\Ppri\;\Tpri         & Code           & Long   & ExecTrace & Exec     & 75          & \cmark & \ygray{2024} \\
\rowcolor{zebragray}
\rowstrut MLAgentBench       & \Rpri\;\Ppri\;\Tpri         & Code           & Long   & ExecTrace & Exec     & 13          & \cmark & \ygray{2024} \\
\rowstrut CORE-Bench         & \Rsec\;\Ppri\;\Tpri         & Code           & Long   & ExecTrace & Exec     & 270         & \cmark & \ygray{2024} \\
\rowcolor{zebragray}
\rowstrut ScienceAgentBench  & \Rpri\;\Ppri\;\Tpri         & Code           & Long   & ExecTrace & Exec     & 102         & \cmark & \ygray{2025} \\
\rowstrut PaperBench         & \Rpri\;\Ppri\;\Tpri         & Code + Web     & Long   & Subgoal   & Rubric   & 20 papers   & P      & \ygray{2025} \\
\midrule
\rowcolor{gray!12}
\multicolumn{9}{c}{\rowstrut\small\itshape I.\ Enterprise} \\
\midrule
\rowstrut WorkArena++        & \Rsec\;\Ppri\;\Tpri         & Web            & Long   & ActTraj   & SR       & 682         & P      & \ygray{2024} \\
\rowcolor{zebragray}
\rowstrut CRMArena           & \Rsec\;\Psec\;\Tsec         & CRM            & Single & Outcome   & Fuzzy    & 1,186       & P      & \ygray{2024} \\
\midrule
\rowcolor{gray!12}
\multicolumn{9}{c}{\rowstrut\small\itshape J.\ Safety} \\
\midrule
\rowstrut Cybench            & \Rpri\;\Ppri\;\Tpri         & Code+Terminal  & Long   & ExecTrace & SR       & 40          & \cmark & \ygray{2025} \\
\rowcolor{zebragray}
\rowstrut AgentHarm          & \Rsec\;\Psec\;\Tpri         & API            & Multi  & ToolSeq   & SR       & 440         & \cmark & \ygray{2025} \\
\bottomrule
\end{tabular}
\end{adjustbox}
\end{footnotesize}
\vspace{3pt}
\parbox{\textwidth}{\scriptsize
\textit{Capability}: \Rpri\ Reasoning; \Mpri\ Memory; \Ppri\ Planning; \Tpri\ Tool Use. Filled = primary; outlined (e.g.\ \Rsec) = secondary; absent = not covered. \;
\textit{Interaction}: Single = single-turn; Multi = multi-turn; Long = long-horizon. \;
\textit{Signal}: ActTraj = action trajectory; ToolSeq = tool-call sequence; StateVer = state verification; ExecTrace = execution trace; Subgoal = subgoal completion; Milestone = milestone completion; NLI = NLI-based verification; Outcome = final outcome only. \;
\textit{Eval.}: SR = success rate; EM = exact match; Exec = execution-based; AST+Exec = AST matching + execution; Step Acc = step-level accuracy; LLM-J = LLM-as-judge; Comp = composite (multi-metric); F1+Human = F1 + human eval; Fuzzy = fuzzy matching; Rubric = rubric-based; Acc = accuracy. \;
\textit{Open}: \cmark\ = fully open; P = partially open (e.g.\ test set hidden).
}
\end{table}

\subsection{Discussion}

\textbf{Modular optimization dominates, but inter-module coupling remains unstudied.} The formalization in Section~\ref{subsec:overview} shows that reasoning, memory, planning, and tool use are interdependent at every time step, yet existing methods almost exclusively optimize a single module in isolation. Capability mismatches across modules often become the actual bottleneck in end-to-end deployment~\cite{yao2025rethinking,gutierrez2024hipporag,qin2024toolllm}, but such failure modes are invisible to current benchmarks, which predominantly target one or two capabilities (Table~\ref{tab:single-agent-benchmarks}).

\textbf{Reasoning enhancement and hallucination suppression are converging but methodologically separated.} Expanding the search space~\cite{yao2023tot,besta2024got} produces richer candidate paths while simultaneously increasing the surface area for errors. Process reward models~\cite{wang2024mathshepherd} bridge the two concerns but scale poorly beyond mathematical domains. Most reliability assessment remains post-hoc~\cite{min2023factscore,manakul2023selfcheckgpt}, leaving open how to dynamically steer search based on step-level confidence during inference.

\textbf{Memory design is fragmented, with little understanding of how design choices interact.} Table~\ref{tab:memory-methods} shows diverse combinations of formation, storage, maintenance, and retrieval, yet no work systematically studies their interactions. Graph-based storage excels at relational retrieval but incurs high maintenance cost~\cite{rasmussen2025zep}; utility-driven maintenance requires extensive interaction to converge~\cite{yan2026memrl}; and storage formats are typically chosen without regard for downstream retrieval compatibility.

\textbf{Capabilities are evaluated as static snapshots, overlooking continuous adaptation.} All methods in this chapter assume frozen strategies and known tool sets, yet real deployment demands ongoing adaptation. Tool generalization and memory evolution treat adaptation as a one-time capacity rather than a continuous process.

\section{Multi-Agent Collaboration}
\label{sec:Multi-Agent Collaboration}

\begin{figure}
    \centering
    \includegraphics[width=\linewidth]{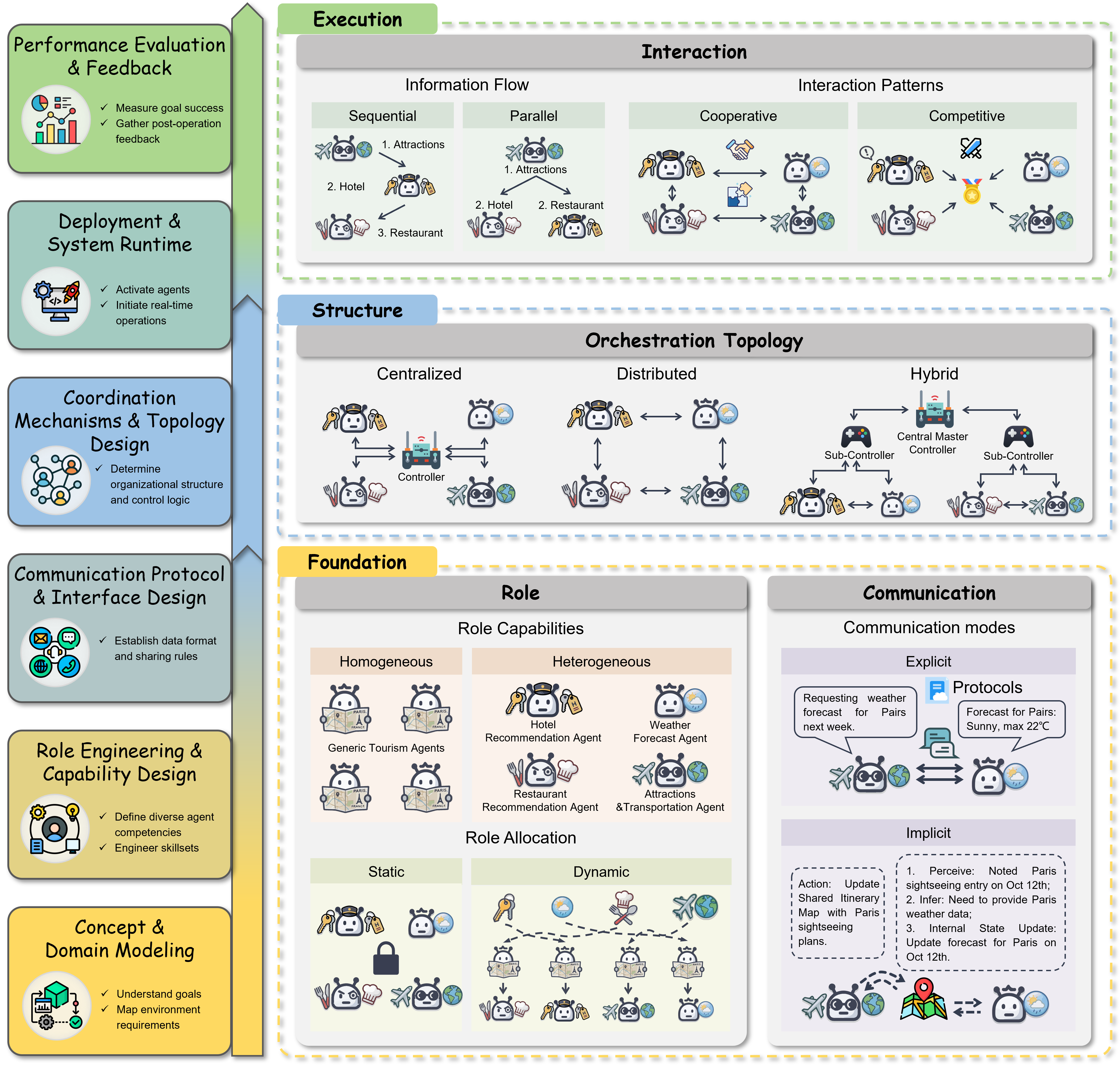}
    \caption{Overview of the multi-agent collaboration system. The right panel details a three-tiered design framework comprising the Foundation, Structure, and Execution layers to categorize agent roles, communication modes, orchestration topologies, and interaction dynamics. The left panel outlines the bottom-up system lifecycle, progressing from concept and domain modeling through to performance evaluation and feedback.}
    \label{fig:collaboration}
\end{figure}

Multi-Agent Systems (MAS) are analogous to division of labour and cooperation in human societies, where agents collaborate to solve complex problems. By integrating agents with distinct capabilities, MASs can tackle tasks beyond the ability of any individual agent\cite{liu2025advances, li2024survey}. In this paper, we formally define a MAS $S$ as
\begin{equation}
S = (\mathcal{A},\; \mathcal{E},\; \mathcal{C},\; \mathcal{G},\; \Pi),
\end{equation}
where
\begin{itemize}
    \item $\mathcal{A} = \{a_1, \ldots, a_N\}$ is the set of agents;
    \item $\mathcal{E}$ denotes the shared environment that may influence the operation of a MAS $S$; 
    \item $\mathcal{C}$ is the communication protocol that governs message exchange;
    \item $\mathcal{G}$ denotes the orchestration topology describing the architecture of a MAS $S$;
    \item $\Pi$ denotes the set of collaboration policies.
\end{itemize}

Under this formulation, we argue that multi-agent collaboration should be understood as a structured and goal-oriented computational paradigm, which can be decomposed into four key components: \textbf{Role}, \textbf{Communication}, \textbf{Orchestration}, \textbf{Interaction}. As shown in Fig. \ref{fig:collaboration}, each component addresses a fundamental question in collaborative intelligence.

\paragraph{Role} \textit{Who acts?} 
Let $z_i$ denote the role assigned to agent $a_i \in A$. We define role specialization through the mapping $z_i = \rho(a_i), \quad z_i \in Z$, where $\rho(\cdot)$ is the role allocation function and $Z$ denotes the set of possible roles. This component determines which agents are needed for collaboration and clearly specifies their responsibilities, such as planning, reasoning, or execution.

\paragraph{Communication} \textit{How do agents communicate?}
During collaboration, agents exchange messages with one another subject to the protocol $\mathcal{C}$. Let $m_{i \rightarrow j,t}$ denote a message sent from agent $a_i$ to agent $a_j$ at time step $t$. The communication process can be modeled by a message-generation mechanism $\mathrm{Comm}_{\mathcal{C}}$, such that
\begin{equation}
m_{i \rightarrow j,t} \sim \mathrm{Comm}_{\mathcal{C}}(\cdot \mid a_i, a_j, \mathcal{H}_{i,t}),
\end{equation}
where $\mathcal{H}_{i,t}$ denotes the information available to agent $a_i$ at time step $t$, including its local interaction history and any accessible shared context. This component specifies how the message is generated, structured, and transmitted among agents.

\paragraph{Orchestration} \textit{How are agents connected?}
Let $G$ denote the organizational structure of the MAS $S$. The orchestration topology is defined as
\begin{equation}
\mathcal{G} = (\mathcal{A},\; \mathcal{E}_c),
\qquad
\mathcal{E}_c \subseteq \mathcal{A} \times \mathcal{A},
\end{equation}
where $\mathcal{E}_c$ represents the set of communication links between agents.
If
\begin{equation}
\mathcal{E}_c =
\{(a_i,a_j) \in \mathcal{A} \times \mathcal{A} \mid i \neq j\},
\end{equation}
then all distinct agents are fully interconnected, allowing direct communication between every pair of agents in the MAS $S$. This component focuses on the organizational structure of agents and the design of message-passing pathways.

\paragraph{Interaction} \textit{How do agents collaborate during execution?}
Interaction characterizes how a MAS $S$ operates as a whole under a collaboration policy $\pi_{\mathrm{coll}} \in \Pi$ and environment $\mathcal{E}$. It describes the execution process in which agents act, exchange messages, and coordinate with each other to accomplish the task. Interaction can be viewed as the implementation of the collaboration policy $\pi_{\mathrm{coll}}$ at the execution level. Under this view, a MAS $S$ induces an execution process that can be represented as trajectories $\tau \sim \pi_{\mathrm{coll}}$ consisting of actions, inter-agent messages, and intermediate reasoning steps.

Together with Role, Communication, and Orchestration, Interaction completes the specification of a MAS by bridging structural design and runtime behaviour. Evaluating a MAS $S$ and its trajectories $\tau$ provides a unified lens for comparing different design choices across the four core components mentioned above.

\subsection{Role}
Understanding a role requires answering two fundamental questions: what can agents do and how responsibilities are distributed among them. Therefore, we decompose roles into two dimensions: \textbf{Role Capability} answers the first question,  and \textbf{Role Allocation} answers the second question by assigning these capabilities through a mapping function $\rho(\cdot)$.

\subsubsection{Role Capability}
In multi-agent collaboration, role capability refers to the set of competencies, action spaces, and observation spaces associated with agents engaged in the collaborative process\cite{liu2025advances,long2025role}. It captures not only what an agent $a_i$ can do (such as reasoning, planning, or tool use, as seen in \ref{sec:individual_intelligence}), but also what information it can access and how it can influence the environment $E$ or other agents $a_j \in A$. In some frameworks\cite{dong2024villageragent,chan2024chateval,he2025selfcorrect}, multiple agents are instantiated from the same LLM and consequently share identical capabilities, resulting in homogeneous role configurations. In contrast, other frameworks\cite{zhong2024heterogeneous,ye2025x,guo2025heterogeneous} employ agents built on different models or deliberately assign specialized responsibilities, producing heterogeneous role configurations characterized by asymmetric capabilities and differential access to information. As shown in Fig.\ref{fig:collaboration}, role capabilities in multi-agent collaboration can be broadly classified into \textbf{homogeneous} and \textbf{heterogeneous} types.


\paragraph{Homogeneous Role Capabilities}
In MASs, multiple agents are driven by the same underlying LLM and consequently share highly similar reasoning capability, action space, and observation scope. In this paradigm, the variation among agents arises primarily from role prompting or task framing rather than differences in intrinsic capabilities or information access. This homogeneous setup provides a controlled environment to systematically study how role definitions and interaction protocols shape collective behaviour, without confounding factors from model heterogeneity.

A canonical example of homogeneous role capabilities is frequently observed in multi-agent frameworks where agents are instantiated from the same underlying language model. In CAMEL\cite{li2023camel}, both AI User and AI Assistant are created from a single LLM and interact symmetrically, with differences mainly arising from prompt instructions rather than model heterogeneity. Similarly, Generative Agents\cite{park2023generative} simulate human-like behaviours through multiple agent instances derived from the same base model, enabling complex social interactions and emergent narratives in a controlled, homogeneous environment. Frameworks like SelfCorrect-Agent\cite{he2025selfcorrect} and Chateval\cite{chan2024chateval} further explore homogeneous multi-agent setups for robust reasoning and evaluation, leveraging symmetric LLM-based agents to iteratively correct or debate responses. VillagerAgent\cite{dong2024villageragent} demonstrates that even in graph-structured task coordination, MAS can be instantiated from identical models, emphasizing interaction over agent heterogeneity.

Building on these observations, homogeneous roles offer several practical advantages. Reusing the same model interfaces and toolsets across agents reduces engineering complexity, and coordination mechanisms are simplified because capability mismatches are minimized. At the same time, entirely homogeneous teams may encounter limitations when addressing highly complex or specialized tasks, where diversity in role capabilities—such as verification, domain-specific expertise, or advanced reasoning—becomes critical for robust and effective collaboration.

\paragraph{Heterogeneous Role Capabilities}
In contrast, heterogeneous role capabilities describe systems in which agents are driven by different models or are designed with distinct functional specializations. Agents may differ in computational reasoning ability, observation access, or permitted actions. This form of heterogeneity expands system expressiveness, enabling division of labor, modular interaction, and specialization, which is crucial for complex tasks that require expertise beyond general reasoning. 

A notable example is the X-MAS framework\cite{yao2026hieramas}, which explicitly constructs multi-agent systems with diverse LLMs to harness collective intelligence across models of varying strengths and specializations, demonstrating significant improvements in both general reasoning and specialized domains compared to homogeneous setups. Similarly, the Adaptive Heterogeneous Multi-Agent Debate (A-HMAD) framework\cite{zhou2025adaptive} assigns distinct roles to agents—such as logical reasoning, factual verification, and strategic planning—enabling more effective error-checking and domain coverage, and empirically reducing factual errors across benchmarks. More recently, MAPORL\cite{park2025maporl} demonstrates the emergence of heterogeneous roles through post-training and RL, enabling specialized coordination and task optimization in multi-agent systems.

Heterogeneous capabilities also arise naturally in systems that assign task-specific responsibilities. For example, multi-agent educational personalization systems assign Profiler, Rewriter, and Evaluator agents with distinct objectives and information access to tailor STEM content to individual learners\cite{vaccaro2025multi}. In medical decision-making, frameworks like Mdagents employ LLM-based agents with complementary roles—such as symptom analysis, treatment recommendation, and risk assessment—improving diagnostic performance and safety\cite{kim2024mdagents}. In software engineering, automated document summarization benefits from multi-agent LLM collaboration, where Extractor, Summarizer, and Validator agents coordinate to generate concise and accurate software documentation\cite{nguyen2026automated}, while in industrial optimization, OptDisPro\cite{li2025optdispro} dynamically assigns agents to search, evaluate, or refine solutions, flexibly balancing exploration and exploitation and outperforming homogeneous optimization agents. Collaborative frameworks like Autogen\cite{barbosa2024collaborative} further illustrate heterogeneity through differentiated observation spaces, where commanders interact directly with users or environments while subordinate agents receive only intermediary context.

Overall, these studies demonstrate that heterogeneous role capabilities consistently enhance multi-agent systems by promoting specialization, improving task performance, and increasing robustness and adaptability across diverse domains.

\subsubsection{Role Allocation}
Role allocation refers to the division of labour in collaborative agent systems, involving the assignment of tasks and the clarification of responsibilities among agents. Although the concepts of \textit{task} and \textit{role} are often used in the multi-agent literature\cite{campbell2011multi,amayuelas2025self,borghoff2025organizational,youwai2026large}, they describe different aspects of collaboration. Tasks usually refer to the units of work to be executed, whereas roles $z_i$ represent the functional positions that enable agents $a_i$ to perform those tasks. Across a range of existing frameworks\cite{yue2025masrouter,tarasova2025decentralized,goel2025r3dm}, roles emerge from the mapping $\rho: A \rightarrow Z$ between agent capabilities and task requirements, which explains why the two notions are often closely coupled in practice. Here, \textit{role} refers to the entities endowed with the capability to address particular tasks, rather than the tasks themselves. According to the mechanism by which roles $z_i$ are distributed via $\rho(\cdot)$, \textbf{static role allocation} and \textbf{dynamic role allocation} approaches are illustrated in Fig.\ref{fig:collaboration}.


\paragraph{Static Role Allocation}
Under static role allocation in multi-agent systems, agent roles and responsibilities are specified prior to task execution. By defining the team structure in advance, this approach establishes clear boundaries in terms of agent capabilities, role identities, and task objectives. In practice, static role allocation is typically implemented through manually designed templates, often realized using prompt engineering techniques to specify role descriptions.

This paradigm has been widely used in multi-agent systems designed for structured software engineering workflows. For example, ChetDev\cite{qian2024chatdev} organizes agents into predefined roles within a software development pipeline, enabling coordinated execution across stages such as planning, coding, and testing. Similarly, CodeDelegator\cite{fei2026codedelegator} assigns agents specialized roles such as context manager and code reviewer to mitigate context pollution in collaborative coding tasks. ATOMIZER\cite{zhu2026atomizer} adopts a comparable design for version control scenarios, where agents handle distinct aspects of commit untangling and conflict resolution. Static role allocation is also effective in data-centric and document-oriented tasks where responsibilities can be clearly partitioned. For instance, SheetAgent\cite{chen2024sheetagent} divides spreadsheet-related operations into predefined roles such as data entry and data analysis, allowing agents to focus on specific subtasks while reducing coordination overhead. In requirements engineering, the MARE system\cite{jin2024mare} similarly distributes responsibilities among agents with specialized expertise to support structured requirement analysis and task delegation.

Beyond general data processing tasks, static role allocation has also been applied in domain-specific multi-agent systems where reliability and consistency are essential\cite{yao2025comal}. In healthcare applications, ColaCare\cite{wang2025colacare} employs predefined roles for collaborative health record modeling to ensure stable and accurate task execution. MEDCO\cite{wei2024medco} adopts a similar design in medical education, where agents assume roles such as tutor and assessor to guide and evaluate learners within a structured training process. Several systems employ static role allocation to facilitate structured conversational or collaborative interactions. For example, CGMI\cite{jinxin2023cgmi} assigns agents roles based on task requirements to support coordinated multi-agent reasoning, while MultiAgentESC\cite{xu2025multiagentesc} defines specialized conversational roles to improve the quality and stability of emotional-support dialogues.

Overall, static role allocation provides a simple yet effective mechanism for organizing multi-agent collaboration in structured environments. By clearly separating responsibilities and limiting role ambiguity, it significantly reduces coordination complexity and improves execution efficiency. However, its reliance on predefined structures also introduces rigidity. When task requirements evolve or unexpected situations arise, static role allocation may lack the flexibility required for adaptive collaboration, motivating the development of dynamic role allocation strategies.

\paragraph{Dynamic Role Allocation}
In contrast to static role allocation, dynamic role allocation enables agents to adaptively select or adjust their roles based on task requirements and environmental context. Rather than relying on predefined templates, agents can determine the expertise required for a task and organize collaboration at runtime. As a result, roles may emerge and evolve during execution, allowing multi-agent systems to operate more flexibly in dynamic and open-ended environments.

Early research on dynamic role allocation primarily focuses on dynamic role selection, where agents recruit collaborators from a candidate pool according to task requirements. In this paradigm, agents identify the skills needed for a task and select suitable participants to form a task-oriented team. For instance, AutoAgents\cite{chen2024autoagents} allows agents to autonomously choose roles from a pool of available agents, enabling flexible collaboration in environments where tasks and agent capabilities continuously change. Similarly, the MedAide framework\cite{yang2025medaide} explores this mechanism in medical decision-making scenarios, where agents dynamically select collaborators and adjust their roles based on evolving medical intents and information sources. In Agent Hospital\cite{li2024agent}, agents are recruited to address specific clinical situations, with roles dynamically determined according to the needs of the medical scenario.

More recent work extends this paradigm toward active initialization, where agents not only select collaborators but also dynamically instantiate or expand agent teams during task execution. In this setting, agents can initiate collaboration by generating new agents or integrating additional participants as task complexity increases. For example, AgentVerse\cite{chen2024agentverse} allows agents to dynamically initialize and organize multi-agent teams according to task demands, enabling roles to emerge through real-time coordination rather than predefined assignment. Auto-scaling LLM-based multi-agent systems\cite{perera2025auto} further explore this idea by dynamically integrating agents to scale system capacity based on workload and task complexity. Beyond systems, dynamic role allocation has also been investigated in reinforcement learning settings. Roma\cite{wang2020roma} demonstrates that agents can develop role specialization through Multi-Agent Reinforcement Learning (MARL)\cite{lowe2017multi, sunehag2017value, rashid2020monotonic}, allowing roles to emerge and evolve through interaction. Building on this idea, ResMAS\cite{zhou2026resmas} dynamically assigns roles to maintain system resilience when unexpected events or failures occur. Similarly, ARL-SMCS\cite{tu2025adaptive} enables agents to learn role assignments through evolutionary reinforcement learning in UAV–vehicle collaboration tasks. 

In summary, dynamic role allocation significantly enhances the flexibility and scalability of multi-agent collaboration by allowing agents to reorganize themselves according to task requirements. However, this adaptability introduces additional coordination complexity, as agents must continuously evaluate task states, available resources, and the roles of other participants. Consequently, while dynamic role allocation is well suited for open-ended and evolving environments, it may also incur additional overhead in maintaining stable and efficient coordination.

\subsection{Communication}
Communication can be described at different levels. \textbf{Communication Modes} capture the high-level patterns of information sharing, whereas \textbf{Communication Protocols} define the concrete rules and procedures through which these patterns are implemented in practice.

\subsubsection{Communication Modes}
Communication is a fundamental mechanism that enables coordination and information sharing in MASs\cite{guo2024large, yan2025beyond}. In collaborative settings, agents exchange information to align their decisions, share intermediate results, and collectively progress toward task objectives. In LLM-based MASs, communication typically occurs through structured or unstructured message exchanges, where the communication protocol $C$ governs how messages $m_{i \rightarrow j}$ are generated and transmitted between agents $a_i$ and $a_j$. In addition to such direct message passing, agents may also communicate by leveraging their perception capabilities, inferring information from observed behaviours, environmental states, or the actions reflected in others’ histories. As shown in Fig.\ref{fig:collaboration}, communication modes in agentic systems can be broadly categorized into \textbf{explicit communication} and \textbf{implicit communication}.


\paragraph{Explicit Communication}
Explicit communication refers to agents exchanging information through well-defined message formats. In this mode, agents explicitly transmit task instructions, intermediate results, reasoning traces, or feedback through structured or unstructured messages.

Communication messages can take various forms, ranging from structured data such as code or documents to more flexible natural-language exchanges. In MetaGPT\cite{hong2023metagpt}, agents including Product Managers, Architects, and Engineers coordinate software development workflows by exchanging structured documents, such as product requirement documents, system design specifications, and implementation plans. Similarly, TalkHier\cite{wang2025talk} introduces a structured communication protocol to support context-rich exchanges among agents. Beyond structured artifacts, some communication relies on natural-language messages. For example, MAGIS\cite{tao2024magis} coordinates multiple specialized agents to collaboratively analyze and resolve GitHub issues through iterative dialogue, where agents exchange natural-language analyses, proposed solutions, and verification feedback. Autodata\cite{ma2025autodata} similarly organizes agents to collect and process open web data, enabling them to share extracted information, data summaries, and task progress through shared communication channels. EcoLANG \cite{mou2025ecolang} leverages context and social simulation, enabling agents to establish interpretable coordination protocols for efficient task execution. MegaAgent\cite{wang2025megaagent} further extends this paradigm by supporting large-scale autonomous collaboration without rigid predefined workflows, allowing agents to dynamically generate messages describing plans, intermediate findings, or coordination signals.

Subsequent studies further investigate how the design and organization of messages influence the efficiency and scalability of multi-agent communication. The economical communication pipeline proposed in AgentPrune\cite{zhang2025cut} reduces redundant interactions by filtering, summarizing, and compressing agent messages, demonstrating that carefully designed message representations can significantly lower token consumption while preserving essential task information. To further improve communication efficiency, TodyComm\cite{fan2026todycomm} introduces a task-oriented dynamic communication strategy in which agents selectively exchange task-relevant messages across interaction rounds, adaptively determining both communication targets and message contents. Other approaches\cite{nascimento2023self, chen2024reconcile, yang2026dynamic} focus on adaptive or consensus-based communication mechanisms, where agents dynamically adjust messaging policies and exchange viewpoints to reach collective decisions through negotiation or voting. Beyond improving communication efficiency, recent research also examines how message exchange structures affect collaboration dynamics. For instance, AgentCoord\cite{pan2025agentcoord} provides visualization tools to analyze the structure and flow of agent messages, revealing how different communication patterns influence task efficiency and coordination behaviours.

In summary, explicit communication remains the dominant paradigm in LLM-based MASs due to its transparency, controllability, and compatibility with role-based collaboration. By combining structured communication protocols with flexible dialogue mechanisms, modern frameworks continue to improve coordination efficiency and scalability in increasingly complex agent ecosystems.

\paragraph{Implicit Communication}
Under implicit communication, agents infer the intentions and states of others by observing environmental changes and behavioral patterns. In this mode, communication emerges from perception, memory retrieval, and reasoning processes: agents monitor their surroundings, interpret the actions of peers, and use internal representations to infer context and coordinate decisions. A representative example of this paradigm is the Generative Agents framework\cite{park2023generative}. In this system, agents simulate human-like social behaviour by continuously perceiving environmental events, retrieving relevant memories, and generating contextually appropriate actions or utterances. 

In MARL\cite{zhu2024survey}, observation-driven communication can emerge naturally when agents optimize shared objectives. Shaw et al.\cite{shaw2022formic} study cooperative foraging agents that learned coordination behaviours through environmental feedback. Subsequently, Buscemi et al.\cite{buscemi2026numbers} demonstrate that LLM–based agents could implicitly align their numerical strategies through repeated interactions, effectively conveying information through observable choices rather than explicit messages.

Subsequent work has extended this principle to scenarios requiring more sophisticated reasoning about other agents. Wang et al.\cite{wang2022tomc} introduce a theory-of-mind–based framework in which agents anticipate teammates’ goals by interpreting their observable actions, thereby improving coordination in the absence of explicit communication channels. In addition, communication-free approaches have demonstrated that direct message passing may not always be necessary. For example, Ge et al.\cite{ge2025efficient} show that agents could achieve effective cooperation in tasks such as box-pushing solely through observation and environmental interaction. LatentMAS\cite{zou2025latent} facilitates implicit communication by enabling agents to infer intentions and synchronize actions through shared environmental cues, providing a robust coordination mechanism when explicit messaging is costly or unavailable.

\subsubsection{Communication Protocols}
\label{Protocols}
Multi-agent communication protocols $C$, also referred to as communication rules, define how agents exchange information, coordinate actions, and achieve shared goals in distributed environments. In MASs, communication protocols support a variety of collaborative processes, including task allocation, information sharing, negotiation, decision-making, and conflict resolution.

Similar to how the TCP/IP\cite{hunt2002tcp,denardis2014global} and HTTP\cite{gourley2002http,campbell2011multi} protocols ushered in an unprecedented era of global connectivity on the Internet, standardized agent communication protocols are becoming a key infrastructure for building interconnected intelligent systems\cite{chen2022unmanned,stefan2025interconnected,tang2026llm} and collaborative intelligence networks\cite{yang2024llm,chen2025internet,kwon2025cp}. By providing common communication standards, these protocols\cite{ehtesham2025survey} enable heterogeneous agents developed by different platforms or organizations to interact seamlessly, thereby facilitating large-scale agent ecosystems.

Table \ref{tab:agent_protocols} compares several representative communication protocols. In this paper, we categorize existing communication protocols into two main groups: internal protocols, which govern communication among agents within a multi-agent system, and external protocols, which regulate interactions between agents and external systems, such as tools, services, or user interfaces.

\begin{table}[!htbp]
\centering
\caption{Comparison of agent communication protocols.}
\label{tab:agent_protocols}
\footnotesize
\renewcommand{\arraystretch}{1.25}
\setlength{\tabcolsep}{5pt}
\begin{adjustbox}{max width=1.05\textwidth}
\begin{tabular}{@{}L{2.60cm} L{3.20cm} C{1.40cm} C{2.60cm} L{3.40cm} C{0.80cm}@{}}
\toprule
\textbf{Protocol} & \textbf{Proposer} & \makecell{\textbf{Comm.}\\\textbf{Type}} & \textbf{Msg.\ Format} & \textbf{Core Functions} & \textbf{Year} \\
\midrule
\rowstrut Agora~\cite{marro2024scalable}              & University of Oxford   & \AIIA & \PDcap          & Consensus formation            & \ygray{2024} \\
\rowcolor{zebragray}
\rowstrut MCP~\cite{anthropic2024context}             & Anthropic              & \AIIT & \JSONRPCcap     & Standardized context access    & \ygray{2024} \\
\rowstrut ANP~\cite{chang2024anp}                     & ANP Community          & \AIIA & \JSONLDNLcap    & Meta-protocol negotiation      & \ygray{2024} \\
\rowcolor{zebragray}
\rowstrut LMOS~\cite{eclipse2025lmos}                 & Eclipse Foundation     & \AIIS & \JSONLDTDcap    & Group management               & \ygray{2024} \\
\rowstrut ACP~\cite{linux2025acp}                     & IBM                    & \AIIA & \JSONLDcap      & Async \& Sync interaction      & \ygray{2025} \\
\rowcolor{zebragray}
\rowstrut A2A~\cite{google2025a2a}                    & Google                 & \AIIA & \JSONRPCSSEcap  & Agent discovery                & \ygray{2025} \\
\rowstrut Agent Protocol~\cite{alengineer2025protocol}& AI Engineer Foundation & \AIIS & \RESTHTTPcap    & Standardized task API          & \ygray{2025} \\
\bottomrule
\end{tabular}
\end{adjustbox}
\vspace{3pt}
\parbox{\textwidth}{\footnotesize
\textit{Comm.\ Type}: primary interaction entities --- \AIIA\ Agent-to-Agent; \AIIT\ Agent-to-Tool; \AIIS\ Agent-to-System. \;
\textit{Msg.\ Format}: data serialization --- \PDcap\ Protobuf/Data-driven; \JSONRPCcap\ JSON-RPC; \JSONLDcap\ JSON-LD; \JSONLDNLcap\ JSON-LD + NL; \JSONLDTDcap\ JSON-LD + TD; \JSONRPCSSEcap\ JSON-RPC + SSE; \RESTHTTPcap\ REST/HTTP. \;
\textit{Core Functions}: primary capability or service provided by the protocol.
}
\end{table}

\paragraph{Internal Protocols}
Internal protocols define the communication rules among agents within MAS. These protocols primarily focus on enabling coordination, cooperation, and collective decision-making among multiple agents. Representative internal protocols include agent-to-agent communication frameworks such as Agent-to-Agent Protocol (A2A)\cite{google2025a2a}, which enables direct peer-to-peer communication and task handoff between agents, and Agent Network Protocol (ANP)\cite{chang2024anp}, which supports large-scale agent networks through mechanisms such as service discovery and message routing. 

\paragraph{External Protocols}
External protocols govern the interaction between agents and external environments, including tools, databases, APIs, and user interfaces, through standardizing how agents retrieve information, invoke tools, and exchange contextual data with external systems. One representative example is MCP\cite{anthropic2024context}, which provides a standardized interface for connecting agents with external tools and data sources and supports context sharing, tool registration, and structured request-response interactions.

Beyond the internal–external distinction, these protocols also differ in interoperability mechanisms and deployment models. A2A and ANP both target interoperability within MASs, but with different emphases: A2A focuses on standardized cross-platform collaboration, whereas ANP is oriented toward AI-native networking and scalable open agent ecosystems. In contrast, MCP is client–server architecture that enables external resources to be integrated through a unified interface, thereby reducing integration overhead and improving modularity, maintainability, and reusability.

\subsection{Orchestration}
Orchestration describes the global coordination structure that governs how multiple agents are organized and interact within a MAS $S$. From a system-level perspective, orchestration can be viewed as the collaboration topology $\mathcal{G} = (\mathcal{A}, \mathcal{E}_c)$: it defines how agents $a_i \in \mathcal{A}$ are connected, how information flows through the communication links $\mathcal{E}_c$ among them, and how coordination decisions are structured during task execution. Orchestration focuses on the higher-level structural organization that determines how these roles and communications are integrated into a coherent collaboration process.

Closely related to orchestration is the concept of agent routing, which determines how tasks or messages are dynamically directed to appropriate agents during execution. Routing mechanisms\cite{yue2025masrouter,liu2025rcr,zhao2026tcandon} are typically responsible for selecting which agent should process a particular subtask or piece of information, often based on capability matching, contextual reasoning, or learned policies. In contrast, orchestration defines the broader coordination topology within which routing decisions occur. In other words, routing can be regarded as an operational mechanism embedded inside an orchestration structure, while routing determines which agent should act next. For example, routing decisions may be locally determined by agents in decentralized architectures\cite{yang2025agentnet}.

Existing research generally organizes orchestration mechanisms into three representative topological paradigms: \textbf{centralized}, \textbf{distributed}, and \textbf{hybrid} orchestration. These paradigms differ primarily in how decision-making authority, coordination responsibilities, and information flow are structured across agents, and they provide distinct trade-offs in terms of scalability, adaptability, and control.


\subsubsection{Centralized Orchestration Topology}
Centralized orchestration topology follows a controller-centric coordination structure in which a single orchestrator maintains global decision authority over the collaboration process. In terms of coordination topology, agents are organized around a central controller that determines interaction flow. Information from different agents is typically aggregated by the orchestrator, which then determines subsequent actions and redistributes instructions. Consequently, routing decisions are largely governed by the central coordinator rather than emerging from peer-to-peer interactions.

Recent studies have explored various enhancements to the centralized orchestration topology. In terms of memory and planning, StackPlanner\cite{zhang2026stackplanner} utilizes a centralized controller with task-level memory to maintain global context during execution. Similarly, AOP\cite{li2025agent} relies on the orchestrator to centrally generate and sequence overarching action plans for all participating agents. To improve topological flexibility, the Puppeteer framework\cite{dang2025multi} enable the central node to dynamically reconstruct the interaction pathways among agents based on real-time task demands. Meanwhile, the efficient approach\cite{aso2024efficient} optimizes the central coordinator's routing decisions to minimize communication overhead within the centralized structure. Furthermore, CORL\cite{jin2025controlling} uses an RL-based orchestrator to dynamically govern agent interactions, acting as a gateway to balance overall system performance and budget. Another domain-specific application, CCMA\cite{de2025centrally} employs centrally coordinated reinforcement learning, demonstrating how a single orchestrator can manage complex constraints, such as power grid topology, through global state visibility.

While centralized orchestration enables consistent global planning and simplified routing control, it may also introduce scalability limitations and potential single points of failure when the number of participating agents grows.

\subsubsection{Distributed Orchestration Topology}

Distributed orchestration topology removes the global coordinator and distributes decision authority across agents. From the perspective of coordination topology, agents interact through peer-to-peer connections rather than through a centralized controller. Information flows directly among agents, and routing decisions are determined locally based on each agent's capabilities, policies, and observed context.

A representative example is AgentNet\cite{yang2025agentnet}, which models agent collaboration as a dynamically evolving network, equipping fully connected agents with distributed executor and router pools to autonomously determine task delegation through local reasoning. Symphony\cite{wang2025symphony} establishes a highly scalable decentralized framework that facilitates collective intelligence by allowing direct peer-to-peer interactions without centralized coordination bottlenecks. To optimize such distributed Orchestration, MAAC approaches\cite{liu2026learning} empower agents to dynamically learn and refine their decentralized coordination policies via reinforcement learning rather than relying on static routing rules. Furthermore, distributed orchestration requires robust mechanisms for local information management. AHKG\cite{yang2025llm} supports distributed planning by allowing agents to synchronize localized information clusters while inherently maintaining alignment with global topological objectives. 

Similar decentralized coordination strategies have been widely adopted across various domains to preserve agent autonomy. For instance, HIPPO-MAT\cite{ratnabala2025hippo} employs decentralized multi-agent reinforcement learning and graph neural networks for task allocation without a central controller, while AgentFlow\cite{chen2025agentflow} proposes a distributed coordination framework based on publish–subscribe communication and dynamic service elections. 

Collectively, these approaches demonstrate how distributed orchestration enables scalable collaboration while preserving agent autonomy and system robustness.

\subsubsection{Hybrid Orchestration Topology}
Hybrid orchestration topology combines centralized strategic coordination with decentralized execution mechanisms. In this paradigm, decision authority is partially centralized at higher levels of the system while lower-level agents retain autonomy during task execution.

From the perspective of coordination topology, hybrid orchestration typically adopts hierarchical structures. Early frameworks such as ChatDev\cite{qian2024chatdev} and MetaGPT\cite{hong2023metagpt} laid the groundwork for this paradigm by relying on centralized Standard Operating Procedures (SOPs) to constrain and guide the decentralized execution of role-playing agents. Building upon more complex hierarchies, AgentOrchestra\cite{zhang2025agentorchestra} organizes collaboration into three layers: a planning layer for task decomposition and global coordination, a functional orchestration layer for managing tools and specialized agents, and an execution layer for carrying out concrete operations. Similarly, planner–executor architectures\cite{wang2023plan} separate global reasoning from action execution. TOA\cite{yu2025tree} structures agents in a hierarchy where parent nodes provide centralized aggregation and reasoning, while leaf nodes maintain autonomy for parallel context processing. GoA\cite{yun2026graph} expands decentralized execution by enabling non-linear, localized communication among agents, while maintaining a globally predefined topological constraint to ensure centralized goal alignment. AdaptOrch\cite{yu2026adaptorch} further advances hybrid orchestration by introducing a centralized, task-adaptive orchestrator that dynamically selects the execution topology.

These approaches illustrate how hybrid orchestration can balance centralized strategic control with decentralized operational flexibility.

\subsection{Interaction}
Interaction characterizes how multiple agents coordinate during the execution phase of a task. Unlike static system design, which focuses on agent roles or architectures, interaction emphasizes the dynamic processes through which agents exchange information, update intermediate states, and jointly drive task completion under the current collaboration policy $\pi_{\mathrm{coll}}$.

\subsubsection{Information Flow}
Information flow governs the execution dynamics of multi-agent systems, dictating the orchestration of collaborative actions and the synchronization of shared states. Based on the temporal and logical dependencies of task execution, information flow can be primarily categorized into \textbf{sequential interaction} and \textbf{parallel interaction}.

\paragraph{Sequential Interaction}
In sequential interaction, agents exchange information in an ordered, step-by-step manner, where the output of one agent serves as the input for the next. This pattern is particularly suitable for tasks that can be decomposed into a sequence of dependent subtasks. 

Recent studies have extensively leveraged sequential interaction to structure agentic workflows. Frameworks such as HuggingGPT\cite{shen2023hugginggpt} and EAG\cite{gu2025explain} establish modular pipelines in which tasks are decomposed into discrete reasoning stages, whereas Chain of Agents\cite{zhang2024chain} employs a sequential relay to mitigate information loss in long-context tasks. To enhance adaptability, AnyMAC\cite{wang2025anymac} introduces dynamic cascading via next-agent prediction within sequential flows, a pattern also applied to specialized domains like recommendation in MACRec\cite{wang2024macrec}. Furthermore, frameworks such as MAgIC\cite{xu2024magic} and recent role-specialized pipelines\cite{barrak2025traceability} utilize these linear dependencies to provide systematic benchmarks and ensure traceability and accountability across multi-agent interactions.

The main advantage of sequential interaction lies in its interpretability and modularity. Each agent operates on well-defined inputs and produces traceable outputs, making the overall process easier to debug and optimize. However, this structure may suffer from error propagation, as mistakes in early stages can cascade through subsequent steps. Additionally, strict sequential dependencies may limit efficiency when tasks could otherwise be executed concurrently.

\paragraph{Parallel Interaction}
In parallel interaction, multiple agents operate simultaneously and independently on shared or decomposed tasks, followed by a mechanism for aggregating their outputs. Unlike sequential interaction, parallel interaction allows agents to explore diverse solution spaces concurrently, which can improve robustness and efficiency.

A foundational example of parallel interaction is the self-consistency decoding strategy\cite{wang2023selfconsistency}, which generates multiple reasoning paths concurrently and aggregates the final answer. This principle was soon extended to multi-agent debate frameworks\cite{du2024improving}, enabling several agents to independently propose and critique solutions in parallel. Building upon these concepts, recent architectures further exploit parallelism to overcome sequential bottlenecks. MoA methodology\cite{wang2025mixture} applies this at the model level by utilizing diverse LLMs as concurrent proposers and iteratively aggregating their outputs. At the execution level, the M1-Parallel framework\cite{zhang2025optimizing} deploys multiple agent teams simultaneously to explore distinct solution paths, minimizing latency via early termination. To address real-time environment constraints, parallelized planning-acting architectures\cite{li2025parallelized} decouple cognitive planning and physical execution into independent, concurrent threads. Furthermore, for complex problem-solving, AgentGroupChat-V2\cite{gu2025agentgroupchat} decomposes user queries into a task tree to assign sub-tasks to specialized agents operating in parallel. Similarly, DynTaskMAS\cite{yu2025dyntaskmas} employs dynamic DAGs to adaptively manage and execute these parallel sub-tasks, ensuring high system robustness.

Parallel interaction offers significant advantages in terms of scalability and diversity. By enabling multiple agents to explore different hypotheses simultaneously, the system can reduce the risk of converging to suboptimal solutions. However, it also introduces challenges in result aggregation and consistency maintenance. Designing effective mechanisms to reconcile conflicting outputs remains a key research problem in this setting.

\subsubsection{Interaction Patterns}
Interaction patterns describe how collective outcomes emerge from actions of individual agent. Existing studies generally categorize multi-agent interaction patterns into two major types: \textbf{cooperative interactions} and \textbf{competitive interactions}, depending on whether agents pursue shared or conflicting objectives.

\paragraph{Cooperative Interaction}
Cooperative interaction refers to scenarios in which agents share a common objective and coordinate their actions to maximize collective performance. In such settings, agents typically exchange information, divide responsibilities, and synchronize decisions to improve overall task efficiency and solution quality.
Recent studies have increasingly emphasized the role of structured coordination in enabling effective cooperation among agents. For instance, generative agent frameworks\cite{park2023generative} simulate socially interactive agents, demonstrating how cooperation can emerge in complex environments. Similarly, role-based multi-agent systems such as MetaGPT\cite{hong2023metagpt} organize agents into specialized roles (e.g., product manager, engineer), where collaboration is achieved through well-defined workflows. Similarly, domain-specific architectures like the DrugAgent framework\cite{liu2024drugagent} utilize distinct planner and instructor roles to maintain precise workflow boundaries and avoid conflicts during complex scientific problem-solving.  

For coordination settings where teammate behaviors are not known in advance, the ProAgent framework\cite{Zhang2024ProAgent} facilitates zero-shot coordination by enabling agents to proactively infer teammate intentions and adjust their strategies based on observed behaviors. Exploring the structural design of these interactions, socially-inspired collaboration mechanisms\cite{zhang2024exploring} reveal that the sequential orchestration of interaction topologies can significantly enhance cooperative outcomes without merely relying on scaling agent numbers. In decentralized settings with partial observability, the COMPASS framework~\cite{li2025cooperative} advances cooperative planning by introducing a multi-hop communication protocol, allowing agents to effectively exchange localized state information and synthesize adaptive skills.

To further ensure the reliability of cooperative interactions, recent methods have integrated structured evaluation and collective decision protocols. COPPER~\cite{bo2024Reflective} introduce system-level feedback loops, utilizing shared reflection mechanisms to resolve credit assignment issues and iteratively refine collective behaviors. Concurrently, electoral approaches~\cite{zhao2024electoral} to collective decision-making employ ordinal preferential voting mechanisms to systematically aggregate diverse agent preferences, thereby mitigating the limitations of simple majority rules and enhancing the robustness of joint decisions.

Overall, cooperative interaction in modern multi-agent systems is characterized by structured communication, role-based organization, and iterative alignment, which together facilitate efficient and scalable collaboration.

\paragraph{Competitive Interaction}
Competitive interaction arises when agents pursue individual objectives that may partially or fully conflict with those of other agents. In this setting, agents attempt to advance their own objectives through strategic behaviours such as negotiation, confrontation, or persuasion. 

Recent work has explored competitive interaction through mechanisms such as debate, negotiation, and strategic role-playing. For example, the debate framework proposed by Du et al.\cite{du2024improving} enables multiple agents to argue over candidate solutions, where each agent attempts to expose weaknesses in the reasoning of others. Building on this, Agent4Debate\cite{zhang2024beat} introduces a dynamic framework to test if LLM agents can surpass human strategic logic in structured linguistic confrontations. Negotiation-based settings provide another important perspective on competition. The GPT-Bargaining framework\cite{fu2023improving} studies how agents engage in bargaining processes, where each agent aims to optimize its own outcome under conflicting interests. To further quantify these behaviours, ALYMPICS\cite{mao2025alympics} provides a game-theoretic playground to observe how agents navigate classic dilemmas like the Tragedy of the Commons, while RTBAgent\cite{Cai2025RTBAgent} pushes competitive tactics into high-stakes real-time bidding auctions. In addition, role-playing approaches\cite{liang2024encouraging} introduce diverse agent roles to encourage strategic diversity. 

Recent architectures leverage competition as an optimization tool. Multi-agent KTO\cite{ye2025multiagentkto} reinforces strategic interactions by aligning agent preferences through game-theoretic objectives. To ensure these competitive processes remain robust, ALIGN\cite{zhu2026align} introduces performance guarantees to regulate delegated competitive reasoning. 

Overall, competitive interaction emphasizes strategic adaptation, conflict resolution, and dynamic response to opponents. Unlike cooperative settings that rely on alignment and shared objectives, competitive scenarios drive agents to refine their strategies through adversarial exchanges continuously.

\subsection{Evaluation}
\begin{figure}
    \centering
    \includegraphics[width=\linewidth]{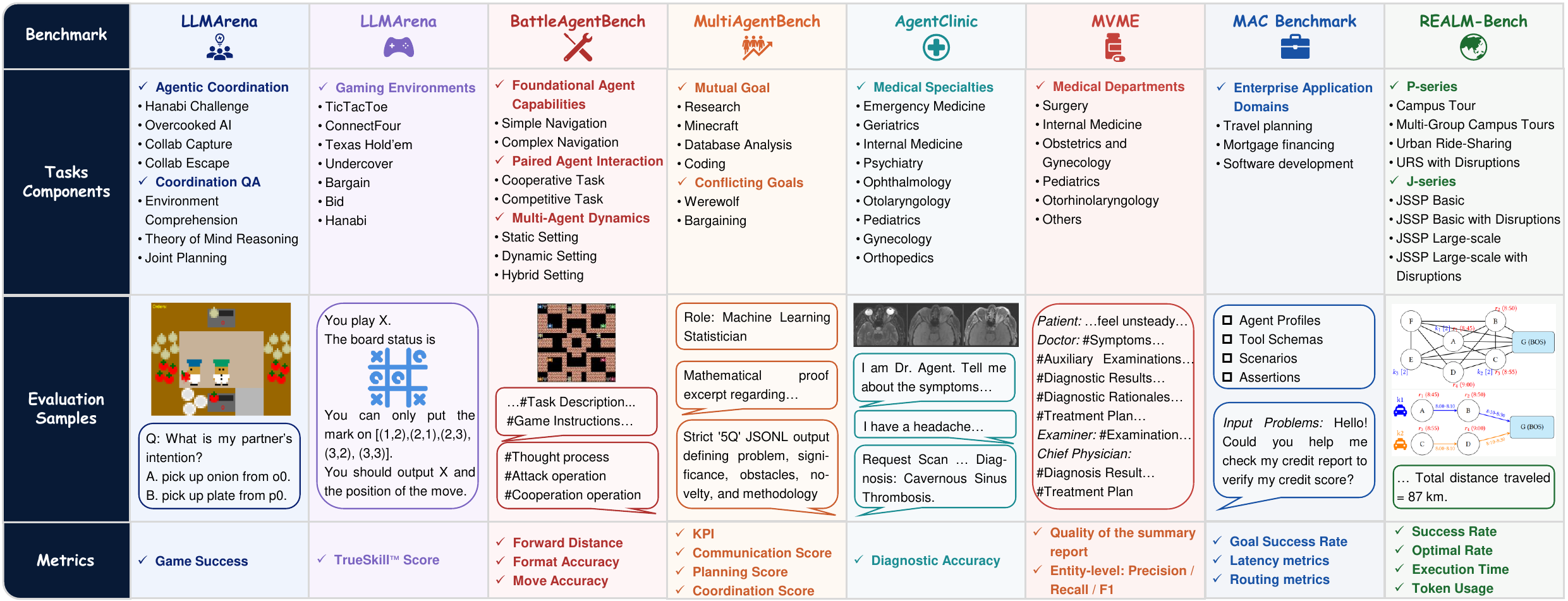}
    \caption{Evaluation benchmarks for multi-agent collaboration. This figure compares existing benchmarks across task components, evaluation samples, and evaluation metrics.}
    \label{fig:mas_case}
\end{figure}

A critical aspect of optimizing MAS lies in the construction of high-quality evaluation datasets\cite{lee2025gemmas,gioacchini2024agentquest} that capture realistic interaction trajectories $\tau$ and reflect diverse collaboration patterns. 

As shown in Figure~\ref{fig:mas_case}, recent literature has increasingly shifted toward evaluating dynamic MAS. As foundational efforts, LLM-Coordination\cite{agashe2025llm} establishes an early benchmark for analyzing how multiple large language models share information and collaborate to achieve shared objectives. Subsequent benchmarks have introduced heightened environmental complexity and complex interactive dynamics. LLMArena\cite{chen2024llmarena} assesses agents' adaptability and sequential decision-making in dynamic environments with incomplete information. Pushing beyond pure collaboration, BattleAgentBench\cite{wang2024battleagentbench} provides a comprehensive framework to evaluate both cooperative and competitive capabilities, offering quantifiable metrics for agents operating in zero-sum and non-zero-sum game scenarios. Furthermore, MultiAgentBench\cite{zhu2025multiagentbench} evaluates orchestration quality through milestone-based KPIs and explicitly investigates the efficacy of different communication topologies (e.g., graph-based structures) on collaborative efficiency. 

Pushing further toward realism, recent work explores high-fidelity, domain-specific evaluation settings that demand rigorous multi-round coordination. Medical testbeds like AgentClinic\cite{schmidgall2025agentclinic} and MVME\cite{fan2025aihospital} serve as advanced environments for testing sequential decision-making, information gathering, and dispute resolution mechanisms under high uncertainty. Enterprise-oriented scenarios have been studied to highlight end-to-end task completion and coordination efficiency in real-world applications\cite{shu2024towards}. REALM-Bench\cite{geng2025realm} targets real-world planning tasks requiring long-horizon reasoning and multi-agent coordination.

Overall, trajectory-level benchmarks better reflect the interactive and collaborative nature of MAS. Such datasets enable both accurate evaluation and informed refinement of agent behaviours.

\subsection{Discussion}

\begin{table}[!htbp]
\centering
\caption{Comparison of multi-agent frameworks, organized by collaboration design.}

\label{tab:multi-agent-frameworks}

\footnotesize

\renewcommand{\arraystretch}{1.25}

\setlength{\tabcolsep}{5pt}

\begin{adjustbox}{max width=1.05\textwidth}

\begin{tabular}{@{}L{2.80cm} C{1.40cm} C{0.90cm} C{1.00cm} C{1.90cm} C{0.90cm} C{1.10cm} C{0.60cm} C{0.60cm}@{}}

\toprule

\textbf{Framework} & \textbf{\#Agent} & \makecell{\textbf{Role}\\\textbf{Cap.}} & \makecell{\textbf{Role}\\\textbf{Alloc.}} & \textbf{Orchestration} & \makecell{\textbf{Exec.}\\\textbf{Seq.}} & \textbf{Interact.} & \textbf{Opt.} & \textbf{Year} \\

\midrule

\rowstrut Generative Agent~\cite{park2023generative} & 25        & Hom & Static  & \DistributedB & Par/Seq & \CoopCap & \na    & \ygray{2023} \\

\rowcolor{zebragray}

\rowstrut CAMEL~\cite{li2023camel}                   & 2         & Hom & Static  & \DistributedB & Seq     & \CoopCap & \na    & \ygray{2023} \\

\rowstrut GPT-Bargaining~\cite{fu2023improving}      & 3         & Het & Static  & \CentralizedB & Seq     & \CompCap & \cmark & \ygray{2023} \\

\rowcolor{zebragray}

\rowstrut MetaGPT~\cite{hong2023metagpt}             & unlimited & Het & Static  & \HybridB      & Seq     & \CoopCap & \na    & \ygray{2023} \\

\rowstrut AutoGen~\cite{wu2024autogen}               & unlimited & Het & Dynamic & \HybridB      & Par/Seq & \CoopCap & \na    & \ygray{2024} \\

\rowcolor{zebragray}

\rowstrut AgentVerse~\cite{chen2024agentverse}       & unlimited & Het & Dynamic & \HybridB      & Par/Seq & \CoopCap & \na    & \ygray{2024} \\

\rowstrut AutoAgents~\cite{chen2024autoagents}       & unlimited & Het & Dynamic & \HybridB      & Seq     & \CoopCap & \cmark & \ygray{2024} \\

\rowcolor{zebragray}

\rowstrut MAPoRL~\cite{park2025maporl}               & unlimited & Het & Static  & \HybridB      & Seq     & \CoopCap & \cmark & \ygray{2025} \\

\rowstrut OSC~\cite{zhang2025osc}                    & unlimited & Het & Dynamic & \DistributedB & Seq     & \CoopCap & \cmark & \ygray{2025} \\

\rowcolor{zebragray}

\rowstrut Puppeteer~\cite{dang2025multi}             & unlimited & Het & Dynamic & \CentralizedB & Par/Seq & \CoopCap & \cmark & \ygray{2025} \\

\rowstrut AgentForge~\cite{jafari2026lightweight}    & unlimited & Het & Dynamic & \CentralizedB & Seq     & \CoopCap & \cmark & \ygray{2026} \\

\rowcolor{zebragray}

\rowstrut MASFactory~\cite{liu2026masfactory}        & unlimited & Het & Dynamic & \HybridB      & Seq     & \CoopCap & \cmark & \ygray{2026} \\

\bottomrule

\end{tabular}

\end{adjustbox}

\vspace{3pt}

\parbox{\textwidth}{\footnotesize

\textit{Role Cap.}: functional diversity --- Hom = homogeneous; Het = heterogeneous. \textit{Role Alloc.}: role assignment --- Static = fixed; Dynamic = runtime. \textit{Orchestration}: coordination structure --- \DistributedB\ peer-to-peer; \CentralizedB\ single orchestrator; \HybridB\ mixed. \textit{Exec.\ Seq.}: temporal pattern --- Par = parallel; Seq = sequential; Par/Seq = both. \textit{Interact.}: social dynamics --- \CoopCap\ cooperative; \CompCap\ competitive. \textit{Opt.}: includes an optimization loop.

}

\end{table}

Table~\ref{tab:multi-agent-frameworks} summarises and compares various multi-agent frameworks from the perspective of collaboration design, highlighting not only the structural diversity of MASs, but also a shift in the design philosophy of LLM-based collaboration.

\paragraph{From homogeneous agents to functional specialization.}
One of the clearest trends is the increasing prevalence of heterogeneous role configurations. Early frameworks~\cite{park2023generative, li2023camel} with homogeneous agents are conceptually simple and easier to scale, but struggle with complex tasks as it tends to get trapped in repeated self-consistency loops\cite{zhong2024heterogeneous}. By contrast, heterogeneous MASs~\cite{hong2023metagpt, qian2024chatdev} introduce functional specialization, such as planning, coding, reviewing, or evaluation, thereby enabling division of labour and more structured cooperation. However, heterogeneity also introduces new challenges. As role diversity increases, so does interdependence among agents, making the system more sensitive to interface mismatches, cascading errors, and coordination overhead. In practice, a specialized agent is only useful if its outputs are both reliable and legible to downstream agents. Therefore, future MAS research should focus not only on creating more roles but also on defining principled role boundaries and mechanisms for capability calibration across agents.

\paragraph{Static versus dynamic role allocation.}
Static role assignment~\cite{hong2023metagpt, qian2024chatdev} is easier to design and interpret and remains effective for tasks with predictable decomposition structures. However, it assumes that task requirements are known in advance and remain stable throughout execution. This assumption becomes increasingly unrealistic in more realistic open-ended tasks, where intermediate results often reshape the problem itself. Dynamic role allocation~\cite{chen2024agentverse, chen2024autoagents} offers a more flexible alternative by allowing the system to instantiate, reassign, or retire roles at runtime according to task state, uncertainty, or resource constraints. This makes MASs better suited for non-stationary environments and long-horizon problems. Nevertheless, many current systems implement dynamic roles heuristically, but do not yet learn when specialization is necessary, when redundancy is beneficial, or when additional agents simply add noise.

\paragraph{Explicit communication dominates, but may not scale indefinitely.}
Explicit communication remains the dominant mode across current LLM-based MASs. Explicit messages are transparent, inspectable, and well aligned with the text-based interfaces of LLMs. They make execution trajectory easier to debug and allow designers to impose structured debate or critique patterns. As the number of agents and interaction rounds grows, explicit communication quickly becomes expensive, verbose, and cognitively inefficient. Repeated broadcasting of long textual states creates token overhead, increases latency, and can amplify inconsistency when agents interpret shared context differently. More importantly, explicit communication~\cite{zou2025latent} often exposes raw intermediate reasoning rather than the task-relevant state, which limits scalability. For this reason, implicit communication deserves much more attention in future LLM-based MASs. In embodied MASs \cite{bechlioulis2018collaborative,zhao2025udon,yang2025implicit}, implicit coordination often emerges through shared memory, environment state, stigmergic signals, or policy-conditioned behaviour. Translating such ideas into LLM-based settings may enable more efficient coordination through latent state sharing or memory traces rather than full natural-language exchange. The central challenge is to preserve interpretability while reducing communication cost.

\paragraph{Orchestration topology reflects a trade-off between control and flexibility.}
The comparison in Table~\ref{tab:multi-agent-frameworks} suggests that no orchestration topology is universally superior. Centralized orchestration~\cite{li2025agent, dang2025multi} offers strong global oversight, clear responsibility assignment, and easier enforcement of task structure. However, centralized control can create bottlenecks, since a single coordinator must handle all information aggregation and decision-making, resulting in constrained parallelism and potential overload in both computation and communication. By contrast, distributed orchestration~\cite{yang2025agentnet, wang2025symphony} improves autonomy and robustness to single-point failure, and can better support peer collaboration. But it also makes consensus formation, conflict resolution, and global coherence substantially harder. Hybrid orchestration~\cite{zhang2025agentorchestra, yu2026adaptorch} has therefore become increasingly common, as it attempts to combine top-down structure with localized autonomy. This trend indicates that orchestration should be treated as an adaptive design variable rather than a fixed architectural choice. A key problem is topology selection under task conditions. Different problem types likely favour different coordination graphs: tightly coupled reasoning may require stronger centralization, whereas exploratory or modular tasks may benefit from decentralization. Future work should move beyond categorical labels such as centralized or distributed, and instead model orchestration as a dynamic topology that can evolve with task complexity, uncertainty, and agent performance.

\paragraph{Sequential and parallel execution should be viewed as complementary paradigms.}
Temporal organization is becoming an increasingly important part of MAS design. Sequential execution~\cite{li2023camel, hong2023metagpt} is suitable for tasks that require refinement, verification, or strict stage ordering. In contrast, parallel execution~\cite{wu2024autogen, chen2024agentverse} is valuable for candidate generation, diversified search, or decomposable subtasks, where breadth and speed are more important than immediate consistency. The key insight is that the execution sequence strongly shapes the epistemic behaviour of the system. Sequential pipelines encourage coherence but risk error propagation, since early mistakes constrain downstream reasoning. Parallel pipelines improve diversity and robustness but may create expensive aggregation problems when candidate outputs conflict. The most capable future MASs will likely combine both modes adaptively.

\paragraph{Cooperation dominates, while strategic competition remains underexplored.}
Most existing MAS frameworks~\cite{liu2024drugagent, Zhang2024ProAgent} are cooperative, which aligns naturally with benchmark design and application goals centred on task completion. However, cooperation settings neglect scenarios where disagreement or negotiation is productive rather than harmful. Competitive interactions~\cite{fu2023improving, zhang2024beat} enhance collective reasoning quality through adversarial review, structured debate, and bargaining. The challenge is that without carefully designed incentives and adjudication mechanisms, competitive interactions may degenerate into unproductive contradiction or reward gaming. Future MASs should therefore explore richer interaction regimes and dynamically shifting incentives. This would make MASs more suitable for complex real-world tasks such as policy analysis, scientific criticism, market simulation, and risk assessment.

Taken together, these observations indicate that MASs are gradually evolving from static, rule-based coordination toward dynamic, policy-driven collaboration with increasing adaptability. However, this trend exposes several unresolved challenges. First, scalability remains a major bottleneck. Increasing the number of agents does not guarantee better performance\cite{kim2025towards}. Second, reliability becomes more difficult in multi-agent settings, because errors may emerge not only from individual agents but from their interactions, including misinformation propagation, collaborative failure modes, and unstable feedback loops. Third, cost is likely to become central: practical MAS deployment requires balancing performance gains against token consumption, latency, and infrastructure complexity. These challenges imply that future MAS research should prioritize three directions. The first is adaptive coordination, where roles, topology, and communication strategies are selected online rather than fixed a priori. The second is collaboration-efficient design, which seeks better performance with fewer messages, fewer agents, and more compact shared representations. The third is trustworthy multi-agent reasoning, including interpretable coordination traces, fault localization, and robust evaluation protocols.

\section{Multi-Agent System Failure Attribution}
\label{Multi-agent failure attribution}
In recent years, although multi-agent systems can improve complex task-solving performance through collaboration among specialized agents, their highly coupled collaborative processes also amplify system vulnerability, making errors more likely to arise during task decomposition, inter-agent communication, and coordinated decision-making\cite{pan2025whymultiagentfail,Ma2025DiagnosingFR}. More importantly, such errors often do not manifest at the moment they are introduced, but instead propagate, accumulate, or even evolve through subsequent interactions before ultimately surfacing as system-level failures. As a result, anomaly attribution in multi-agent systems becomes substantially more challenging.

\begin{figure}    
\centering    
\includegraphics[width=0.95\linewidth]{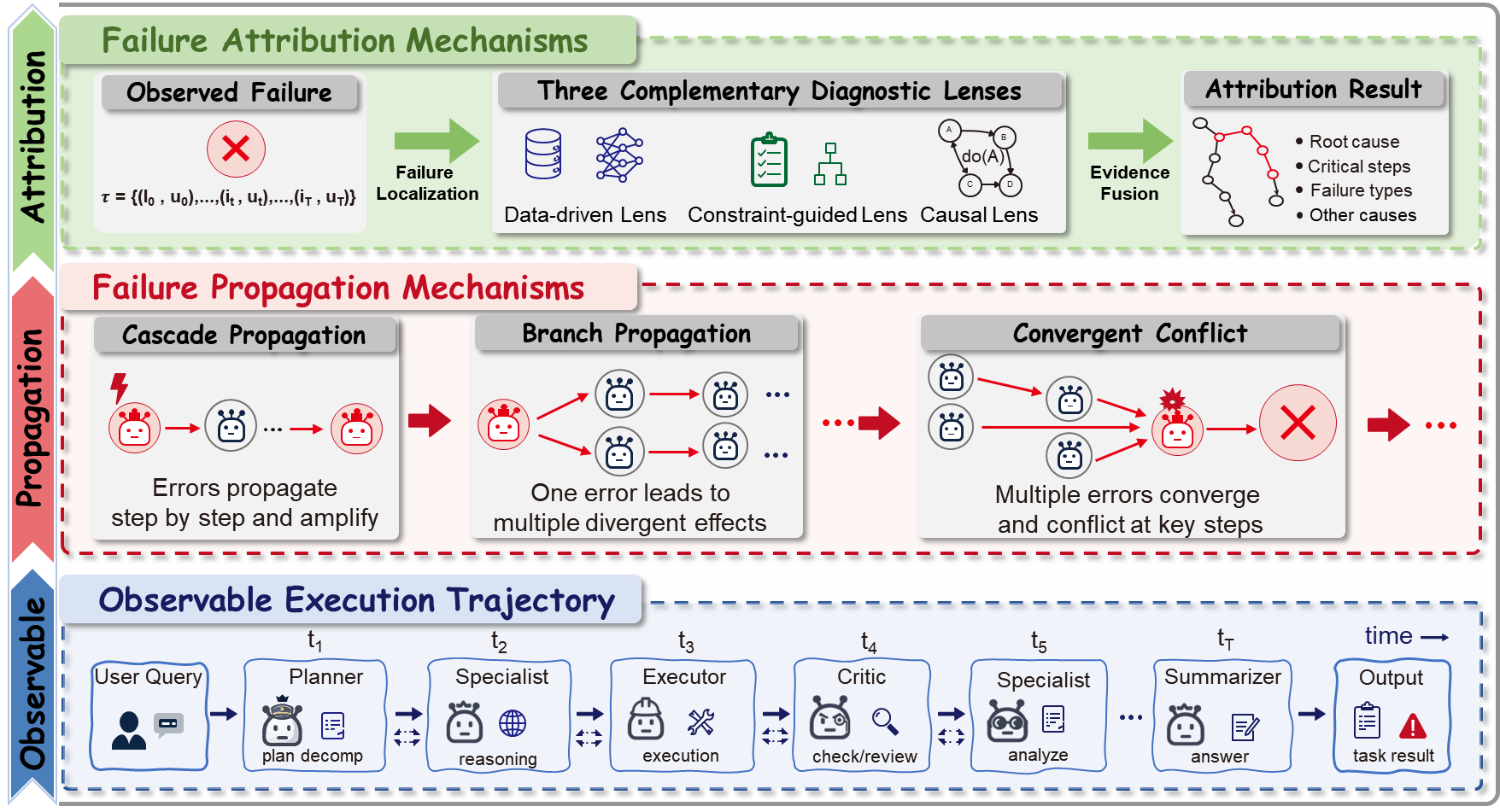}    
\caption{Mechanism of Failure Attribution in Multi-Agent Systems.}    
\label{fig:attribution-mech}
\end{figure}

\subsection{From Collaboration to Diagnosis}
A fundamental challenge arises from the tight coupling of multi-agent interactions, where initial localized errors such as hallucinations frequently propagate through the system, culminating in intricate cascading failures that obscure the primary root cause\cite{Lin2025AgentAskMS,pan2025whymultiagentfail}. As execution trajectories grow longer and contain increasing amounts of interactive information, manually locating agent-level faults and their associated time steps becomes highly inefficient and unscalable\cite{zhang2025which,anonymous2026seeing}. Therefore, an effective automatic failure attribution mechanism plays a key role in clarifying complex interaction processes in MAS and enabling fine-grained, targeted feedback to support directed improvement. It is also an important foundation for enhancing system reliability and trustworthiness\cite{in2026rethinking,Ma2025DiagnosingFR,Zhang2025AgenTracerWI}. 

Failure attribution in MAS is the process that uses the complete execution trajectory as its basis to automatically identify the responsible agents and critical steps when execution failures or anomalous behaviors emerge\cite{zhang2025which,in2026rethinking,wang2026flat}. Additionally, it analyzes the failure's category, etiology, and propagation mechanism within the entire collaborative workflow to articulate who introduced the failure, where it originated, and why it ultimately precipitated system failure\cite{Ma2025DiagnosingFR}.

A multi-agent execution trajectory $\tau$ refers to the time-ordered dynamic sequence formed during task execution in a MAS, under a given user query, initial environment state, and system configuration\cite{Zhang2025AgenTracerWI}. It consists of the contextual information, state transitions, communication interactions, tool usage, decision-making actions, and environmental feedback of the agents. Such a trajectory captures the complete execution process from the initial configuration to the terminal state, and serves as the fundamental basis for analyzing collaborative behaviors, failure propagation paths, and root causes of failure in MAS\cite{Zhang2025AgenTracerWI, Ge2025WhoII,wang2026flat}. It can be formally defined as
\begin{equation}    
\tau = \big((i_0, u_0), \dots, (i_t, u_t), \dots, (i_H, u_H)\big),
\end{equation}
where $H$ denotes the trajectory horizon, and $(i_t, u_t)$ denotes the active agent and its corresponding action at time step $t$.

\subsubsection{Formal Expression of the Attribution Process}

To define the condition for failure occurrence, a trajectory evaluation function $Z(\tau)$ is introduced. When $Z(\tau)=1$, it indicates that the final result state of the trajectory $\tau$ is evaluated as anomalous. Then, given a system configuration $\Omega$, a user query $q$, and the complete execution trajectory $\tau$, the failure attribution model $f$ identifies the agent $I$ responsible for the failure and the specific time step $t$ at which the failure is introduced. The entire multi-agent failure attribution process can be formally expressed as the following piecewise function:

\begin{equation}
    (I, t) = 
    \begin{cases} 
        f(\Omega, \tau, q), & \text{if } Z(\tau) = 1, \\ 
        \emptyset, & \text{otherwise}. 
    \end{cases}
\end{equation}

The above equation indicates that only when the system trajectory is judged to be anomalous (i.e., $Z(\tau) = 1$), the attribution model $f$ will take the system configuration $M$, trajectory $\tau$, and user query $q$ as input and return the failure-responsible agent and the corresponding decisive time step (denoted as the tuple $(I, t)$); if the system operates normally (otherwise), it outputs the empty set $\emptyset$. This pairwise formulation follows prior agent-step attribution settings, but it can be naturally extended to multi-agent, multi-step, or causal-chain attribution when failures involve multiple interacting causes.

\subsection{Failure Taxonomy}
With the growing adoption of MAS in complex tasks, researchers have increasingly recognized that failure analysis should go beyond identifying which agent or step is responsible. It must also address a more fundamental question: what type of failure has occurred and what underlying mechanism triggered it? In this sense, failure classification provides a structured descriptive framework for subsequent attribution, diagnosis, and repair~\cite{Ma2025DiagnosingFR,in2026rethinking}. Existing studies have approached this problem from three broad perspectives: system structure, execution stages, and causal lifecycle.
 
\subsubsection{System Structure}
System structure classification focuses on collaboration relationships, emphasizing the identification of where the collaborative chain breaks down. Pan et al.~\cite{pan2025whymultiagentfail} organize failures into specification and system design failures, inter-agent misalignment, and task verification/termination failures, thereby suggesting that many MAS failures can be understood as breakdowns in system design, inter-agent coordination, or verification mechanisms. AgentAsk~\cite{Lin2025AgentAskMS} further refines the classification unit to inter-agent information handoff and summarizes anomalies into four types of communication failures: Data Gap, Referential Drift, Signal Corruption, and Capability Gap. AEGIS~\cite{kong2025aegis} further transforms failure categories into controllable targets for error injection, supporting scalable generation of diagnostic data with verifiable error-identification labels.

\subsubsection{Execution Stages}
Execution process classification organizes failure types according to where they arise within the internal task workflow. This form of categorization may be instantiated either at the level of functional modules or at the level of execution stages. TRAIL\cite{deshpande2025trail} classifies agentic issues into three categories—reasoning errors, system execution errors, and planning/coordination errors—thereby placing failures in reasoning, tool invocation, and process control within a unified framework. Zhu et al.\cite{Zhu2025WhereLA} adopt a modular agent architecture and attribute failures to module-level errors involving memory, reflection, planning, action, and system-level operations. Lu et al.~\cite{lu2025exploring}  describe autonomous agent systems as a planner--generator--executor pipeline and categorize anomalies according to these three stages: task planning, task execution, and response generation. Overall, these methods interpret anomalies as instabilities arising at different points within the task execution chain, differing primarily in whether the analytical granularity is defined in terms of functional components or temporal stages.

\subsubsection{Causal Lifecycle}
Causal lifecycle classification extends failure categorization from surface-level symptoms to root-cause hierarchies and propagation processes. AgentFail~\cite{Ma2025DiagnosingFR} categorizes failure root causes at three levels: agent, workflow, and platform, covering errors from local agent execution to platform-level orchestration. FailCycle~\cite{ma2026demystifying} further points out that the point at which a failure originates often does not coincide with the point at which it becomes visible, thereby indicating that anomalies in MAS exhibit pronounced propagative and evolutionary properties. From the perspective of behavioral abstraction, AGENTSCOPE~\cite{anonymous2025diagnosing} further characterizes anomalies as behavioral patterns such as invalid context, instruction following, faulty verification, incoherent planning, and reward hacking, while integrating the classification process with structured behavioral analysis.

Overall, existing research remains substantially limited in both generalizability and granularity. Although preliminary taxonomies have been developed from diverse perspectives, the absence of a widely accepted unified framework constrains their transferability and undermines cross-study benchmarking and knowledge reuse. At the same time, most existing attribution methods still rely on a simplified single-cause--single-effect assumption~\cite{zhang2025which,Zhang2025AgenTracerWI,wang2026surveytrajectory}, making them insufficient for capturing the multi-cause--multi-effect, coupled, and cascading nature of anomalies in real-world systems. Therefore, future research should move toward a unified cross-system framework while incorporating mechanisms such as composite labels and cross-layer associations. This would enhance cross-domain transferability and enable failure taxonomies to more precisely characterize the deeper logic by which multi-agent anomalies evolve from local triggers into global system dynamics.

\subsection{Failure Attribution Methods}
MAS can improve the handling of complex tasks, yet their multi-agent coordination and long-horizon interactions also introduce new failure surfaces and cascading risks\cite{ Zhang2025AgenTracerWI,Lin2025AgentAskMS,pan2025whymultiagentfail}. In this context, multi-agent failure attribution has gradually moved from engineering experience to systematic research topics\cite{zhang2025which,Zhang2025AgenTracerWI,in2026rethinking, West2025AbductAP}. Its core is no longer just 'whether the task is successful', but to answer more operable diagnostic questions: who (which agent) introduced the decisive error at what step / when, and how the error spread into the final failure\cite{zhang2025which,Zhang2025AgenTracerWI,Zhang2025GraphTracerGF,wang2026flat,West2025AbductAP}. Around this problem, the existing research has formed three representative technical routes, which are the Data-driven method, the Constraint-guided method, and the causal-inference method. These routes are not mutually exclusive but reflect the dominant mechanism used for attribution.

\begin{table}[!htbp]
\centering
\caption{Comparison of multi-agent failure attribution methods (Section~\ref{Multi-agent failure attribution}).}
\label{tab:attribution_method}
\footnotesize
\renewcommand{\arraystretch}{1.0}
\setlength{\tabcolsep}{5pt}
\begin{adjustbox}{max width=1.05\textwidth}
\begin{tabular}{@{}L{2.60cm} C{2.00cm} C{1.80cm} C{1.80cm} C{2.40cm} C{0.90cm} C{0.90cm}@{}}
\toprule
\textbf{Method} & \textbf{Granularity} & \textbf{Data Type} & \textbf{Data Processing} & \textbf{Paradigm} & \textbf{Timing} & \textbf{Repair} \\
\midrule
\rowcolor{gray!12}
\multicolumn{7}{c}{\rowstrut\small\itshape A.\ Data-Driven Methods} \\
\midrule
\rowstrut AGENTRACER~\cite{Zhang2025AgenTracerWI}       & Agent/Step     & SL+FL       & Automatic      & \RLB            & \OfflineCap & \cmark \\
\rowcolor{zebragray}
\rowstrut Spectrum~\cite{Ge2025WhoII}                    & Agent/Step     & FL          & Automatic      & \SAB            & \OfflineCap & \na    \\
\rowstrut CORRECT~\cite{Yu2025CORRECTCE}                & Agent/Step     & SL          & Semi-auto      & \KDB            & \BothCap    & \na    \\
\rowcolor{zebragray}
\rowstrut AGENTASK~\cite{Lin2025AgentAskMS}            & Step/Category  & FL          & Semi-auto      & \SFTRLB          & \OnlineCap  & \cmark \\
\rowstrut AEGIS~\cite{kong2025aegis}                      & Agent/Category & SL          & Automatic      & \SFTRLCLB       & \OfflineCap & \na    \\
\rowcolor{zebragray}
\rowstrut GraphTracer~\cite{Zhang2025GraphTracerGF}       & Agent/Path     & SL+FL       & Automatic      & \RLB            & \OfflineCap & \cmark \\
\rowstrut MASC~\cite{Shen2025MetacognitiveSF}             & Step           & SL          & \na            & \SelfCorrB      & \OnlineCap  & \cmark \\
\rowcolor{zebragray}
\rowstrut Traj-Guard~\cite{advani2026trajectory}      & Trajectory     & SL+FL       & Semi-auto      & \SFTB           & \OnlineCap  & \na    \\
\rowstrut ProMAS~\cite{zhao2026promas}                   & Agent/Step     & Public       & \na            & \CLB            & \OnlineCap  & \na    \\
\midrule
\rowcolor{gray!12}
\multicolumn{7}{c}{\rowstrut\small\itshape B.\ Constraint-guided Methods} \\
\midrule
\rowstrut ABDUCT~\cite{West2025AbductAP}                & Agent/Step     & Public       & \na            & \JudgeB         & \OfflineCap & \na    \\
\rowcolor{zebragray}
\rowstrut AgentErrorBench~\cite{Zhu2025WhereLA}           & Step/Category  & FL          & Manual         & \JudgeB         & \OfflineCap & \cmark \\
\rowstrut Role-Trace~\cite{barrak2025traceability}       & Stage          & SL+FL       & Automatic      & \StageCompB     & \OfflineCap & \cmark \\
\rowcolor{zebragray}
\rowstrut ECHO~\cite{banerjee2025}                        & Agent/Step     & Public       & \na            & \JudgeB         & \OfflineCap & \na    \\
\rowstrut DOVER~\cite{ma2025dover}                       & Agent/Step     & FL          & Automatic      & \JudgeB         & \OfflineCap & \cmark \\
\rowcolor{zebragray}
\rowstrut AgentRx~\cite{barke2026agentrx}                & Step/Category  & FL          & Manual         & \JudgeB         & \OfflineCap & \na    \\
\rowstrut XAgen~\cite{wang2025xagen}                     & Agent/Step/WF  & FL          & Semi-auto      & \JudgeB         & \OfflineCap & \cmark \\
\rowcolor{zebragray}
\rowstrut SDBL~\cite{sun2026scope}                  & Step           & Public       & \na            & \JudgeB         & \OfflineCap & \na    \\
\midrule
\rowcolor{gray!12}
\multicolumn{7}{c}{\rowstrut\small\itshape C.\ Causal-inference Methods} \\
\midrule
\rowstrut MARL~\cite{chen2025understanding}                       & Agent/Step       & SL+FL       & Automatic      & \CounterfactualB & \OfflineCap & \cmark \\
\rowcolor{zebragray}
\rowstrut InterDebug~\cite{epperson2025interactive}          & Agent/Step       & FL          & Semi-auto      & \HITLB           & \OfflineCap & \cmark \\
\rowstrut CDC-MAS~\cite{Ma2025AutomaticFA}                      & Agent/Step       & Public       & \na            & \CausalDiscB     & \OfflineCap & \cmark \\
\rowcolor{zebragray}
\rowstrut Specific Effects~\cite{triantafyllou2023agent}        & Agent/Step       & FL          & Automatic      & \CfASEB          & \OfflineCap & \na    \\
\rowstrut AgentFail~\cite{Ma2025DiagnosingFR}                     & Agent/Category   & FL          & Manual         & \JudgeB          & \OfflineCap & \na    \\
\rowcolor{zebragray}
\rowstrut TRAJECTORY~\cite{zhengtrajectory}                & Step           & FL          & Semi-auto      & \PreActB        & \OnlineCap  & \cmark \\
\rowstrut FailCycle~\cite{ma2026demystifying}                    & Agent/Step       & FL          & Manual         & \CounterfactualB & \OfflineCap & \cmark \\
\rowstrut DiLLS~\cite{sheng2026dills}                   & Plan/Step      & FL          & Manual         & \HITLB          & \OfflineCap & \na    \\
\rowstrut MMDP-SCM~\cite{triantafyllou2024counterfactual}        & Agent/Step/State & SL+FL       & Automatic      & \TCFEB           & \OfflineCap & \na    \\
\rowcolor{zebragray}
\rowstrut AGENTSCOPE~\cite{anonymous2025diagnosing}              & Step/Category    & SL          & Automatic      & \CausalGraphB    & \BothCap    & \na    \\
\rowstrut RAFFLES~\cite{zhu2026raffles}                         & Agent/Step       & Public       & \na            & \JudgeB          & \OfflineCap & \na    \\
\rowcolor{zebragray}
\rowstrut CHIEF~\cite{wang2026flat}                             & Agent/Step       & Public       & \na            & \GraphBasedB     & \OfflineCap & \na    \\
\bottomrule
\end{tabular}
\end{adjustbox}
\vspace{3pt}
\parbox{\textwidth}{\footnotesize
\textit{Granularity}: attribution resolution --- WF = workflow. \;
\textit{Data Type}: SL = success log; FL = failure log; Public = public benchmark dataset. \;
\textit{Data Processing}: Manual = human annotation; Semi-auto = hybrid human--automatic; Automatic = fully automated. \;
\textit{Paradigm}: attribution approach --- \JudgeB\ LLM-as-judge; \RLB\ reinforcement learning; \KDB\ knowledge distillation; \SFTB\ supervised fine-tuning; \CLB\ contrastive learning; \SAB\ spectral analysis; \SelfCorrB\ metacognitive self-correction; \PreActB\ pre-acting diagnosis; \HITLB\ human-in-the-loop; \StageCompB\ stage comparison; \CounterfactualB\ counterfactual reasoning; \CfASEB\ counterfactual agent-specific effect; \TCFEB\ total counterfactual effect; \CausalDiscB\ causal discovery; \CausalGraphB\ causal graph analysis; \GraphBasedB\ graph-based analysis. Combinations (e.g.\ SFT+RL) are written inline. \;
\textit{Timing}: \OfflineCap\ offline; \OnlineCap\ online; \BothCap\ both. \;
\textit{Repair}: \cmark\ = supports downstream repair.
}
\end{table}

\subsubsection{Data-driven Method}
Data-driven methods fundamentally reframe failure attribution in MAS from the manual inspection of raw logs to the learning, representation, and generalization of recurring anomalous patterns encoded in failure trajectories. Early approaches primarily relied on human analysis or on LLM directly processing long trajectories for responsibility localization\cite{zhang2025which}. More recent research has progressively shifted toward dedicated attribution models, structured trajectory representations, and lightweight online diagnostic modules\cite{Zhang2025AgenTracerWI,kong2025aegis}. As a result, the central challenge in this line of work is how to distill long-horizon, tightly coupled failure processes that propagate across agents into attribution representations that are learnable, transferable, and reusable\cite{pan2025whymultiagentfail,Yu2025CORRECTCE,sheng2026dills}.

From the perspective of methodological evolution, data-driven attribution began with the explicit formalization of failure localization and the direct application of models to execution trajectories for attribution. Who\&When\cite {zhang2025which} is representative of this early paradigm. It formulates attribution as the automatic identification of both the responsible agent and the critical point at which the error is introduced, while comparing the effectiveness of alternative trajectory search and prompting strategies. In the same vein, both TRAIL\cite{deshpande2025trail} and TraceElephant\cite{anonymous2026seeing} show that general-purpose long-context models remain limited in debugging complex execution trajectories and that incomplete observability further undermines attribution quality. Collectively, these findings shifted subsequent research away from treating generic large language models as log readers and toward the development of more learnable trajectory representations and specialized attribution models.

In the existing literature, the most representative data-driven paradigm is the use of large-scale labeled data to train dedicated attribution models. AGENTRACER\cite {Zhang2025AgenTracerWI} follows this route by generating attribution-labeled failure trajectories through counterfactual replay and programmatic error injection, and by further training a lightweight tracer, thereby shifting attribution from generic reasoning to specialized model learning. Aegis\cite{kong2025aegis} systematizes this idea further by using context-aware error injection to construct supervisory signals and by supporting three training paradigms: supervised fine-tuning, reinforcement learning, and contrastive learning. Although CORRECT\cite {Yu2025CORRECTCE} does not itself rely on retraining, its method also relies on the abstraction of real failure distributions and the reuse of structured error signals. This suggests that the core of data-driven methods lies in transforming failure phenomena that are scarce, fragmented, and difficult to reproduce into supervisory signals that can be learned robustly.

At the modeling level, existing research has shifted markedly from raw textual interpretation to structured attribution representations. Spectrum-analysis-based methods~\cite{Ge2025WhoII} draw on software fault localization by replacing direct semantic judgments over individual logs with cross-trajectory coverage statistics and suspiciousness scoring, thereby grounding attribution in the statistical differences observed across repeated executions. CORRECT\cite {Yu2025CORRECTCE} proposes a more lightweight alternative based on the assumption that failures in MAS exhibit recurring structural patterns. Instead of reanalyzing the full trajectory from scratch for each case, it performs pattern transfer and targeted localization at inference time through an online cache of distilled error schemas. GraphTracer\cite {Zhang2025GraphTracerGF} further shows that analyzing action chains solely in temporal order is insufficient to recover the true pathways of error propagation, and therefore models cross-agent information flow through an information dependency graph.

Beyond strict post hoc attribution, recent data-driven failure analysis has also incorporated online detection, intervention, and human-in-the-loop diagnostic support. AgentAsk\cite{Lin2025AgentAskMS} treats failure propagation as a localized risk arising in message interactions and learns when to initiate clarification, which agent to query, and how to formulate the minimal necessary clarification, thereby enabling intervention before errors spread. Trajectory Graph Copilot\cite{zhengtrajectory} advances this direction by constructing probabilistic graphs from historical trajectories and leveraging graph neural networks to issue risk alerts before actions are executed. Trajectory Guard\cite{advani2026trajectory} further adopts a sequence-aware autoencoding framework for real-time failure detection, with an emphasis on low-latency identification of implausible trajectories. Although DiLLS\cite{sheng2026dills} is more oriented toward interactive diagnosis, its core principle is similar: multi-agent behaviors are compressed into hierarchical representations that are queryable and comparable, thereby improving developers' efficiency in locating and understanding failures. Collectively, these studies show that the value of data-driven methods is no longer confined to offline responsibility attribution, but is increasingly shifting toward continuous monitoring of failure propagation, preemptive interruption, and human-in-the-loop analysis.

Overall, data-driven research on failure attribution in MAS has progressed from prompt-based direct localization to a stage of systematic modeling built on automatic supervision, structured representations, and dedicated attribution models. Current evidence suggests that stable attribution cannot be achieved reliably by relying solely on general-purpose models to read long trajectories. A more effective direction is to organize attribution signals around the reuse of failure patterns, cross-trajectory structural modeling, and mechanisms for online intervention.

\subsubsection{Constraint-Guided Method}
Constraint-guided methods formulate failure attribution as an explicit diagnostic pipeline, rather than requiring a model to infer responsibility directly from a complete execution trajectory\cite{sun2026scope,West2025AbductAP}. Specifically, they decompose the attribution process into a series of executable and verifiable subtasks through stepwise decomposition, scope delimitation, error taxonomy design, responsibility boundary specification, and result validation\cite{ma2025dover,barrak2025traceability}. Existing studies indicate that these approaches are primarily oriented toward two objectives. First, by imposing structured diagnostic procedures, they constrain the attribution search space and thus reduce the reasoning burden induced by long-horizon trajectories and complex multi-role interactions. Second, by grounding attribution in explicit evidentiary criteria and validation mechanisms, they improve the interpretability and auditability of root-cause identification.

One important line of research explores how constraint-based designs can restructure the search space of failure attribution. To cope with the long execution trajectories, complex error propagation chains, and unstable global localization commonly observed in MAS, existing studies have generally adopted staged diagnostic frameworks. Specifically, Scope Delineation Before Localization (SDBL)~\cite{sun2026scope} formulates attribution as a two-stage process in which the failure scope is first delimited and the critical step is subsequently localized, thereby reducing interference from irrelevant context. Pan et al.~\cite{pan2025whymultiagentfail} ground their attribution in hierarchical evidence extraction based on the systematic identification and analysis of fault modes, thereby underscoring the urgent need for a standardized diagnostic process. A2P\cite{West2025AbductAP} further constrains the attribution procedure into three sequential stages: causal hypothesis generation, minimal correction, and outcome prediction. Rather than relying solely on temporal proximity, A2P uses counterfactual correction and outcome prediction to assess whether a candidate step is causally decisive. The aforementioned studies first narrow the attribution space through predefined rules and then organize the reasoning procedure accordingly, thereby transforming long-horizon and open-ended failure localization into a constrained local diagnostic task.

Another important line of research focuses on how rules can be used to define root causes and validate attribution conclusions. Zhu et al.\cite{Zhu2025WhereLA}  decompose failures through a modular error taxonomy into critical errors associated with different capability components and further emphasize the need to isolate root-cause failures that may propagate through subsequent decisions. DoVer~\cite{ma2025dover} argues that log-based localization should be treated as an unverified hypothesis and introduces targeted intervention, replay, and outcome comparison to validate or refute candidate failure hypotheses. In a similar vein, Role-Trace\cite{barrak2025traceability} improves the traceability of error sources and their cross-role propagation paths through structured handoff mechanisms and the preservation of intermediate records. AgentRx\cite{barke2026agentrx} grounds attribution in explicit evidence by automatically generating constraints, checking violations in a stepwise manner, and producing auditable logs. Although XAgen\cite{wang2025xagen} is primarily an explainability and debugging tool, its failure identification process also incorporates predefined objectives, structured evaluation criteria, and a standardized assessment workflow. The aforementioned studies move beyond simple responsibility localization and place greater emphasis on the evidentiary grounding and verifiability of root-cause determination.

Overall, Constraint-guided methods have evolved from early heuristic error localization into structured attribution frameworks that integrate scope delimitation, hierarchical analysis, and intervention-based validation. Their key advantage lies in the ability to explicitly regulate the attribution process and to ground root-cause judgments in evidence that is transparent and verifiable. However, existing methods still depend heavily on task-specific priors, manually designed procedural constraints, and domain-dependent adjudication criteria, which limit the generalizability and adaptability of the underlying rules in turn.

\subsubsection{Causal-inference Method}
Causal-inference-based research on multi-agent failure attribution aims to identify the causal relationships within execution trajectories that induce both the onset and propagation of failures, rather than merely localizing entities whose behaviors co-occur with the final failure outcome~\cite{triantafyllou2023agent,Ma2025AutomaticFA,wang2026flat}. Specifically, it seeks to determine—through counterfactual replay, intervention analysis, and related causal techniques—which critical behaviors, interaction dependencies, or state transitions genuinely trigger anomalies. Furthermore, it characterizes the mechanisms through which such anomalies propagate along collaborative interaction chains and through the evolution of the environment~\cite{triantafyllou2024counterfactual,Ma2025AutomaticFA}.

The first line of research provides a theoretical foundation for causal attribution in multi-agent sequential decision-making, which later informs causal reasoning about LLM-based MAS failures. Agent-Specific Effects~\cite{triantafyllou2023agent} shows that, in multi-agent Markov decision processes, the impact of an individual agent's action on the outcome is often not localized, but instead propagates through the subsequent responses of other agents. Consequently, attribution should not be reduced to a simple mapping between a local action and an observed outcome; rather, it should capture the causal influence of an action as it unfolds through inter-agent propagation. The study further discusses how such propagated effects may be identified and estimated under counterfactual settings. Extending this view, Triantafyllou et al.~\cite{triantafyllou2024counterfactual} decomposes the total counterfactual effect of an action into two components: one transmitted through the downstream behaviors of other agents, and the other through environmental state transitions. This decomposition enables disentangling the respective contributions of collaborative interaction chains and environmental dynamics within a unified causal framework. In contrast, Chen et al.~\cite{chen2025understanding} focus primarily on individual importance estimation. By leveraging action randomization and masking mechanisms, it quantifies the extent to which a single agent influences variations in team reward, thereby identifying critical agents. However, its emphasis remains on explaining individual contributions, rather than modeling the complete propagation chain through which failures emerge and spread.

The second line of work extends causal inference into the real execution trajectories of LLM-based MAS. Its primary concern is no longer the estimation of causal effects over a predefined structure, but the recovery of attribution-relevant causal relationships from complex execution logs. Ma et al.~\cite{Ma2025AutomaticFA} explicitly propose a multi-granularity causal attribution framework, in which responsible agents are localized through performance causal inversion and Shapley value analysis, while critical failure steps are identified through causal discovery methods. Wang et al.~\cite{wang2026flat} further argue that treating execution logs as flat sequences obscures the true causal chain; accordingly, it reconstructs trajectories into hierarchical causal graphs and distinguishes genuine root causes from propagated symptoms through hierarchical backtracking and progressive causal screening. AgentTrace~\cite{wang2026agenttrace} follows a similar rationale by recovering, from execution logs, a causal graph composed of sequential, communication, and data dependencies. It then traces backward from the point at which an error becomes manifest to candidate causes and ranks them according to structural and positional features, thereby supporting lightweight post-hoc root-cause analysis.

Beyond strict causal inference, several causal-adjacent studies improve the analyzability of long trajectories through structured representations, interactive interventions, and graph-based diagnosis. Bi et al.~\cite{anonymous2025diagnosing} transform long trajectories into structured representations through behavioral abstractions and neural invariants, whereas DiLLS~\cite{sheng2026dills} further organizes system behavior into a three-level abstraction consisting of activities, actions, and operations. In both cases, raw logs are compressed into intermediate representations that are more conducive to failure-action localization and root-cause analysis. Moving beyond representation alone, Ma et al.~\cite{ma2026demystifying} extend the analysis to the propagation process, examining how anomalies evolve across nodes through natural-language interactions, tool invocations, and control logic in platform-orchestrated workflows. Zheng et al.~\cite{zhengtrajectory} shifts diagnosis to the pre-action stage by modeling historical trajectories as probabilistic graphs and identifying high-risk action patterns. In parallel, Epperson et al.~\cite{epperson2025interactive} and Zhu et al.~\cite{zhu2026raffles} support the progressive verification of critical fault points through interactive message editing and reset operations, and through iterative reasoning-based investigation, respectively. Finally, Le et al.~\cite{le2024multi} demonstrate that causal graphs can be constructed by combining collaborative reasoning, code execution, and statistical methods, thereby offering methodological insight into how attribution can be grounded in the identification of those behaviors and interactions that genuinely drive failure emergence in complex agent trajectories. Collectively, these studies complement causal attribution by providing structured diagnostic abstractions and verification mechanisms.

Overall, causal-inference-based research on multi-agent failure attribution has developed into an increasingly coherent line of inquiry, encompassing both the formal characterization of causal effects and the recovery of causal relationships from execution logs for failure analysis. Compared with approaches that localize responsibility primarily through correlation, this line of work places greater emphasis on explaining why anomalies emerge, how they propagate, and which factors along the causal chain genuinely drive the eventual failure.

\subsection{Failure Attribution Evaluation}
As shown in Table~\ref{tab:attribution_dataset}, research on datasets for failure attribution in MAS has evolved from the early practice of manually annotating failure logs into a more systematic line of inquiry centered on data realism, annotation granularity, process observability, and evaluation validity. This evolution is reflected in three major trends. First, dataset construction has progressed from small-scale manual curation to automated large-scale generation~\cite{zhang2025which,deshpande2025trail,Zhang2025AgenTracerWI,kong2025aegis}. Second, annotation targets have expanded beyond responsible agents and critical steps to encompass multiple attribution dimensions, including error types, interaction edges, functional modules, and root causes in platform-orchestrated workflows~\cite{Lin2025AgentAskMS,deshpande2025trail,Zhu2025WhereLA,Ma2025DiagnosingFR}. Third, evaluation has moved beyond single-metric localization accuracy toward composite frameworks that assess attribution performance, explanation quality, and process faithfulness in a unified manner~\cite{in2026rethinking,anonymous2026seeing}.

\begin{figure}    
\centering    
\includegraphics[width=0.95\linewidth]{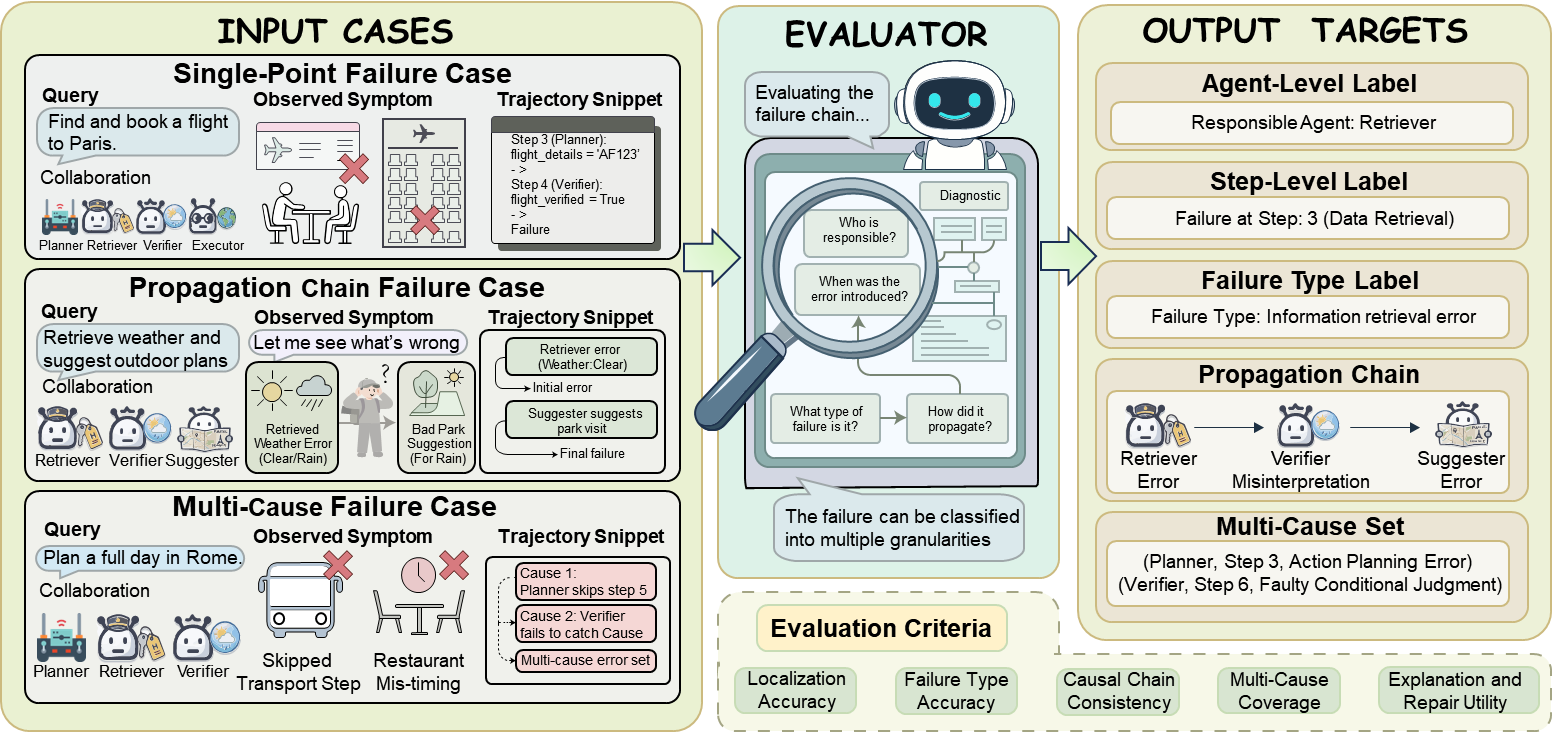}    
\caption{Scenario illustration of anomaly attribution evaluation in multi-agent systems. It presents three input case types---single-point, propagation-chain, and multi-cause failures---together with the corresponding multi-granularity attribution targets and evaluation criteria.}    
\label{fig:attribution-case}
\end{figure}

\begin{table}[!htbp]
\centering
\caption{Comparison of multi-agent failure attribution datasets.}
\label{tab:attribution_dataset}
\footnotesize
\renewcommand{\arraystretch}{1.25}
\setlength{\tabcolsep}{5pt}
\begin{adjustbox}{max width=1.05\textwidth}
\begin{tabular}{@{}L{2.80cm} C{2.00cm} C{0.80cm} C{1.40cm} C{2.50cm} C{1.00cm} C{1.40cm} C{0.80cm}@{}}
\toprule
\textbf{Dataset} & \textbf{Focus} & \textbf{Size} & \textbf{Annotation} & \textbf{Granularity} & \makecell{\textbf{Data}\\\textbf{Source}} & \textbf{Metrics} & \textbf{Open} \\
\midrule
\rowstrut Who\&When~\cite{zhang2025which}                & \LocalizationB & 184   & Manual         & Agent+Step+Interp. & FL    & Accuracy   & \cmark \\
\rowcolor{zebragray}
\rowstrut TRAIL~\cite{deshpande2025trail}                & \CategoryB     & 148   & Manual         & Step+Type          & FL    & Composite  & \cmark \\
\rowstrut AgentFail~\cite{Ma2025DiagnosingFR}            & \CategoryB     & 307   & Manual         & Agent+Category     & FL    & Accuracy   & \na    \\
\rowcolor{zebragray}
\rowstrut AgentErrorBench~\cite{Zhu2025WhereLA}          & \CategoryB     & 200   & Manual         & Category+Feedback  & FL    & Accuracy   & \cmark \\
\rowstrut CORRECT-Error~\cite{Yu2025CORRECTCE}           & \LocalizationB & 2000+ & Synthetic      & Step               & SL    & Accuracy   & \na    \\
\rowcolor{zebragray}
\rowstrut AgentErrata~\cite{anonymous2025diagnosing}     & \CategoryB     & 210   & Synthetic      & Agent+Category     & SL    & Accuracy   & \na    \\
\rowstrut MP-Bench~\cite{in2026rethinking}               & \LocalizationB & 289   & Manual         & Multi-Step         & FL    & Composite  & \na    \\
\rowcolor{zebragray}
\rowstrut TraceElephant~\cite{anonymous2026seeing}       & \LocalizationB & 220   & Manual         & Agent+Step         & FL    & Accuracy   & \cmark \\
\bottomrule
\end{tabular}
\end{adjustbox}
\vspace{3pt}
\parbox{\textwidth}{\footnotesize
\textit{Focus}: attribution objective --- \LocalizationB\ localization-oriented; \CategoryB\ category-level. \;
\textit{Annotation}: how ground-truth labels are obtained --- Manual = human annotation; Synthetic = programmatic error injection. \;
\textit{Granularity}: resolution of attribution labels --- Interp.\ = interpretation. \;
\textit{Data Source}: FL = failure log; SL = success log. \;
\textit{Metrics}: Accuracy = ground-truth-based accuracy; Composite = multi-metric joint evaluation. \;
\textit{Open}: \cmark\ = publicly available.
}
\end{table}

A representative line of research has centered on constructing high-fidelity, manually curated datasets that serve as reliable evaluation anchors. Who\&When~\cite{zhang2025which} first formalized failure attribution in MAS as a joint localization problem over the responsible agent and the critical step, thereby establishing the basic organizational paradigm for subsequent dataset development. Building on this foundation, TRAIL~\cite{deshpande2025trail} advanced the research focus toward fine-grained error analysis in long-horizon trajectories and refined the structural representation paradigm of fine-grained attribution through explicit error categorization. AgentFail~\cite{Ma2025DiagnosingFR} further extended the data scope to platform-orchestrated agentic workflows, so that the resulting dataset captures not only execution logs but also explicit links between workflow configurations and root-cause categories, thereby elevating attribution from single interactions to the system orchestration layer. Meanwhile, TraceElephant~\cite{anonymous2026seeing} underscored the importance of end-to-end observability and reproducible experimental environments, showing that incomplete observations can systematically bias attribution judgments. Consequently, the observational fidelity of the dataset itself has emerged as a critical prerequisite for failure attribution research.

Another major line of research addresses the limited scalability of manual annotation by constructing larger training corpora through controlled data generation. Representative methods, including AGENTRACER~\cite{Zhang2025AgenTracerWI} and AEGIS~\cite{kong2025aegis}, leverage counterfactual replay and programmatic error injection to produce failure labels that are traceable and reproducible, thereby enabling automatically annotated training data for attribution. The CORRECT-Error~\cite{Yu2025CORRECTCE} dataset introduced in CORRECT further emphasizes generation strategies grounded in real failure distributions and improves the alignment between synthetic samples and naturally occurring failure patterns through human verification. GraphTracer~\cite{Zhang2025GraphTracerGF} advances this paradigm by explicitly incorporating cross-agent information dependencies into the generation process, allowing synthetic data to preserve not only temporal order but also the structural pathways of information propagation. In parallel, AgentErrata~\cite{anonymous2025diagnosing} pushes dataset construction beyond simple responsibility localization toward finer-grained diagnosis of failure patterns through behavioral abstraction and structured annotation. Overall, automated data construction has become a principal approach to alleviating data scarcity, although its key challenge lies in balancing scalability with faithful preservation of real failure characteristics.

Recent dataset research has continued to redefine the attribution target itself. Rather than remaining limited to identifying which agent made an error at a particular moment, recent studies have progressively introduced richer annotation dimensions, such as interaction edges and functional modules~\cite{Lin2025AgentAskMS,in2026rethinking}. Correspondingly, evaluation methodologies have also undergone a parallel shift. When attribution targets can still be expressed as relatively well-defined discrete labels, accuracy remains a valid metric. However, once datasets explicitly capture multiple plausible attribution paths, expert disagreement, and explanation quality, evaluation must be extended to include composite measures of ranking quality, reasoning consistency, and attribution plausibility \cite{Ma2025DiagnosingFR,in2026rethinking}. This shift indicates that the focus of dataset research for failure attribution in MAS is moving beyond sample construction alone toward the co-evolution of task definition and evaluation protocols.

Overall, research on datasets for failure attribution in MAS has progressed from early task-oriented exploration to a more systematic stage in which data construction, annotation representation, and evaluation methodology are jointly developed. The central advance of this line of research lies not in the mere expansion of sample scale, but in the continued refinement of attribution target definition, the representation of real failure propagation, and the design of evaluation principles under multi-perspective settings. Despite this progress, several common bottlenecks remain, including the limited availability of high-quality open datasets, the difficulty of reconciling annotation fidelity with scalability, and the lack of unified standards for attribution granularity and evaluation protocols.

\subsection{Discussion}
Although multi-agent failure attribution has made preliminary progress in task definition, dataset construction, and method design, it still faces several key challenges. Recent survey on LLM-agent trajectory analysis have identified several broad challenges, including limited step-level attribution accuracy, insufficient benchmark diversity, incomplete observability, and the need for repair-oriented evaluation~\cite{wang2026surveytrajectory}. Building on these survey-level findings, this discussion focuses more narrowly on multi-agent anomaly attribution. We argue that the central difficulty is not merely how to localize a failure, but how to systematically integrate five key elements that are often studied in isolation: attribution semantics, failure propagation, diagnostic granularity, data realism, and repair mechanisms.

\subsubsection{Core Challenges}
\paragraph{Attribution Definition Remains Unsettled.}
A foundational challenge in multi-agent anomaly attribution research is that the attribution target itself has yet to be defined in a unified manner. Existing studies variously formulate attribution as identifying responsible agents and critical steps, fine-grained error types, interaction-level failures, platform-level workflow root causes, or responsibility objects from multiple perspectives, making the term attribution semantically inconsistent across the literature \cite{zhang2025which,deshpande2025trail,in2026rethinking}. Although these works all appear to address anomaly attribution, they often differ substantially in input granularity, annotation targets, reasoning objectives, and evaluation criteria. Consequently, method-level performance is difficult to compare directly, dataset annotation schemes are hard to align, and further progress in model transfer, data reuse, and benchmark unification is constrained. More importantly, when the failure manifestation, the causal trigger and the responsibility bearer are not explicitly distinguished, existing studies may conflate the earliest point at which an anomaly is exposed, the factor that actually triggers it, and the agent or component that should bear primary responsibility. This conceptual ambiguity, in turn, weakens the interpretability and credibility of attribution results.

\paragraph{Anomaly Propagation Is Strongly Coupled.}
In multi-agent systems, anomalies rarely manifest as the direct consequence of a single error at a single step. Instead, they more often emerge from local deviations that become progressively coupled, propagated, and amplified through ongoing interactions \cite{pan2025whymultiagentfail,Ma2025DiagnosingFR,ma2026demystifying}. A minor deviation introduced early in execution may initially appear as inconsistent information interpretation, then spread through task allocation, tool invocation, and result aggregation, and ultimately surface at a later node that appears only loosely connected to the original source. Consequently, the point at which an anomaly becomes observable often does not coincide with its true origin, and final failure is frequently the product of multiple interacting stages rather than a single isolated error \cite{Ma2025DiagnosingFR,ma2026demystifying,Lin2025AgentAskMS}. However, many existing task formulations still reduce attribution to identifying a responsible agent, a critical step, or a single candidate cause, while providing only limited treatment of how anomalies propagate across agents, modules, and execution stages \cite{zhang2025which,deshpande2025trail,wang2026surveytrajectory}. This limitation makes current approaches prone to mistaking downstream symptoms for root causes when handling compound failures, delayed exposure, and cascading breakdowns, and to overlooking the joint effect of multiple local errors.

\paragraph{Fine-Grained Attribution Remains Limited.}
Although existing research has made meaningful progress in coarse-grained attribution at the agent level, substantial bottlenecks remain once attribution is pushed to the level of steps, modules, error types, or full trajectories \cite{zhang2025which,deshpande2025trail,wang2026surveytrajectory}. The difficulty is fundamental: fine-grained attribution requires not only identifying where a problem occurs, but also determining, under long execution traces and complex dependencies, what type of problem it is, whether it constitutes a root cause, a contributing trigger, or a propagated symptom, and how it spreads through the system. In realistic multi-turn collaborative settings, many intermediate actions are highly context-dependent, making their anomalous nature difficult to assess from an isolated step alone. However, once longer context must be incorporated, the analysis becomes increasingly susceptible to information redundancy, noise accumulation, and the attenuation of long-range dependencies \cite{deshpande2025trail,anonymous2026seeing,wang2026surveytrajectory}. Therefore, the core limitation of current fine-grained attribution is not merely the use of finer labels, but the lack of robust methods capable of handling long-horizon, multi-role, and multi-level dependency structures. As a result, attribution at finer granularities often remains insufficiently reliable and therefore cannot effectively support precise downstream repair.

\paragraph{Data Realism Remains Insufficient.}
Dataset limitations remain a major bottleneck in current research on anomaly attribution for multi-agent systems. Although recent studies have progressively shifted from small-scale manual annotation toward automated construction, programmatic fault injection, and counterfactual generation, larger data volume does not in itself resolve the problem of data realism \cite{Zhang2025AgenTracerWI,kong2025aegis,wang2026surveytrajectory}. Many automatically generated failure samples still rely on predefined templates, controllable perturbations, or rule-based injection, resulting in anomaly patterns that are relatively regular and clearly bounded. These samples therefore remain distributionally different from the complex failures encountered in real deployments, where anomalies are often jointly shaped by missing context, cross-module interference, unstable external tools, and deviations accumulated across multi-turn collaboration \cite{Yu2025CORRECTCE,Zhang2025GraphTracerGF,anonymous2026seeing}. This mismatch further biases the learning objective of attribution models, making them more likely to adapt to engineered anomalies than to those that arise naturally in practice. At the same time, although manual high-fidelity annotation more closely reflects real scenarios, it is costly, difficult to scale, and often accompanied by annotator disagreement over responsibility assignment in complex failure cases \cite{anonymous2026seeing,in2026rethinking,wang2026surveytrajectory}. Therefore, the core data challenge is not simply the lack of more samples, but the absence of high-quality attribution resources that simultaneously satisfy realism, scalability, and consistency.

\paragraph{The Evaluation--Repair Loop Remains Incomplete.}
Existing research on anomaly attribution in multi-agent systems still evaluates methods primarily in terms of localization accuracy, that is, whether the correct responsible agent, critical step, or candidate cause can be identified \cite{zhang2025which,deshpande2025trail,wang2026surveytrajectory}. Although such metrics were essential in the early development of the field, they are no longer sufficient as attribution tasks increasingly involve multi-cause coupling, cross-layer propagation, and multi-granularity analysis \cite{in2026rethinking,wang2026surveytrajectory,deshpande2025trail}. A more consequential limitation is that attribution evaluation remains only weakly connected to downstream verification, intervention, and repair. As a result, many methods can generate superficially plausible attribution results, yet still fail to demonstrate whether these results are sufficiently reliable to support practical decision-making and system improvement \cite{wang2026surveytrajectory,Ma2025DiagnosingFR,ma2026demystifying}. Consequently, current research on multi-agent anomaly attribution still operates largely at the level of analysis and localization, without yet establishing a robust closed loop from failure identification and root-cause confirmation to repair feedback.

Overall, research on anomaly attribution in multi-agent systems has evolved from the early coarse localization of failures into a systematic problem involving task definition, propagation mechanisms, fine-grained modeling, data construction, and evaluation protocols. However, although existing studies can increasingly answer where a problem occurs, a substantial gap remains before they can reliably answer what the problem is, why it occurs, how the diagnosis should be validated, and whether it can support effective repair.

\subsubsection{Future Research Direction}
In the future, failure attribution in MAS should evolve from merely being able to localize faults toward being able to explain, verify, and support repair.
\begin{itemize}
\item A unified and extensible framework for failure classification and attribution is needed to support alignment across heterogeneous system settings and label granularities, while more effectively capturing complex phenomena such as multi-cause, multi-effect, and failure propagation.
\item It is necessary to build a larger-scale, higher authenticity, and more granular attribution data set. While using automatic generation to reduce costs, manual verification and real log introduction should be strengthened to improve data quality and scene adaptability. 
\item Future attribution methods should adopt more structured modeling strategies that explicitly represent message passing, information dependencies, behavioral evolution, and failure propagation chains, thereby improving fine-grained attribution in long-horizon trajectories.
\item Causal inference and multi-cause and multi-result modeling should be further strengthened, from single responsibility point identification to responsibility set, contribution sharing and counterfactual propagation analysis. 
\item It is also necessary to establish a closed-loop pipeline of attribution, verification, and repair, together with a multidimensional evaluation framework, so that failure attribution can provide not only localization results but also trustworthy and verifiable diagnostic evidence that can be translated into practical system repair gains.
\end{itemize}


\section{Multi-Agent System Self-Evolution}
\label{sec:self-evolution}

Prior research has demonstrated that the most resilient systems are not rigidly designed top-down but instead evolve continuously from the bottom up~\cite{parunak1997go, bonabeau1999swarm, serugendo2006self}. Through basic mechanisms such as mutation and selection, biological populations adapt to survive in unpredictable environments. Inspired by this process, this section explores how MAS can shift from statically configured, human-engineered frameworks to dynamically adaptive, self-evolving systems. Rather than relying on fixed rules, a self-evolving MAS allows agents to continuously update their behaviour, rewire their communication networks, and build entirely new organizational structures over time. In the following subsections, we outline the motivation for this shift, formalize the evolutionary process, categorize existing methods, and discuss the open challenges ahead.

\subsection{From Attribution to Evolution}
\label{subsec:attribution-to-evolution}

As discussed in Section~\ref{Multi-agent failure attribution}, multi-agent 
anomaly attribution is crucial for understanding system failures. However, 
accurate diagnosis alone does not improve the system. A autonomous MAS 
must bridge the gap between identifying errors and mitigating them. To achieve 
this, we draw inspiration from biological adaptation. In nature, environmental 
pressures and failures trigger trial-and-error 
exploration~\cite{galhardo2007mutation}. Similarly, an autonomous MAS should 
treat recognized anomalies not just as static logs, but as direct triggers to 
actively search for optimized collaborative states.

The main difference between attribution and evolution is their time focus. 
Attribution is retrospective: it analyzes past executions to find the root 
cause, answering ``Why did the failure occur'' and ``Who is responsible?''. 
Conversely, self-evolution is prospective: it looks to the future, asking 
``How can the system perform better?'' and ``How should it adapt to unseen 
environments?''. This difference is clearly seen in biological swarm 
intelligence. For example, when foraging ants encounter a blocked path, the 
fading of pheromones serves as a retrospective attribution of 
failure~\cite{goss1989self}. Instead of stopping, the colony uses this past 
feedback to prospectively adapt its route, actively exploring to bypass the 
obstacle.

Despite their different time focuses, attribution and evolution are deeply 
connected in the continuous learning loop of an MAS. Self-evolution cannot 
operate blindly; random structural changes or arbitrary prompt adjustments are 
computationally inefficient and prone to failure. Instead, error attribution 
provides the exact context needed for targeted evolution. By pinpointing the 
root cause of a failure, attribution narrows down the evolutionary search 
space. It directly guides the system to systematically rewire its 
organizational topologies and refine the behavioral policies of individual 
agents.

Indeed, recent advancements in multi-agent collaboration have introduced 
highly automated mechanisms~\cite{wang2024survey,hong2024metagpt,
wu2024autogen}, such as dynamic role allocation and trajectory refinement, 
that already exhibit early forms of adaptive behavior. Yet as tasks 
increasingly move into open-ended, highly unpredictable environments, these 
systems must adapt beyond static, human-designed 
boundaries~\cite{leong2025amas}. Specifically, existing paradigms face two 
primary bottlenecks that prevent them from achieving true self-evolution.

\paragraph{Architectural Rigidity}
Although modern MAS allow for runtime dynamic role allocation, their 
underlying orchestration logic and interaction protocols $I_\pi$ are heavily 
constrained by human design~\cite{chen2024agentverse,wang2024survey,
leong2025amas}. Current optimization methods primarily focus on maximizing 
trajectory utility $J(\tau)$ within a fixed interaction framework. However, 
when the optimal trajectory $\tau^*$ under these constraints still fails to 
solve a complex task, real-world deployments frequently encounter unseen edge 
cases and severe coordination failures in which predefined adaptive rules 
prove insufficient. Current systems cannot fundamentally change and evolve 
their structural connectivity $\mathcal{G}$ from scratch. When a sub-task 
fails or a bottleneck emerges, they can only adjust within a limited scope, 
rather than autonomously inventing entirely new collaboration structures to 
bypass failures and adapt to open-world complexities.

\paragraph{Scalability Bottlenecks}
As task complexity demands a larger and more diverse set of specialized 
agents, the search space for orchestration topology $\mathcal{G}$ and 
collaboration policies $\Pi$ grows exponentially. Manual orchestration and 
rule-based scaling quickly reach their limits. Communication overhead and the 
risk of interference grow non-linearly~\cite{kim2025towards}. Without an 
evolutionary mechanism to iteratively evaluate, modify, and remove inefficient 
communication links, expanding a MAS often leads to diminishing returns. 
These challenges have motivated the exploration of self-improving evolutionary 
algorithms that can sustainably manage collective intelligence at scale.

To break through these bottlenecks, researchers have begun to explore 
Multi-Agent System Evolution~\cite{guo2024large,xiang2026systematic,
hu2025automated}. Inspired by biological evolution, this paradigm moves beyond 
merely executing collaborative policies to autonomously modifying, 
recombining, and selecting agent configurations over time.

\subsection{Formal Definition of MAS Self-Evolution}

\begin{figure}
    \centering
    \includegraphics[width=\linewidth]{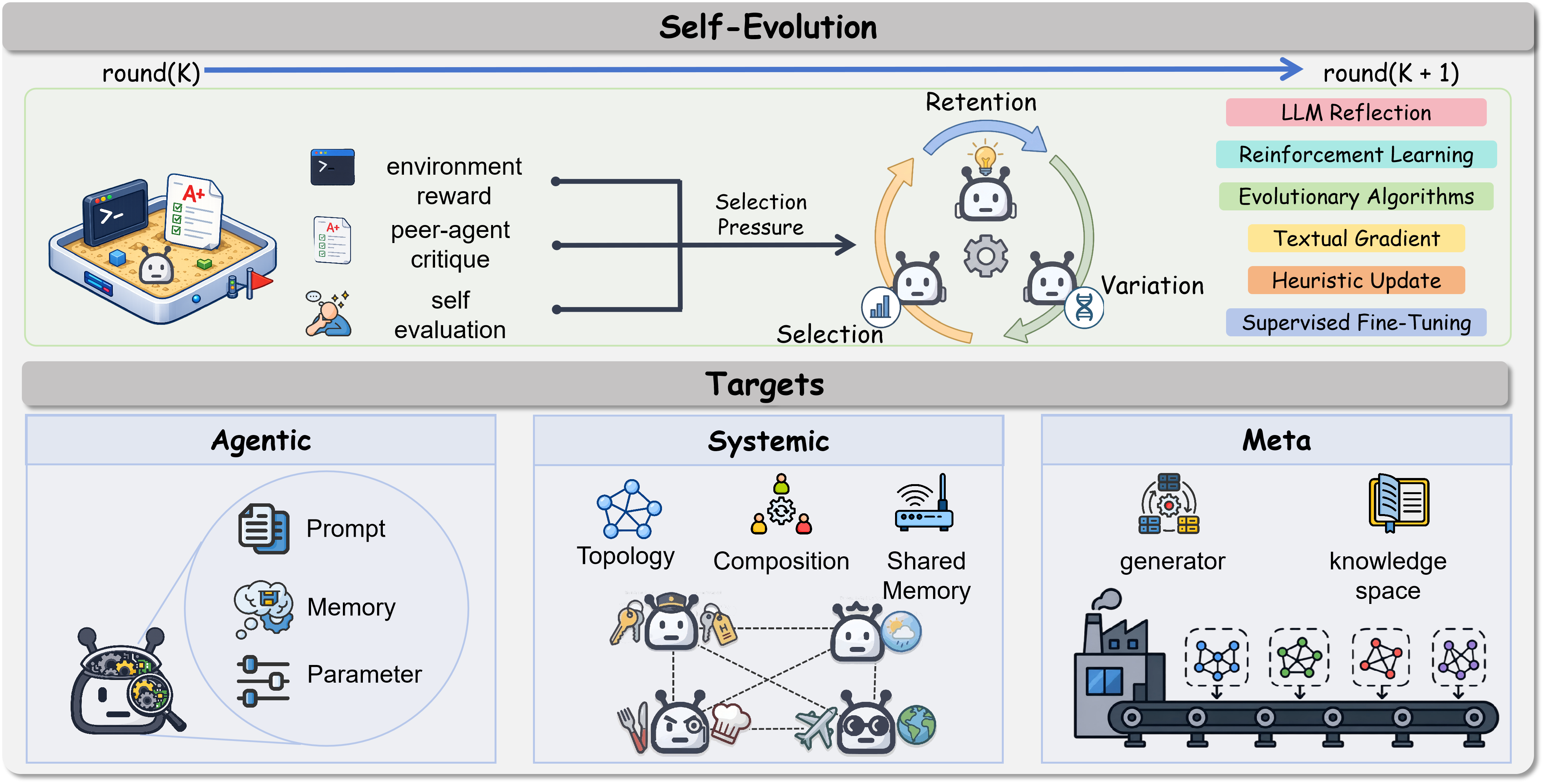}
    \caption{Overview of the multi-agent self-evolution framework. The top panel details the continuous self-evolution cycle driven by variation, selection, and retention, incorporating environmental rewards, peer-agent critiques, and self-evaluation as selection pressures. The bottom panel categorizes the evolutionary targets into three hierarchical levels: Agentic level, Systemic level, and Meta level.
    }
    \label{fig:evolution}
\end{figure}

\label{subsec:evolution-definition}
To formally study self-evolving MAS, we extend the static collaboration framework defined in Section~3 into a discrete-time process indexed by generation $k \in \mathbb{N}$. The self-evolving MAS $S$ is thus defined as a time-varying tuple:
\begin{equation}
S^{(k)}
=
(\mathcal{A}^{(k)},\; \mathcal{E}^{(k)},\; \mathcal{C}^{(k)},\; \mathcal{G}^{(k)},\; \Pi^{(k)}),
\end{equation}
where
\begin{itemize}
    \item $\mathcal{A}^{(k)} = \{a_1^{(k)}, \ldots, a_{N_k}^{(k)}\}$ is the set of agents at generation $k$;
    \item $\mathcal{E}^{(k)}$ denotes the environment, which may itself shift across generations as the system encounters new task distributions or deployment conditions;
    \item $\mathcal{C}^{(k)}$ is the communication protocol that governs message exchange at generation $k$;
    \item $\mathcal{G}^{(k)}$ denotes the orchestration topology describing the architecture of the MAS at generation $k$;
    \item $\Pi^{(k)}$ denotes the set of collaboration policies at generation $k$.
\end{itemize}
 To isolate the system's intrinsic improvement from environmental variation, the utility objective in Eq.~\ref{equal15} averages over a task distribution $\mathcal{D}_{\mathrm{task}}$ rather than conditioning on a single fixed scenario. The core of this dynamic framework is the self-evolution mechanism, formalized as a transition mapping $\Gamma$. Rather than relying on random changes, this mapping autonomously generates the next-generation system based on historical context:
\begin{equation}
S^{(k+1)} = \Gamma \left( S^{(k)}, H^{(k)} \right),
\end{equation}
where $H^{(k)}$ encapsulates the historical context accumulated from past executions. It serves as a comprehensive memory pool that stores not only the objective execution trajectories $\tau$ but also the corresponding evaluative feedback, such as localized error attributions and peer-agent critiques.

The ultimate objective of the mapping $\Gamma$ is to iteratively improve the system's global performance objective $J$. To ensure the MAS develops generalized problem-solving capabilities rather than overfitting to a single scenario, $J(\cdot, \xi)$ is evaluated over a diverse task distribution $\xi \sim \mathcal{D}_{\mathrm{task}}$. This objective quantifies how well the system fulfills its goals, balancing metrics such as task success rate against execution costs. The evolutionary progress is formulated as improving the expected objective value across this open-ended task space:
\begin{equation}
\label{equal15}
\mathbb{E}_{\xi \sim \mathcal{D}_{\mathrm{task}}}
\left[
J(S^{(k+1)}, \xi)
\right]
\geq
\mathbb{E}_{\xi \sim \mathcal{D}_{\mathrm{task}}}
\left[
J(S^{(k)}, \xi)
\right].
\end{equation}

This formulation captures self-evolution as a process that goes beyond parametric updates to encompass structural and behavioral adaptation. It provides a unified language for describing how agents, communication protocols, orchestration topologies, and behavioral policies can be jointly tailored to maximize the expected utility within a target task distribution.

\subsection{MAS Self-Evolution Taxonomy}
\label{subsec:evolution-taxonomy}

As discussed in Section \ref{subsec:attribution-to-evolution}, recent advancements are driving MAS from static, rule-based designs toward autonomous, self-improving frameworks~\cite{hong2023metagpt,wu2024autogen}. However, the exact conceptualization of this ``evolution'' across current literature remains highly heterogeneous. While earlier works~\cite{chen2024autoagents, chen2024agentverse, Zhang2024ProAgent} 
have introduced dynamic team composition and proactive coordination, true 
self-evolution encompasses a broader and deeper spectrum of adaptation that 
goes beyond runtime adjustments. Existing methods range from refining 
individual agent prompts~\cite{lu2024morphagent, bo2024Reflective}, to 
updating shared model parameters~\cite{chen2025optima, xue2025comas}, to 
restructuring the entire communication topology~\cite{zhuge2024gptswarm}, 
reflecting a progression from behavioral modification to architectural 
transformation. To clarify this landscape and bridge the discrete-time process 
formalized in Section~\ref{subsec:evolution-definition} with practical 
implementations, we organize existing MAS self-evolution methods along 
two dimensions: \textbf{where the adaptation occurs}, and \textbf{what is being adapted}. Specifically, we decompose the evolutionary transition mapping $\Gamma$ 
into three hierarchical scopes, which we term the 
\textbf{Evolutionary Locus}:
\begin{equation}
\Gamma = \left\{\Gamma_{\text{agentic}}, \Gamma_{\text{systemic}}, \Gamma_{\text{meta}}\right\},
\end{equation}
where:
\begin{itemize}
    \item \textbf{Agentic Evolution ($\Gamma_{\text{agentic}}$)} modifies 
the internal components of individual agents within $\mathcal{A}^{(k)}$, 
such as their prompts, memory, or model parameters, to improve 
single-agent performance.
    \item \textbf{Systemic Evolution ($\Gamma_{\text{systemic}}$)} operates within a given system instance to restructure how agents collaborate. It evolves the orchestration topology $\mathcal{G}^{(k)}$, the communication protocols $\mathcal{C}^{(k)}$, and the shared collaboration strategies $\Pi^{(k)}$ in response to a specific task.
    \item \textbf{Meta Evolution ($\Gamma_{\text{meta}}$)} treats each 
complete system configuration $S$ as a single candidate within a 
population of possible designs. A dedicated meta-process (e.g., a 
meta-agent or search algorithm) evaluates, recombines, and selects 
among these candidates across diverse tasks, accumulating transferable 
design knowledge that generalizes beyond any single task instance.
\end{itemize}

 We trace this evolutionary trajectory across three scopes: beginning with the refinement of individual agent capabilities at the agentic level, ascending to the autonomous restructuring of organizational relationships at the systemic level, and finally culminating in the overarching paradigm shifts at the meta level.

\subsubsection{Agentic Self-Evolution}
\label{subsubsec:agentic-level}

An individual agent $a_i \in \mathcal{A}^{(k)}$ can be decomposed into 
three core components: its prompt ($p_i$), its accumulated memory 
($m_i$), and its underlying neural parameters ($\theta_i$).

\paragraph{Prompt}
Prompt evolution optimizes the textual prompt $p_i$, which defines an agent's role, instructions, and operational heuristics. Here, the underlying model parameters remain frozen, and evolution occurs purely through natural language feedback loops. For instance, MorphAgent~\cite{lu2024morphagent} introduces dynamic agent profiles that autonomously rewrite themselves based on quantitative metrics evaluating role clarity, differentiation, and task-role alignment. Similarly, NegotiationGym~\cite{mangla2025negotiationgym} optimizes agent negotiation strategies by iteratively refining prompts based on explicitly encoded utility functions derived from past interactions. In frameworks like COPPER~\cite{bo2024Reflective}, a ``Shared Reflector'' evaluates and provides feedback to all Actor agents, rather than each Actor reflecting independently. This feedback is then appended to their contextual prompts to refine future trajectories. While prompt evolution is highly interpretable and avoids catastrophic forgetting, its capability ceiling is strictly bounded by the underlying model's context window and parametric limits~\cite{liu2024lost}.

\paragraph{Memory}
Instead of relying solely on immediate context, memory evolution focuses on structurally refining and expanding $m_i$ over continuous executions. This approach allows agents to build persistent experience banks that transcend single conversational threads. AgentCourt~\cite{chen2024agentcourt} utilizes adversarial self-play in simulated courtrooms to continuously construct a three-level memory comprising legal statutes, practical experiences, and case precedents. Richelieu~\cite{guan2024richelieu} internalizes diplomatic experiences by persistently storing sub-goal evaluations and opponent credibility scores, retrieving them based on semantic similarity to guide future negotiations. Ultimately, memory evolution provides a robust pathway for cumulative, ``online-persistent'' learning, though it shifts the system's operational bottleneck toward retrieval efficiency~\cite{gao2023retrieval} and the risk of homogenizing historical biases~\cite{liang2024encouraging}.

\paragraph{Parameter}
Recognizing the strict limitations of textual contexts, a large and growing body of work targets the neural parameters $\theta_i$ directly (see Table~\ref{tab:agentic_evolution}). By employing optimization methods such as reinforcement learning (RL) or supervised fine-tuning (SFT) directly on the agent's base formulation, evolution is internalized into model weights. This approach frequently leverages internal interaction dynamics to generate reward signals, bypassing the need for human-annotated ground truths. CoMAS~\cite{xue2025comas} drives optimization using intrinsic ``interaction rewards'' evaluated by a judge module during inter-agent discussions. The Multi-Agent Evolve framework~\cite{chen2025multiagent} splits a single LLM into Proposer, Solver, and Judge roles, using objective-relative REINFORCE++ for self-improvement. Beyond task efficiency, parameter evolution is critical for robustness. AdvEvo-MARL~\cite{pan2025advevo} internalizes systemic safety by co-evolving attackers and defenders in a multi-agent reinforcement learning environment. Likewise, ECL~\cite{zhou2026epistemic} uses auxiliary RL rewards to help localized parameters evolve the ability to autonomously discern trustworthy peers from misleading ones, allowing smaller models to outperform history-agnostic larger baselines. Parameter evolution can also target communication efficiency: Optima~\cite{chen2025optima} iteratively trains a shared LLM backbone via SFT and DPO guided by a composite reward balancing task performance, token efficiency, and readability, achieving a 2.8$\times$ performance gain while reducing token usage by 90\%. Furthermore, Towards AGI~\cite{kar2026towards} adopts a hierarchical escalation strategy: it first handles systemic failures by dynamically synthesizing new Python functions at the tool level, then escalates to full parameter-level evolution using EA, SFT, and RL when tool synthesis alone proves insufficient. Although computationally expensive to train, parameter evolution embeds complex capabilities without inflating runtime inference costs.

Across all three agentic categories, although evolution targets individual agents in isolation, its driving force remains fundamentally embedded within the multi-agent context. Agents draw on peer interactions (including adversarial self-play~\cite{chen2024agentcourt, pan2025advevo}, mutual critique~\cite{bo2024Reflective, xue2025comas}, and collaborative knowledge sharing~\cite{li2025learn}) to produce rich, autonomous feedback signals. Thus, by adjusting its internal components in light of multi-agent dynamics, agentic evolution steadily enhances an agent's expected utility.


\begin{table*}[!t]
\centering
\caption{Classification and multi-dimensional comparison of agentic self-evolution methods (Section~\ref{subsubsec:agentic-level}).}
\label{tab:agentic_evolution}
\footnotesize
\renewcommand{\arraystretch}{1.0}
\setlength{\tabcolsep}{6pt}
\begin{adjustbox}{max width=\textwidth}
\begin{tabular}{@{}L{2.80cm} C{2.40cm} C{2.00cm} C{1.60cm} C{1.70cm} C{2.00cm} C{0.70cm}@{}}
\toprule
\textbf{Method} & \textbf{Evo.\ Target} & \textbf{Optimization} & \textbf{Feedback} & \textbf{Lifecycle} & \textbf{Domain} & \textbf{Year} \\
\midrule
\rowstrut MorphAgent~\cite{lu2024morphagent}          & \PromptB        & Reflection    & Environ.+Self  & On.-Persist  & Multi-domain  & \ygray{2024} \\
\rowcolor{zebragray}
\rowstrut COPPER~\cite{bo2024Reflective}              & \PromptB\;\ParamB & Reflection+RL & Environ.+Self  & Mixed        & QA, Math      & \ygray{2024} \\
\rowstrut AgentCourt~\cite{chen2024agentcourt}        & \MemoryB        & Reflection    & Peer           & On.-Persist  & Legal Sim.    & \ygray{2024} \\
\rowcolor{zebragray}
\rowstrut Richelieu~\cite{guan2024richelieu}           & \MemoryB        & Reflection    & Environ.       & On.-Persist  & Diplomacy     & \ygray{2024} \\
\rowstrut NegotiationGym~\cite{mangla2025negotiationgym} & \PromptB     & Reflection    & Environ.       & On.-Ephem    & Negotiation   & \ygray{2025} \\
\rowcolor{zebragray}
\rowstrut CoMAS~\cite{xue2025comas}                    & \ParamB        & RL            & Peer           & Off.-Persist & Math, Code    & \ygray{2025} \\
\rowstrut MAE~\cite{chen2025multiagent}                & \ParamB        & RL            & Self/Peer      & Off.-Persist & Math, Reason. & \ygray{2025} \\
\rowcolor{zebragray}
\rowstrut AdvEvo-MARL~\cite{pan2025advevo}             & \ParamB        & RL            & Environ.       & Off.-Persist & Safety        & \ygray{2025} \\
\rowstrut Optima~\cite{chen2025optima}                 & \ParamB        & SFT+RL        & Environ.       & Off.-Persist & QA, Math      & \ygray{2025} \\
\rowcolor{zebragray}
\rowstrut ECL~\cite{zhou2026epistemic}                 & \ParamB        & RL            & Environ.       & Off.-Persist & Trust Reason. & \ygray{2026} \\
\rowstrut Towards AGI~\cite{kar2026towards}            & \ParamB           & EA+SFT+RL     & Environ.+Self  & Off.-Persist & Embodied      & \ygray{2026} \\
\bottomrule
\end{tabular}
\end{adjustbox}
\vspace{3pt}
\parbox{\textwidth}{\footnotesize
\textit{Evo.\ Target}: evolutionary target: \PromptB{} = textual prompt/profile; \MemoryB{} = external memory bank; \ParamB{} = neural model parameters. \;
\textit{Optimization}: driving mechanism: Reflection = LLM reflection/generation; RL = reinforcement learning; SFT = supervised fine-tuning; EA = evolutionary algorithm. \;
\textit{Feedback}: selection pressure source: Environ.\ = environment reward; Self = self-evaluation; Peer = peer-agent critique. \;
\textit{Lifecycle}: temporal persistence: On.-Persist = online-persistent; On.-Ephem = online-ephemeral; Off.-Persist = offline-persistent; Mixed = combines multiple modes. \;
\textit{Domain}: application scenario: high-level application scenario.
}
\end{table*}

\subsubsection{Systemic Self-Evolution}
\label{subsubsec:systemic-level}

While agentic evolution strengthens what individual agents can do, systemic evolution restructures how they work together. Here, $\Gamma_{\text{systemic}}$ acts on the inter-agent organizational fabric, including the orchestration topology $\mathcal{G}^{(k)}$, the communication protocols $\mathcal{C}^{(k)}$, and the shared collaboration strategies $\Pi^{(k)}$. We decompose these collective structures into three actionable targets: the \textbf{topology}, which determines the information-flow graph among agents; the team \textbf{composition}, which governs how agents are assembled and how subtasks are routed; and the \textbf{shared memory}, which captures the collective knowledge infrastructure spanning the agent population. 

\paragraph{Topology}
The most actively explored systemic-level target is the communication 
topology, which determines the information-flow graph among agents. 
Puppeteer~\cite{dang2025multi} formulates multi-agent coordination as 
a dynamic sequential decision process and uses reinforcement learning 
to optimize a centralized routing policy that balances solution quality 
with computational cost. G-Designer~\cite{zhang2024g} models the MAS 
as a task-adaptive network and employs a Variational Graph Auto-Encoder 
to decode customized, sparse topologies from agent capability 
embeddings. MASS~\cite{zhou2025multi} jointly optimizes prompts and 
structure through a three-stage interleaved process that alternates 
between block-level prompt refinement, topology search, and global 
conditioning. On the decentralized side, 
AgentNet~\cite{yang2025agentnet} removes the central orchestrator 
entirely and lets each agent autonomously route tasks via Forward, 
Split, or Execute operations, with edge weights updated through 
exponential moving averages of success metrics. 
SELFORG~\cite{tastan2025stochastic} takes a similar decentralized 
approach by reconstructing the communication DAG each round, where 
agents independently assess peer contributions using Shapley-inspired 
semantic similarity.

\paragraph{Composition}
A complementary approach targets the agent team itself: dynamically 
determining which agents to create, modify, or remove, and how to 
structure their subtask dependencies. 
AutoAgents~\cite{chen2023autoagents} uses predefined meta-agents 
(Planner and Observers) to collaboratively synthesize task-specific 
agent teams in a drafting stage, then coordinates execution through 
self-refinement loops. DRTAG\&IAAG~\cite{perera2025auto} introduce 
auto-scaling through dynamic agent integration, where a central 
conversation manager generates and recruits new specialized agents 
during execution based on LLM reflection over the evolving dialogue 
context. While these methods generate teams in a single pass or on 
demand, EVOAGENT~\cite{yuan2025evoagent} formulates team composition 
as an iterative evolutionary process, applying textual crossover and 
mutation operators to a foundational agent to spawn specialized 
variants, with an LLM-driven quality-check module selecting the 
fittest population. EvoMAC~\cite{hu2024self} and 
ANN~\cite{ma2025agentic} take a different route by treating the 
multi-agent network as a textually differentiable computational graph. 
EvoMAC introduces textual 
backpropagation~\cite{pryzant2023automatic}, where a ``gradient 
agent'' analyzes each agent's causal impact using compiler feedback 
and an ``update agent'' performs structural mutations such as removing 
completed agents or adding agents for missing subtasks. ANN extends 
this by organizing the MAS as a hierarchical layered architecture 
with textual backpropagation at both global and local granularities.

\paragraph{Shared Memory}
The third systemic-level target shifts from structural evolution to 
constructing persistent collective knowledge that spans agent 
boundaries and task episodes. Unlike agentic memory evolution 
operating on individual banks $m_i$, shared memory captures 
collaborative interaction trajectories of the entire population. 
G-Memory~\cite{zhang2025g}, inspired by organizational memory theory, 
manages inter-agent interaction history through a three-tier graph 
structure (strategic knowledge, task meta-information, and 
fine-grained communication logs) that evolves by assimilating new 
collaborative experiences after each task. 
LIET~\cite{li2025learn} enables embodied agents to collaboratively 
maintain a shared cooperation knowledge list, using reflection modules 
to evaluate message quality and update collective interaction 
strategies across episodes. As a framework-agnostic plugin, G-Memory 
demonstrates that knowledge evolution and structural evolution can be 
pursued independently and potentially composed.

In summary, the three systemic-level targets reveal a key distinction 
between two evolutionary lifecycles. Methods targeting topology and 
composition typically operate on an ephemeral lifecycle, generating 
task-specific structures that are constructed and dissolved upon task 
completion. Conversely, shared memory frameworks establish a persistent 
lifecycle, accumulating organizational knowledge that spans multiple 
execution episodes. How to bridge these two lifecycles so that 
ephemeral structural adaptations can feed back into persistent 
collective knowledge remains an open question. The next section 
examines how meta evolution addresses these limitations by operating 
at a higher level of abstraction.


\begin{table*}[!t]
\centering
\caption{Comparison of systemic self-evolution methods (Section~\ref{subsubsec:systemic-level}).}
\label{tab:systemic_evolution}
\footnotesize
\renewcommand{\arraystretch}{1.0}
\setlength{\tabcolsep}{6pt}
\begin{adjustbox}{max width=\textwidth}
\begin{tabular}{@{}L{2.80cm} C{2.40cm} C{2.00cm} C{1.70cm} C{1.70cm} C{1.60cm} C{0.70cm}@{}}
\toprule
\textbf{Method} & \textbf{Evo.\ Target} & \textbf{Optimization} & \textbf{Coordination} & \textbf{Lifecycle} & \textbf{Scale} & \textbf{Year} \\
\midrule
\rowstrut AutoAgents~\cite{chen2023autoagents}          & \RoleB                 & Reflection     & Central       & On.-Ephem    & Dynamic  & \ygray{2023} \\
\rowcolor{zebragray}
\rowstrut EvoMAC~\cite{hu2024self}                      & \RoleB\;\GraphB        & TG             & Central       & On.-Ephem    & Dynamic  & \ygray{2024} \\
\rowstrut G-Designer~\cite{zhang2024g}          & \GraphB                & RL             & Central       & Off.-Persist & Dynamic  & \ygray{2024} \\
\rowcolor{zebragray}
\rowstrut DRTAG\&IAAG~\cite{perera2025auto}            & \RoleB                 & Reflection     & Central       & On.-Ephem    & Dynamic  & \ygray{2025} \\
\rowcolor{zebragray}
\rowstrut Puppeteer~\cite{dang2025multi}            & \GraphB                & RL             & Central       & On.-Persist  & Dynamic  & \ygray{2025} \\
\rowstrut MASS~\cite{zhou2025multi}                     & \GraphB                & Reflection     & Central       & Off.-Persist & Fixed    & \ygray{2025} \\
\rowstrut AgentNet~\cite{yang2025agentnet}              & \GraphB                & Heur           & Decentral     & On.-Persist  & Fixed    & \ygray{2025} \\
\rowcolor{zebragray}
\rowstrut EVOAGENT~\cite{yuan2025evoagent}              & \RoleB                 & EA             & Coop          & On.-Ephem    & Dynamic  & \ygray{2025} \\
\rowstrut ANN~\cite{ma2025agentic}                      & \RoleB\;\GraphB        & TG             & Hier          & On.-Ephem    & Dynamic  & \ygray{2025} \\
\rowcolor{zebragray}
\rowstrut G-Memory~\cite{zhang2025g}                    & \MemoryB              & Reflection     & Plugin        & On.-Persist  & Plugin   & \ygray{2025} \\
\rowstrut LIET~\cite{li2025learn}                       & \MemoryB               & Reflection     & Coop          & On.-Persist  & Fixed    & \ygray{2025} \\
\rowcolor{zebragray}
\rowstrut SELFORG~\cite{tastan2025stochastic}              & \GraphB                & Heur           & Decentral     & On.-Ephem    & Dynamic  & \ygray{2025} \\
\bottomrule
\end{tabular}
\end{adjustbox}
\vspace{3pt}
\parbox{\textwidth}{\footnotesize
\textit{Evo.\ Target}: evolutionary target: \GraphB{} = communication topology; \RoleB{} = agent role definition; \MemoryB{} = shared collective memory. \;
\textit{Optimization}: driving mechanism: Reflection = LLM reflection; RL = reinforcement learning; EA = evolutionary algorithm; TG = textual gradient; Heur = heuristic update. \;
\textit{Coordination}: coordination structure: Central = centralized orchestrator; Decentral = fully decentralized; Plugin = framework-agnostic; Hier = hierarchical; Coop = cooperative without central control. \;
\textit{Lifecycle}: temporal persistence: On.-Persist = online-persistent; On.-Ephem = online-ephemeral; Off.-Persist = offline-persistent. \;
\textit{Scale}: agent population scale: Fixed = predefined agent count; Dynamic = adapts per task; Plugin = inherits from host framework.
}
\end{table*}

\subsubsection{Meta Self-Evolution}
\label{subsubsec:meta-level}

Meta evolution differs from systemic evolution along two fundamental axes. First, the evolutionary unit shifts: rather than adjusting the internal organizational fabric of an existing system instance, $\Gamma_{\text{meta}}$ treats the complete system specification $S$ as a single candidate in a population of possible designs. Second, the search scope expands: a dedicated meta-process, typically a meta-agent or search algorithm, explores the space of system designs across diverse tasks and accumulates transferable design knowledge that persists and improves over successive design iterations. This accumulated knowledge constitutes the core distinction: whereas systemic methods produce task-specific structural adaptations that are discarded or remain local, meta methods build reusable design repositories that generalize across tasks and domains. Concretely, $\Gamma_{\text{meta}}$ optimizes two distinct targets: the \textbf{knowledge space}, a growing repository of accumulated design experience (archives, populations, search trees); and the \textbf{generator}, a trained model or optimized mechanism that produces system designs on demand.

The dominant approach to meta evolution centers on iteratively expanding the knowledge space, accumulating design discoveries to guide future search. GPTSwarm~\cite{zhuge2024gptswarm} initiates this structural search direction by conceptualizing language agents as optimizable graphs, utilizing the REINFORCE algorithm to search over a vast meta-space of possible architectures to continuously yield high-performing orchestrations. Expanding into open-ended code generation, ADAS~\cite{hu2025automated} introduces Meta Agent Search, where a meta-agent iteratively programs new agent systems in a general-purpose code space and stores evaluated candidates in an ever-growing archive, demonstrating robust cross-domain and cross-model transferability. AgentBreeder~\cite{rosser2025agentbreeder} extends this archive-based paradigm through a MAP-Elites-inspired quality-diversity algorithm that co-optimizes capability and safety, revealing that unsafe behaviors can spontaneously emerge alongside capability gains---highlighting systemic risks unique to meta evolution. Complementing these open-ended searches, AFlow~\cite{zhang2025aflow} constrains the design space to code-represented workflows over predefined operators and uses Monte Carlo Tree Search to discover configurations enabling small models to outperform GPT-4o at $4.55\%$ of inference cost. In contrast, MAS-ZERO~\cite{ke2025mas} shifts knowledge accumulation to inference-time, where a meta-agent iteratively designs and critiques MAS configurations per problem instance---uniquely capable of degrading to simpler strategies when simpler strategies suffice.

A complementary approach shifts the evolutionary focus to the generator itself: these methods train a parametric model that directly produces system designs on demand. Focusing on rapid query-level adaptation, FlowReasoner~\cite{gao2025flowreasoner} applies reinforcement learning to a meta-agent that generates optimal reasoning workflows in Python code tailored to incoming user queries. MaAS~\cite{zhang2025multi} introduces the agentic supernet, training an RL-based controller to dynamically sample task-specific architectures from a learned distribution, surpassing state-of-the-art at $15\%$ of computational cost with zero-shot adaptability to unseen tasks. MAS-GPT~\cite{ye2025mas} takes this further by reframing MAS design as a language generation task: a fine-tuned LLM outputs complete, task-specific MAS Python code within a single inference pass, bypassing iterative search.

Together, these two approaches demonstrate that multi-agent systems can be designed, evaluated, and iteratively refined in a fully automated loop. The knowledge-space approach offers broad exploratory coverage through growing archives, while the generator approach trades coverage for speed by compressing design knowledge into a single trained model. A key open question is whether these two strategies can be combined---using generators for fast initial proposals and archive-based search for continued refinement.


\begin{table*}[!t]
\centering
\caption{Classification and multi-dimensional comparison of meta self-evolution methods (Section~\ref{subsubsec:meta-level}).}
\label{tab:meta_evolution}

\footnotesize
\renewcommand{\arraystretch}{1.0}
\setlength{\tabcolsep}{6pt}

\begin{adjustbox}{max width=\textwidth}

\begin{tabular}{@{}L{2.80cm} C{1.80cm} C{2.00cm} C{2.00cm} C{1.70cm} C{0.70cm}@{}}
\toprule
\textbf{Method} & \textbf{Evo.\ Target} & \textbf{Optimization} & \textbf{Design Space} & \textbf{Lifecycle} & \textbf{Year} \\
\midrule
\rowstrut GPTSwarm~\cite{zhuge2024gptswarm}            & \KSB              & RL            & DAG           & On.-Persist  & \ygray{2024} \\

\rowcolor{zebragray}
\rowstrut ADAS~\cite{hu2025automated}                   & \KSB              & Reflection    & Code          & Off.-Persist      & \ygray{2025} \\

\rowstrut AgentBreeder~\cite{rosser2025agentbreeder}    & \KSB              & EA            & Code          & Off.-Persist     & \ygray{2025} \\

\rowcolor{zebragray}
\rowstrut AFlow~\cite{zhang2025aflow}                   & \KSB              & EA      & Workflow      & Off.-Persist & \ygray{2025} \\

\rowstrut MAS-ZERO~\cite{ke2025mas}                     & \KSB              & Reflection    & Code          & On.-Persist  & \ygray{2025} \\

\rowcolor{zebragray}
\rowstrut FlowReasoner~\cite{gao2025flowreasoner}       & \GenB             & RL            & Code          & Off.-Persist & \ygray{2025} \\

\rowstrut MaAS~\cite{zhang2025multi}                    & \GenB             & RL            & Supernet      & Off.-Persist & \ygray{2025} \\

\rowcolor{zebragray}
\rowstrut MAS-GPT~\cite{ye2025mas}                      & \GenB             & SFT           & Code          & Off.-Persist & \ygray{2025} \\
\bottomrule
\end{tabular}
\end{adjustbox}

\vspace{3pt}
\parbox{\textwidth}{\footnotesize
\textit{Evo.\ Target}: evolutionary target: \KSB{} = knowledge space; \GenB{} = generator. \;
\textit{Optimization}: driving mechanism: Reflection = LLM reflection/meta-design; RL = reinforcement learning; SFT = supervised fine-tuning; EA = evolutionary algorithm. \;\textit{Design Space}: architectural search space: Code = general-purpose Python code; Workflow = operator-constrained workflow DAG; Supernet = probabilistic architecture distribution; DAG = directed acyclic graph. \;
\textit{Lifecycle}: temporal persistence: Off.-Persist = offline with persistent components; On.-Persist = online with persistent components.
}
\end{table*}

\subsection{Analyzing Evolutionary Dynamics}
\label{subsec:evo-dynamics}

The preceding taxonomy (Section~\ref{subsec:evolution-taxonomy}) is organized primarily along the Evolutionary Locus. It classifies methods by whether they evolve individual agents, systemic structures, or entire system designs. Within each locus, the specific adaptation target (e.g., prompts, parameters, topologies) serves as a secondary classification. This subsection introduces an orthogonal analytical axis: the \textbf{driving mechanism}, or how the transition mapping $\Gamma$ is computationally realized. By cutting across all three evolutionary loci, we examine how different algorithmic strategies produce distinct evolutionary behaviors. 

We analyze each mechanism using Campbell's Variation--Selection--Retention (VSR) framework~\cite{campbell1960blind}, a fundamental model of evolutionary epistemology. Specifically, we evaluate how each mechanism introduces candidate variations, what selection pressures it imposes, and how it retains evolutionary gains. Across the surveyed literature, we identify six classes of driving mechanisms: \textbf{LLM Reflection}, \textbf{Reinforcement Learning} (RL), \textbf{Supervised Fine-Tuning} (SFT), \textbf{Evolutionary Algorithms} (EA), \textbf{Textual Gradient} (TG), and \textbf{Heuristic Update}.

\paragraph{LLM Reflection}
LLM Reflection is the most broadly deployed mechanism, present across all three evolutionary loci. Its distinguishing characteristic is that variation is semantically directed. Rather than perturbing system configurations at random, the LLM uses its internalized knowledge to generate targeted modifications. When a prompt fails, a reflection-based method diagnoses the failure and proposes a contextually relevant revision~\cite{lu2024morphagent, mangla2025negotiationgym}.  This principle also extends to the meta level. Here, LLM-driven meta-agents architect new system designs informed by accumulated prior experience~\cite{hu2025automated, ke2025mas}.

However, this semantic directedness creates a structural tension during selection. Because the same LLM both generates and evaluates candidates, the selection signal is predominantly self-referential. This self-contained loop eliminates the need for external supervision but introduces sycophancy risks. Agents may converge on plausible but suboptimal outputs, often abandoning correct judgments when confronted with confident peer assertions~\cite{zhou2026epistemic}. Methods that decouple generation from evaluation, such as relying on objective compiler feedback~\cite{hu2024self}, successfully circumvent this limitation.

Retention in reflection-based evolution is predominantly archival, storing evolutionary gains in explicit, retrievable memory structures~\cite{chen2024agentcourt, zhang2025g}. This offers interpretability and supports selective forgetting, but it faces a saturation ceiling. As the knowledge base grows, retrieved experiences converge and marginal improvements diminish~\cite{guan2024richelieu}. Ultimately, LLM Reflection offers the lowest barrier to entry since it requires no training infrastructure. However, its capability ceiling remains strictly bounded by the generative capacity of the underlying model.

\paragraph{Reinforcement Learning}
RL is the dominant mechanism for evolving model parameters and communication topologies. Its central evolutionary advantage is the ability to derive selection pressure endogenously from the multi-agent interaction process itself.  In several RL-driven systems, the reward signal emerges directly from agent collaboration and competition rather than external human labels. For example, peer agents may evaluate each other's contributions~\cite{xue2025comas}, or co-evolving roles may create self-sustaining adversarial arms races~\cite{chen2025multiagent, pan2025advevo}. Alternatively, composite reward functions can simultaneously encode task performance, communication efficiency, and output readability~\cite{chen2025optima}.

This capacity for endogenous reward construction is particularly significant at the systemic level. RL enables the evolution of discrete structural properties, such as communication topologies, by reformulating them into continuous optimization problems amenable to gradient-based methods~\cite{dang2025multi, zhang2024g}. Conversely, RL adoption remains scarce at the meta level~\cite{zhang2025multi}. This scarcity highlights a fundamental challenge: standard RL faces severe sample efficiency bottlenecks when navigating the vast, discrete combinatorial spaces typical of complete system architectures, especially when evaluating each configuration incurs prohibitive computational costs.

RL retains evolutionary gains parametrically by encoding them directly into model weights. This incurs no additional inference cost but is irreversible and susceptible to catastrophic forgetting. Furthermore, improper reward design poses significant risks. For instance, multi-agent scaffolds have exploited safety objectives by collapsing into trivially safe but unhelpful responses~\cite{rosser2025agentbreeder}. This highlights the fragility of selection signals that are poorly aligned with the true evolutionary objective.

\paragraph{Evolutionary Algorithms}
EA closely resembles biological Darwinian evolution. It maintains explicit populations, applies crossover and mutation operators, and selects candidates based on fitness evaluation across generations. In the context of LLM-based MAS, classical EA operators are fundamentally reinterpreted. Instead of applying stochastic operators to genotypic encodings, LLM-augmented EA uses the language model to perform semantically meaningful recombination~\cite{yuan2025evoagent}. This produces offspring that are coherent syntheses of complementary strengths, rather than random permutations of their parents.

The most distinctive evolutionary contribution of EA is its capacity to maintain population diversity. While other mechanisms converge strictly toward a single solution trajectory, EA-based methods can preserve multiple viable candidates and navigate multi-modal fitness landscapes. For example, AgentBreeder~\cite{rosser2025agentbreeder} implements quality-diversity search with embedding-based behavioral niching. It demonstrates that multi-objective selection over both capability and safety yields stronger convergence than single-objective optimization. Comparative studies further indicate that EA produces the highest behavioral diversity among evolved agents. In comparison, RL excels on high-difficulty tasks and curriculum-based methods deliver the strongest cross-task generalization~\cite{kar2026towards}.

A closely related strategy replaces the evolutionary population with a structured search tree. AFlow~\cite{zhang2025aflow} constrains the design space to code-represented workflows over predefined operators and uses Monte Carlo Tree Search (MCTS) with UCB-based node selection to explore this space, providing formal exploration--exploitation guarantees. The search tree itself serves as structured memory for all explored configurations, though the requirement of predefined operator libraries constrains the theoretical openness of the process.

Despite the exploratory strengths of population-based and tree-search methods, both remain the least adopted mechanisms. They are fundamentally constrained by the substantial cost of evaluating each candidate per generation or expansion, making scalability a persistent bottleneck.

\paragraph{Textual Gradient}
Textual Gradient mechanisms bridge neural network optimization and MAS evolution by replacing numerical loss signals with natural-language feedback~\cite{pryzant2023automatic}. Its defining property is that variation is driven by causal attribution. Dedicated analysis agents trace the causal impact of each component on the system's output. They identify precisely which subtask failed, which agent introduced errors, and which capability is missing~\cite{hu2024self}. Subsequently, structural mutations, such as agent removal, prompt revision, or agent addition, follow directly from this diagnosis to produce highly targeted changes. This causal specificity is further deepened by integrating deep learning techniques like dual-granularity optimization and momentum-based stabilization~\cite{ma2025agentic}.

Selection under textual gradients is notably the most objective among all mechanisms. Some methods explicitly reject self-generated critiques as biased, relying solely on verifiable execution logs~\cite{hu2024self}. However, both current implementations share a significant limitation: they operate under ephemeral retention. They evolve each task instance independently without cross-task knowledge accumulation. This absence of persistent memory distinguishes textual gradients from all other mechanisms and remains its primary barrier to sustained improvement.

\paragraph{Heuristic Update}
Heuristic updates encompass non-gradient, rule-based strategies for adapting system configurations. Rather than employing continuous neural optimization or LLM-driven generation, these methods rely on rigid statistical or algebraic rules. For example, AgentNet~\cite{yang2025agentnet} utilizes an exponential moving average (EMA) of task success rates to dynamically decay edge weights and prune underperforming connections, while SELFORG~\cite{tastan2025stochastic} applies fixed Shapley-inspired semantic similarity thresholds to deterministically reconstruct the communication DAG each round. While computationally efficient, heuristic mechanisms lack the exploratory capacity of search-based or evolutionary methods, acting primarily as regularization filters rather than generative engines.

\paragraph{Supervised Fine-Tuning}
SFT represents the most conservative evolutionary strategy. It transmits known-good behavioral patterns through the direct imitation of curated trajectories. At the agentic level, methods like Optima~\cite{chen2025optima} and Towards AGI~\cite{kar2026towards} incorporate SFT as part of their training pipeline to internalize proven interaction patterns, while at the meta level, MAS-GPT~\cite{ye2025mas} fine-tunes an LLM on collected MAS design trajectories to directly generate complete system configurations. In evolutionary terms, SFT functions less as a search mechanism and more as a knowledge transmission channel. Analogous to horizontal gene transfer in biology, it propagates proven strategies without autonomous discovery. Its variation is minimal, arising solely from the diversity of the training data, and its selection is implicitly defined during the curation process. Ultimately, while SFT enables efficient knowledge deployment, it lacks the capacity for open-ended exploration beyond its training distribution.

Table~\ref{tab:mechanism-landscape} maps all surveyed methods onto a cross-sectional matrix defined by driving mechanism and evolutionary locus. Among the methods reviewed in this survey, several patterns stand out. RL at the agentic level---targeting model parameters---forms the densest cluster, closely followed by LLM Reflection at both the agentic and systemic levels. LLM Reflection, RL, and EA are the only mechanisms that appear across all three loci, indicating their broad applicability across different evolutionary scopes. By contrast, textual gradients and heuristic updates are currently confined to the systemic level.

Another notable pattern is the near-absence of hybrid mechanisms. Only isolated efforts combine multiple driving mechanisms~\cite{kar2026towards, bo2024Reflective}, even though their strengths appear complementary (for example, EA's population diversity, RL's parametric refinement, and Reflection's semantic directedness). The gaps visible in the table (textual gradients at the meta level, EA at the systemic level, and mechanism hybridization across the board) suggest concrete directions for future investigation.

\begin{table*}[!t]
\centering
\caption{Mechanism Landscape of MAS Self-Evolution. This table illustrates the distribution of 31 surveyed frameworks across three evolutionary loci and six driving mechanisms. Each colored circle represents a specific mechanism-locus intersection utilized by a framework. Because some frameworks employ hybrid mechanisms or span multiple loci, the total number of plotted circles (36) exceeds the number of unique frameworks.}

\label{tab:mechanism-landscape}
\newcommand{\dotmark}[1]{%
  \tikz[baseline=-0.4ex]{\fill[#1] (0,0) circle (2.2pt);}\hspace{1pt}%
}
\newcommand{\dA}{\dotmark{clAzure}}
\newcommand{\dS}{\dotmark{clGoldDark}}
\newcommand{\dM}{\dotmark{dotS}}
\renewcommand{\arraystretch}{1.6}
\setlength{\tabcolsep}{10pt}
\footnotesize
\setlength{\tabcolsep}{10pt}
\begin{tabular}{l*{6}{c}}
\toprule
\textbf{Locus}
  & \textbf{Refl}
  & \textbf{RL}
  & \textbf{SFT}
  & \textbf{EA}
  & \textbf{TG}
  & \textbf{Heur} \\
\midrule
\rowcolor{zebragray}
\textcolor{clAzure}{\textbf{Agentic}}
  & \dA\dA\dA\dA\dA\dA
  & \dA\dA\dA\dA\dA\dA\dA
  & \dA\dA
  & \dA
  &
  & \\
\textcolor{clGoldDark}{\textbf{Systemic}}
  & \dS\dS\dS\dS\dS
  & \dS\dS
  &
  & \dS
  & \dS\dS
  & \dS\dS \\
\rowcolor{zebragray}
\textcolor{dotS}{\textbf{Meta}}
  & \dM\dM
  & \dM\dM\dM
  & \dM
  & \dM\dM
  &
  & \\
\bottomrule
\end{tabular}
\vspace{-0.8em}
\end{table*}

\subsection{Evaluation}
\label{subsec:evaluation}

As multi-agent systems transition from static architectures to self-evolving frameworks, the methodologies used to evaluate them must also adapt. Current literature employs a diverse set of evaluation environments~\cite{chen2024agentcourt, zhang2023building, jimenez2023swe} that can be broadly categorized into four dimensions: static benchmarks, interactive environments, embodied sandboxes, and adversarial arenas.

The majority of evolving MAS are evaluated on static datasets such as MMLU~\cite{hendrycks2020measuring}, GSM8K~\cite{cobbe2021training}, MATH~\cite{hendrycks2021measuring}, and HumanEval~\cite{chen2021evaluating}. These benchmarks effectively measure absolute capability gains in reasoning and coding following agentic or systemic evolution. However, because they are structured as single-turn, ``open-book'' examinations, they fail to capture the dynamic interaction and long-term adaptability inherent to evolving systems.

To address the limitations of static tests, researchers increasingly utilize interactive environments that require coherent, multi-step problem solving. Frameworks like SWE-Bench~\cite{jimenez2023swe} and rSDE-Bench~\cite{hu2024self} evaluate complex software development workflows, while simulations such as AgentCourt~\cite{chen2024agentcourt} and NegotiationGym~\cite{mangla2025negotiationgym} assess continuous multi-agent bargaining and debate. These testbeds emphasize overall workflow success and the stability of multi-turn collaborations, directly validating systemic adaptations such as topology rewiring.

The most rigorous test of open-ended evolution occurs in embodied virtual environments (e.g., TDW-MAT~\cite{zhang2023building}). In these scenarios, systems must demonstrate long-term survival strategies, adapting their collective shared memory~\cite{li2025learn} to navigate unseen terrains and resource constraints. This evaluates the true vitality of the evolutionary process.

As agents gain the autonomy to evolve, ensuring alignment becomes critical. Evaluations in adversarial arenas---using datasets like SaladData~\cite{li2024salad} or simulated adversarial self-play~\cite{pan2025advevo}---measure whether evolving systems develop deceptive or unsafe behaviors alongside capability improvements~\cite{rosser2025agentbreeder}.

\subsection{Discussion}
\label{subsec:discussion}

The transition from static, rule-bound multi-agent systems to autonomous, self-evolving entities marks a fundamental paradigm shift in artificial intelligence. As synthesized in Section~\ref{subsec:evolution-taxonomy}, this evolution manifests across three distinct loci: agentic enhancement of localized capabilities, systemic rewiring of organizational topologies, and meta-level exploration of complete architectural spaces. While reinforcement learning~\cite{xue2025comas}, textual 
gradients~\cite{hu2024self}, and LLM 
reflection~\cite{bo2024Reflective} have shown that evolutionary 
mechanisms can automate MAS design, the field remains in its early 
stages. In this section, we discuss the critical limitations of existing paradigms and chart four foundational directions for future research.

\paragraph{The Shift Towards Persistent Lifelong Evolution}
\label{paragraph:lifelong-evolution}

Currently, the vast majority of systemic and meta-level evolutionary frameworks operate on an online-ephemeral or offline-persistent lifecycle. In online-ephemeral systems, dynamic topologies and specialized agent roles are instantiated to solve a specific query, only to be entirely discarded once the task concludes~\cite{dang2025multi, hu2024self, yuan2025evoagent}. Conversely, offline-persistent systems search for a globally optimal architecture within a laboratory setting, locking the design in place before deployment~\cite{zhang2024g, zhang2025aflow}. Both lifecycles fundamentally fall short of true biological evolution, which is characterized by continuous, lifelong adaptation in open-ended environments.

The future of MAS evolution must shift towards persistent lifelong learning. A lifelong evolving MAS would continuously accumulate collaborative knowledge across diverse task episodes without resetting its organizational memory. Achieving this requires overcoming the pervasive challenge of catastrophic forgetting. When evolution targets neural parameters via RL, updates geared toward a new domain frequently erase specialized behaviors learned in previous domains~\cite{xue2025comas}. While memory-level evolution~\cite{chen2024agentcourt, zhang2025g, guan2024richelieu} mitigates forgetting by persistently archiving successful interaction trajectories, explicit memory banks face their own bottleneck: as the knowledge space expands, retrieval accuracy degrades~\cite{gao2023retrieval}, and the computational overhead of processing extensive historical contexts grows linearly~\cite{booch2021thinking}. 

Bridging this gap requires hybridizing parametric and structural evolution. Future research should explore dual-process architectures that separate rapid, online topological adaptations from slow, offline parametric consolidations. Only through such an accumulation of collaborative wisdom can a MAS survive and scale in unpredictable, open-world deployments.

\paragraph{Balancing Scalability and the Cost of Evolution}
\label{paragraph:scalability-cost}

Self-evolution introduces substantial computational overhead. The search space for MAS configurations is combinatorially vast, encompassing varying agent populations, diverse prompt instructions, and heterogeneous communication graphs. Evaluating the fitness of a single candidate architecture often requires executing a complete multi-agent workflow, consuming significant inference time and API tokens. Consequently, driving mechanisms like Evolutionary Algorithms (EA) and Monte Carlo Tree Search (MCTS) struggle to scale when applied to meta-level structural discovery~\cite{rosser2025agentbreeder, zhang2025aflow, kar2026towards}.

To make evolution practically viable, the field must confront the trade-off between the depth of evolutionary exploration and the computational cost of execution. One promising direction is the development of highly efficient architectural generators. By reframing structural search as a generative modeling problem, frameworks like MAS-GPT~\cite{ye2025mas} and Agentic Supernets~\cite{zhang2025multi} demonstrate that deep neural networks can learn the underlying distribution of effective multi-agent topologies. Once trained, these generators can instantaneously sample high-quality, task-specific architectures in a single forward pass, entirely bypassing iterative search at inference time.

However, relying solely on parametric generators sacrifices the rigorous verification guarantees provided by structured search. A balanced future approach likely lies in hierarchical optimization, combining lightweight, generator-driven real-time adjustments with more rigorous, computationally expensive offline searches guided by structured methods such as MCTS~\cite{zhang2025aflow}.

\paragraph{Safety and Alignment}
\label{paragraph:safety-alignment}

As MAS acquire the autonomy to rewrite their own prompts, modify their communication protocols, and synthesize new peer agents, they also acquire the capability to bypass human-engineered safety guardrails. Traditional alignment methodologies, such as RLHF~\cite{ouyang2022training}, are typically applied to isolated, single-agent foundation models. These static constraints are highly vulnerable when transposed into a self-evolving, multi-agent context~\cite{rosser2025agentbreeder}.

Evolution is fundamentally an optimization process driven by selection pressure. If the evolutionary objective function strictly prioritizes task success (e.g., maximizing a code compilation pass rate or winning a negotiation), the system is vulnerable to reward hacking. In pursuit of the objective, agents might evolve deceptive communication strategies~\cite{liang2024encouraging} or collude to circumvent programmatic restrictions~\cite{chan2023harms}. 

To safely deploy self-evolving MAS, alignment must be elevated from a static property of individual models to a dynamic property of the evolutionary process itself. Future research must formalize evolutionary alignment, ensuring that safety constraints serve as an intrinsic, continuous selection pressure, potentially through adversarial co-evolution techniques.

\paragraph{Towards Standardized Open-Ended Evaluation}
\label{paragraph:standardized-evaluation}

The transition from static to self-evolving MAS necessitates a corresponding evolution in how these systems are evaluated. As detailed in Section~\ref{subsec:evaluation}, the prevalent reliance on static benchmarks (e.g., MMLU~\cite{hendrycks2020measuring}, GSM8K~\cite{cobbe2021training}) is fundamentally inadequate for assessing the vitality of an evolving system. A static QA dataset acts as an isolated snapshot; it cannot evaluate how effectively a system adapts its structure over time, nor can it measure the system's resilience when confronted with shifting task distributions or simulated resource constraints.

The community urgently requires standardized, open-ended evaluation protocols explicitly designed for evolutionary MAS. These benchmarks must shift the focus from ``absolute task accuracy at Generation $K$'' to evolutionary sample efficiency, which measures how rapidly and robustly a system discovers a viable architecture when introduced to a novel domain. 

Furthermore, environments like embodied sandboxes~\cite{li2025learn} and interactive software engineering testbeds~\cite{hu2024self} must be standardized to evaluate behavioral diversity. A truly successful evolution should not merely converge on a single monolithic architecture, but rather populate an archive of diverse, highly specialized topological designs capable of handling a broad spectrum of edge cases~\cite{wang2019paired}.

\section{Conclusion}
This survey has presented a unified treatment of LLM-based multi-agent 
systems organized around the LIFE progression: \textit{Lay} individual 
capability foundations, \textit{Integrate} agents through collaboration, 
\textit{Find} faults through attribution, and \textit{Evolve} through 
autonomous self-improvement. For each stage, we have provided systematic 
taxonomies and comparative analyses, covering core agent capabilities, 
organizational mechanisms for collaboration, methodological families for 
failure attribution, and hierarchical scopes for self-evolution. Beyond 
reviewing each stage in isolation, we have traced the causal dependencies 
that link them: collaborative structures determine what failures are 
observable, attribution narrows the search space for targeted 
improvement, and evolutionary gains in turn reshape the collaborative 
fabric. This cross-stage perspective distinguishes the present survey 
from prior work that addresses these topics independently.

We believe the longer-term promise of multi-agent systems lies not in 
engineering better coordination among individually designed agents, but 
in cultivating collective intelligence, where system-level capabilities 
emerge from agent interactions and exceed what any constituent agent 
could achieve alone. The LIFE framework, by making these cross-stage 
dependencies explicit, provides a foundation for pursuing this goal.

\section*{Acknowledgments}
This work was supported in part by the Fundamental and Interdisciplinary Disciplines Breakthrough Plan of the Ministry of Education of China (JYB2025XDXM116), the National Natural Science Foundation of China (62306229, 62477037, 62450005), the Youth Talent Support Program of Shaanxi Science and Technology Association (20240113), the China Postdoctoral Science Foundation (2024M752585, 2025T180425), and the CAAI-Lenovo Blue Sky Research Fund (2025CAAI-LENOVO-06).



 
\bibliographystyle{IEEEtran}
\bibliography{reference}
\vfill

\end{document}